\documentclass[]{siamart1116}
\usepackage{mathtools}
\usepackage{amssymb}
\usepackage{stmaryrd}
\usepackage{color}
\usepackage{pdflscape}
\usepackage{epsfig}

\usepackage{epstopdf}
\usepackage{multirow}
\usepackage{hhline}
\usepackage{txfonts}
\usepackage{mathrsfs}
\usepackage{color}
\usepackage{algorithmic}

\usepackage{subcaption}
\usepackage{mathrsfs}
\usepackage{sidecap}
\usepackage{graphicx, array}

\usepackage{mathabx}

\newtheorem{thm}{Theorem}[section]

\newtheorem{rem}[thm]{Remark}
\newtheorem{exm}[thm]{Example}

\numberwithin{equation}{section}
\title{VPINN\lowercase{s}: Variational Physics-Informed Neural Networks \\ For Solving Partial Differential Equations \thanks{This work was supported by the Applied Mathematics Program within the Department of Energy on the PhILMs project (DESC0019434 and DE-SC0019453).}}

\author{
	Ehsan Kharazmi
	\footnote{D\lowercase{ivision of} A\lowercase{pplied} M\lowercase{athematics},	
		B\lowercase{rown} U\lowercase{niversity}, 170 H\lowercase{ope} \lowercase{ST.}, P\lowercase{rovidence} , RI 02906, USA. E\lowercase{mail: ehsan\_kharazmi@brown.edu}}
	, Zhongqiang Zhang
	\footnote{D\lowercase{epartment of} M\lowercase{athematical} S\lowercase{ciences},	
		W\lowercase{orcester} P\lowercase{olytechnic} I\lowercase{nstitute}, 100 I\lowercase{nstitute} R\lowercase{d}, W\lowercase{orcester} , MA 01609, USA. E\lowercase{mail: zzhang7@wpi.edu }}
	, George EM Karniadakis\footnote{D\lowercase{ivision of} A\lowercase{pplied} M\lowercase{athematics},	
		B\lowercase{rown} U\lowercase{niversity}, 170 H\lowercase{ope} \lowercase{ST.}, P\lowercase{rovidence} , RI 02906, USA. E\lowercase{mail: george\_karniadakis@brown.edu}} \footnote{P\lowercase{acific} N\lowercase{orthwest} N\lowercase{ational} L\lowercase{aboratory}, R\lowercase{ichland}, WA 99354, USA.}
}

\allowdisplaybreaks
\begin{document}
	
\maketitle

\begin{abstract}
Physics-informed neural networks (PINNs)~\cite{raissi2019physics} use automatic differentiation to solve partial differential equations (PDEs)
by penalizing the PDE in the loss function at a random set of points in the domain of interest. 
Here, we develop a Petrov-Galerkin version of PINNs based on the nonlinear approximation of deep neural networks (DNNs)
by selecting the {\em trial space} to be the space of neural networks and the {\em test space} to be the space of Legendre polynomials.  We formulate the \textit{variational residual} of the PDE using  the DNN approximation by incorporating the variational form of the problem into the loss function of the network and construct a \textit{variational physics-informed neural network} (VPINN). By integrating by parts the integrand in the variational form, we lower the order of the differential operators represented by the neural networks, hence effectively reducing the training cost in VPINNs while increasing their accuracy compared to PINNs that essentially employ delta test functions. For shallow networks with one hidden layer, we analytically obtain explicit forms of the \textit{variational residual}. We demonstrate the performance of the new formulation for several examples that show clear advantages of VPINNs over PINNs in terms of both accuracy and speed.


\end{abstract}

%\begin{AMS}
%\end{AMS}

\pagestyle{myheadings}
\thispagestyle{plain}

%
%%%%%%%%%%%%%%%%%%%%%%%%%%%
\section{Introduction}
\label{Sec: Introduction NN}
%%%%%%%%%%%%%%%%%%%%%%%%%%%
%

Developing efficient and accurate numerical methods for simulating the multiscale dynamics of physical and biomedical phenomena has been a long-standing challenge in scientific computing. Several methods have been developed in the literature that aim to provide stable, convergent and accurate approximations of the partial differential equations (PDEs) that govern multiscale dynamics. The most popular and standard approaches are finite difference, finite element and spectral methods. These methods generally convert the mathematical model into its discrete counterpart using a grid over the computational domain and then solve the resulting linear or nonlinear system of PDEs for the unknown state variables at the nodes of the grid or for the unknown coefficients of functional expansions describing these state variables. Spectral methods, in particular, consider a linear combination of some known basis (trial) functions as a modal/nodal approximation, where the span of these basis functions constructs a convergent discrete solution space of the problem. The resulting system of unknown coefficients is then obtained by \textit{testing} the equation with suitable test functions. In a Petrov-Galerkin setting, the trial basis and test function spaces are distinct and hence different choices yield different numerical schemes, while in the Galerkin framework the trial basis functions and test functions are identical. These methods have been significantly advanced over the past five decades and successfully employed in simulating physical and biological problems, e.g. \cite{karniadakis2013spectral} and the references therein.

In a more general setting, a different technique put forward more recently is the nonlinear approximation \cite{devore1998nonlinear,devore2009nonlinear}, which extends the approximants to belong to a nonlinear space and does not limit the approximation to linear spaces, leading to a more robust estimation by sparser representation and cheaper computation. The nonlinear approximation contains different approaches including wavelet analysis \cite{daubechies1992ten}, dictionary learning \cite{tariyal2016greedy}, adaptive pursuit and compressed sensing \cite{davis1994adaptive, ohlsson2013nonlinear, candes2006compressive, candes2008introduction}, adaptive splines \cite{devore1998nonlinear}, radial basis functions \cite{devore2010approximation}, Gaussian kernels \cite{hangelbroek2010nonlinear}, and neural networks \cite{mhaskar1992approximation, mhaskar2019function, daubechies2019nonlinear}. For example, for functions in Besov spaces with smoothness $s$ and input dimensionality of $d$, it has been shown that an approximation of order $\mathcal{O}^{(N-s/d)}$ that is almost optimal can be constructed \cite{devore2010approximation, hangelbroek2010nonlinear, lin2014almost}. Moreover, for H\"{o}lder continuous functions of order $1$ on $[0, 1]^d$, an $\mathcal{O}^{(-\frac{1}{2d})}$ approximation was constructed in \cite{xie2013rate} and then further improved to $\mathcal{O}^{(-\frac{2}{d})}$ in \cite{shen2019nonlinear}. While nonlinear approximation presents more capabilities, its nonlinear nature imposes extra complications, and achieving the optimal approximation rate, especially in high dimensional spaces, remains an ongoing challenging problem.

A neural network (NN), and in particular a deep NN (DNN), can be built to construct a relation $u_{NN}(x): \Omega \rightarrow \mathbb{R}$ between a high-dimensional input $x \in \Omega \subset \mathbb{R}^d$ with some positive integer $d$ and the output $u_{NN} \in \mathbb{R}$ via algebraic operations and nonlinear mapping. One of the main advantages of DNNs is that they can represent a large family of functions with a relatively small number of parameters. It was shown in \cite{cybenko1989approximation} that a two-layer (containing only a single hidden layer) network has great expressive power, and in particular a two-layer network with sigmoid activations could approximate any continuous function. It was also shown in \cite{arora2016understanding} that a one-to-one correspondence between the class of ReLU DNNs and piecewise linear (PWL) functions can be established, which shows that any PWL function can be representable by a ReLU DNN. DNNs are generally comprised of input, output, and hidden layer(s), where each layer may contain several neurons. Assuming a fully connected network, the connection between neurons forms a complete graph. In this case, each interior hidden layer receives the information from the previous layer and passes it to the next layer after applying a combination of scaling (by some weights), shifting (by some biases), and a nonlinear mapping. The final output of the network is then a nonlinearly-mapped weighted summation of input and hidden layers with certain biases, where the weights $\textbf{w}$ and biases $\textbf{b}$ are called the unknown parameters of the network. As DNNs use nonlinear compositional mappings, usually there is no closed form solution of the network parameters, and they are obtained via iterative algorithms, e.g. back propagation, that minimizes a properly defined objective (loss) function. Generally, the loss function is designed as a discrepancy measure between the network output $u_{NN}$ and the given data $u$, where its minimization returns the parameters that best approximate the solution. This process of loss minimization is called network training, and a trained network has usually ``optimum" parameters that can accurately represent the training data sets, while its performance over any other admissible input data (test data) should be examined.

In most of practical applications of NN, the constructed network can be thought of being ignorant of any possibly existing underlying mathematical model expressing physical laws, and therefore gives a model-ignorant algorithm as the network training process does not require any knowledge of the underlying mathematical model. Recently, NNs have been constructed such that they incorporate the underlying mathematical model in the network graph and construct a physics-informed neural network (PINN) \cite{raissi2019physics}. The loss function contains extra terms to merge the mathematical model as an additional constraint to ensure that the network output satisfies the mathematical model as well. For accurate and reliable training, model-ignorant networks require a fairly large number of high fidelity training data points, which are in general expensive to obtain. The physics-informed approach, however, replaces such large volume data requirement by infusing information from the underlying physics by forcing the network output to satisfy the corresponding mathematical model at some penalizing points that are available at minimal cost, and thus the network only requires a minimal number of expensive training data. The recent development of PINNs has been well established for forward and inverse problem of solving differential equations \cite{raissi2017machine,raissi2019deep,yazdani2018hidden,raissi2018hidden} and since then a number of extensions has been made to tackle several physical and biomedical problems \cite{pang2019fpinns, jagtap2019adaptive, jagtap2019adaptiveinpress}. All of these formulations employ the strong form of the mathematical models into the network, i.e., the conservation laws are enforced at random points in the space-time domain. 

In this paper, we develop a variational physics-informed neural network (VPINN) within the Petrov-Galerkin framework, where the solution is represented by nonlinear approximation via a (deep) neural network, while the test functions still belong to linear function spaces. Unlike a PINN that incorporates the strong form of the equation into the network, we incorporate the variational (weak) formulation of the problem and construct a \textit{variational loss function}. The advantages of the variational formulation are multi-fold: 
\begin{itemize}
	\item The order of differential operators can be effectively reduced by proper integration-by-parts, which reduces the required regularity in the (nonlinear) solution space. This will further mitigate the complexity of PINNs in taking high-order derivatives of nonlinear compositional functions, leading to less computationally expensive algorithms.
	
	\item In the case of shallow networks and for special activation and test functions, the loss function can be expressed analytically, which opens up the possibility of performing numerical analysis of such formulations.
	
	\item The large number of penalizing points required by PINNs is replaced by a relatively small number of quadrature points that are used to compute the corresponding integrals in the variational formulation.
	
	\item This setting can benefit from domain decomposition into many sub-domains, where in each sub-domain we can use a separate number of test functions based on the local regularity of the solution. This yields a local (elemental) more flexible learning approach.
	
\end{itemize}

The rest of the paper is organized as follows. In section \ref{Sec: fcn appx}, we briefly discuss the nonlinear function approximation of DNNs. In section \ref{Sec: vPINN}, we formulate the proposed VPINN, and then carry out the analytical approach in the derivation of variational residuals for shallow networks in section \ref{Sec: vPINN shallow}. We take the formulation to deeper networks in section \ref{Sec: vPINN Deep}. Each section is supported by several numerical examples to demonstrate the performance of the proposed method. We conclude the paper with a summary section \ref{Sec: sum}.

%%
%\newpage
%
%%%%%%%%%%%%%%%%%%%%%%%%%%%
\section{Nonlinear Function Approximation}
\label{Sec: fcn appx}
%%%%%%%%%%%%%%%%%%%%%%%%%%%
% 
In general, a nonlinear approximation identifies the optimal approximant as a linear composition of some nonlinear approximators. Let $\textbf{x} \in \Omega \subset \mathbb{R}^d $ and $u: \Omega \rightarrow \mathbb{R}$ be a target function in a Hilbert space associated with some proper norm $\Vert \cdot  \Vert_{\star} $ such as $L^2$ or $L^{\infty}$ norm. A DNN, comprised of $L$ hidden layers with $N_i$ neurons in each layer and activation function $\sigma$, is a powerful nonlinear approximation that takes the compositional form
\begin{align}
\label{Eq: deep NN}
u(x) \approx \tilde{u}(\textbf{x}) = l \circ T^{(L)} \circ T^{(L-1)} \circ \cdots \circ T^{(1)}(\textbf{x}) .
\end{align} 
The nonlinear mapping in each hidden layer $i=1,2,\cdots,L$ is $T^{(i)}(\cdot) = \sigma(\textbf{W}_i \times \cdot + \textbf{b}_i)$ with weights $\textbf{W}_i \in \mathbb{R}^{N_i \times N_{i-1}}$ and biases $\textbf{b}_i \in \mathbb{R}^{N_i}$, where $N_0 = d$ is the input dimension. The output of the last hidden layer is finally mapped via the liner mapping $l: \mathbb{R}^{N_L} \rightarrow \mathbb{R}$ to the network output. We note that the activation function $\sigma$ has similar form for each neuron, however, it may also have different domain and image dimensionality based on the structure of network \cite{jagtap2019locallyadaptive, jagtap2019adaptiveinpress}.

For a specific network structure (depth $L$ and width $N$) and a choice of activation function, DNN \eqref{Eq: deep NN} seeks an approximation that minimizes an objective function that measures the discrepancy of NN output and the given target function, i.e. $\Vert u(\textbf{x}) - \tilde{u}(\textbf{x}) \Vert_{\star}$. Therefore, the approximated function is obtained as $u^{*}(\textbf{x}) = \underset{\textbf{W, \textbf{b}}}{\text{arg min}} \,  \Vert u(\textbf{x}) - \tilde{u}(\textbf{x}) \Vert_{\star}$. This minimization problem, however, is not trivial as usually DNNs take high dimensional inputs and have deep structures that can lead to some numerical issues such as local minima traps, and in practice, one finds  $\hat{u}(\textbf{x}) \approx u^{*}(\textbf{x})$ as the approximation. Hence, the accuracy of DNNs can be characterized by dividing the expected error into three main types: approximation error $\Vert \tilde{u} - u \Vert_{\star}$, generalization error $\Vert u^{*} - \tilde{u} \Vert_{\star}$, and optimization error $\Vert \hat{u} - u^{*} \Vert_{\star}$. There have been several works in the literature \cite{jin2019quantifying, daubechies2019nonlinear, mhaskar2016deep, suzuki2018adaptivity, ma2019barron, shen2019nonlinear}, which attempt to identify these different sources of error and develop a proper framework for error analysis of deep NN. Achieving an optimal approximation, however, depends on many factors and it still remains an open problem.

%%
%\newpage
%
%%%%%%%%%%%%%%%%%%%%%%%%%%%
\section{Variational Physics-Informed Neural Networks (VPINNs)}
\label{Sec: vPINN}
%%%%%%%%%%%%%%%%%%%%%%%%%%%
%
In general, the underlying governing equation of a physical problem in steady state can be written as
\begin{align}
\label{Eq: Strong form }
\mathcal{L}^\textbf{q} u(\textbf{x}) & = f(\textbf{x}), \quad \textbf{x}\in \Omega 
\\
\label{Eq: Strong form BC}
u(\textbf{x}) & = h(\textbf{x}), \quad \textbf{x}\in \partial \Omega
\end{align}
over the physical domain $\Omega \subset \mathbb{R}^d$ with dimensionality $d$ and boundaries $\partial \Omega$. The quantity $u(\textbf{x}): \Omega \rightarrow \mathbb{R}$ describes the underlying physics, the forcing $f(\textbf{x})$ is some (sufficiently) known external excitation, and $\mathcal{L}$ usually contains differential and/or integro-differential operators with parameters $\textbf{q}$. A proper numerical method obtains the approximate solution $\tilde u (\textbf{x}) \approx u (\textbf{x}) $ to the above set of equations, while satisfying the boundary/initial conditions is a critical key in the stability and convergence of the method. 

\vspace{0.1 in}
\noindent $\bullet$ \textbf{PINNs}:
A solution of equations \eqref{Eq: Strong form } and \eqref{Eq: Strong form BC} can be obtained by employing a DNN as approximated solution, i.e. $u(\textbf{x}) \approx \tilde u (\textbf{x}) = u_{NN}(\textbf{x}; \textbf{w} , \textbf{b})$, where $u_{NN}$ is a neural network output with weights and biases $\{\textbf{w} , \textbf{b} \}$, respectively. This results in a physics-informed neural network (PINN), which was first introduced in \cite{raissi2019physics}. The PINN algorithm infuses the governing equation to the network by forcing the network output to satisfy the corresponding mathematical model in the interior domain and at boundaries. Let $\textbf{x}_r \in \Omega$ and $\textbf{x}_u \in \partial \Omega$ be some interior and boundary points, respectively. We define the \textit{strong-form residual} as
\begin{align}
\label{Eq: residue strong form}
residual^{\mathfrak{s}} 
&= r(\textbf{x}) - r^b(\textbf{x}),
\\ \nonumber 
r(\textbf{x}) &
= \mathcal{L}^\textbf{q} u_{NN}(\textbf{x}) - f(\textbf{x}), \quad \forall \textbf{x} \in \textbf{x}_r,
\\ \nonumber
r^b(\textbf{x}) 
&= u_{NN}(\textbf{x}) - h(\textbf{x}) , \,\,\quad\quad \forall \textbf{x} \in \textbf{x}_u.
\end{align}
Subsequently, we define the \emph{strong-form loss function} as 
\begin{align}
\label{Eq: loss PINN - 1}
L^{\mathfrak{s}} 
& = L_{r}^{\mathfrak{s}} + L_{u} ,
\\ \nonumber
L_{r}^{\mathfrak{s}}
& 
= \frac{1}{N_r} \sum_{i = 1}^{N_r} \Big|r(\textbf{x}_{r_i})\Big|^2 ,
%= \frac{1}{N_r} \,\, \Big\Vert \, r(\textbf{x}_r) \, \Big\Vert^{2}_{l^2}
\quad
L_{u}
= \tau \, \frac{1}{N_u} \sum_{i = 1}^{N_u} \Big|r^b(\textbf{x}_{u_i}) \Big|^2,
%= \frac{1}{N_r} \,\, \Big\Vert \, r(\textbf{x}_r) \, \Big\Vert^{2}_{l^2}
%\,\, +\, \tau \,\, \frac{1}{N_u} \,\,  \Big\Vert \,  u_{NN}(\textbf{x}_u) - h(\textbf{x}_u) \, \Big\Vert^{2}_{l^2} ,
\end{align}
in which $\tau $ denotes a penalty parameter. Here, we use the superscript $\mathfrak{s}$ to refer to the loss function associated with the strong form of residual and also later to distinguish between this from and other considered cases. In this setting, we pose the problem of solving \eqref{Eq: Strong form } and \eqref{Eq: Strong form BC} as: 
\begin{align}
\text{find } \tilde u(\textbf{x}) = u_{NN}(\textbf{x}; \textbf{w}^*, \textbf{b}^*) \text{ such that } \{\textbf{w}^*, \textbf{b}^*\} = \text{argmin}(L^{\mathfrak{s}}(\textbf{w}, \textbf{b})).
\end{align}

\begin{SCfigure}
	\centering
	\includegraphics[width=0.5\linewidth]{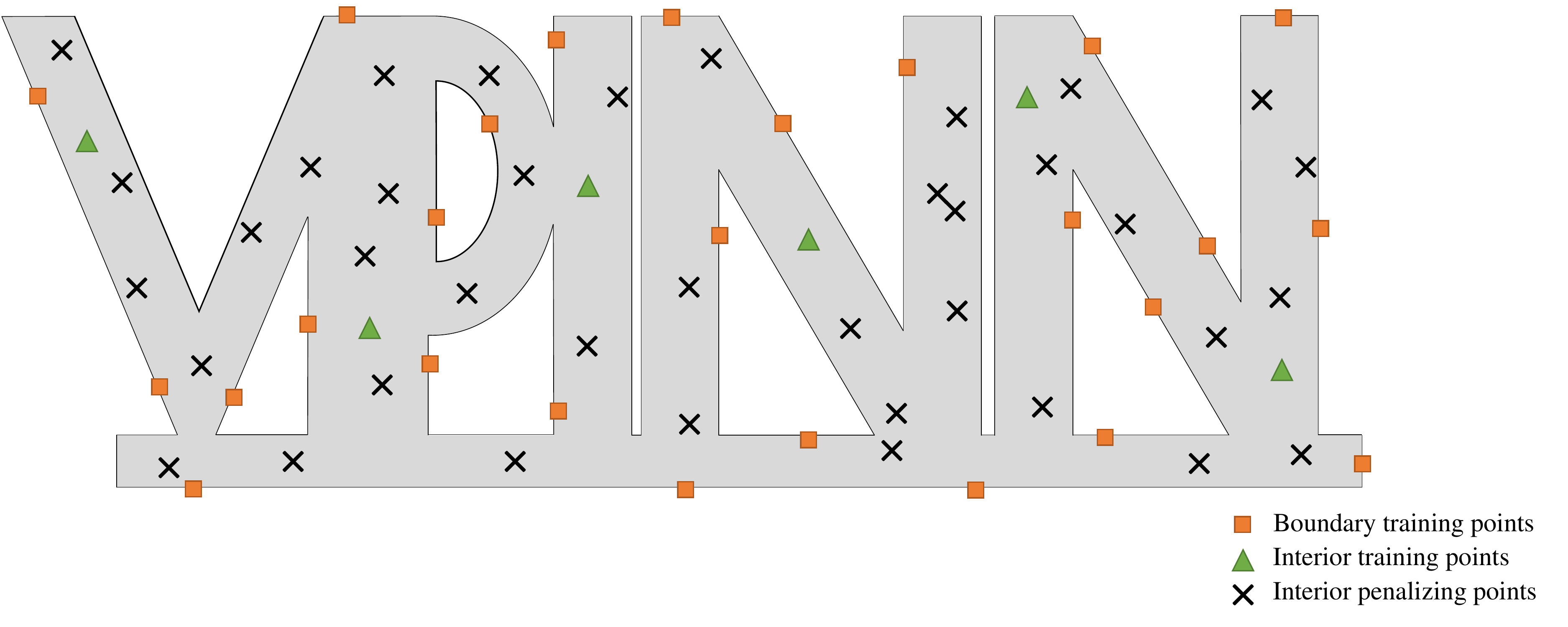}	
	\caption{\scriptsize A schematic of training and penalizing points on a complicated domain in evaluation of \emph{strong-form loss function} $L^{\mathfrak{s}}(\textbf{w} , \textbf{b})$. The squares are training points scattered around on the domain boundary and triangles are a few interior training points. The crosses are penalizing points, which are largely available in millions as mini batch sizes can be used. We note that the training points can include some interior points as well, which can efficiently add the known information from the interior domain to the training algorithm; see \cite{raissi2019physics} for more details.}
	\label{Fig: TR PE Pts}
\end{SCfigure}

The approximated solution $u_{NN} $ in PINNs does not necessarily satisfy the boundary conditions. Therefore, we see that the strong-form loss function is comprised of two parts: the first part that penalizes the strong-form residual at some interior (penalizing) points $\textbf{x}_r$, and the second part that penalizes the solution at boundary (training) points $\textbf{x}_u$, leading to a penalty-like method. Fig. \ref{Fig: TR PE Pts} schematically shows these points in an arbitrary domain. Several spectral penalty methods have been developed in the literature, see e.g. \cite{don1994chebyshev, hesthaven2000spectral, wang2019spectral}. In general, these penalty methods construct a unified equation by efficiently adding the boundary conditions to the interior governing equation via a penalty term. This term includes a penalty function $Q(\textbf{x})$ and a penalty parameter $\tau$ that are designed such that the corresponding scheme maintains the coercivity in proper function spaces and norms. In the context of spectral methods, a penalty term is required when the basis functions do not naturally satisfy the boundary conditions, and in polynomial pseudospectral (or collocation) methods $Q(\textbf{x})$ takes similar polynomial form as the interpolation polynomials \cite{don1994chebyshev}. There is no  analysis in the literature that formulates PINNs in the form of a penalty method. Thus, the penalty parameter $\tau$ in the strong-form loss function depends on the problem at hand and designed based on numerical experiments but it can also be set as a hyperparameter.

%
%*************************************************************************************
%
%
%*************************************************************************************
%

\vspace{0.1 in}

\noindent $\bullet$ \textbf{VPINNs}:
Here, we develop the new setting of variational physics-informed neural network, which is inspired by the variational form of the mathematical problem. We let the approximation form $ \tilde u (\textbf{x}) = u_{NN}(\textbf{x}; \textbf{w} , \textbf{b})$ with weights and biases $\{\textbf{w} , \textbf{b} \}$, respectively. We also let $v(\textbf{x})$ be a properly chosen test function. We multiply \eqref{Eq: Strong form } by the test function and integrate over the whole domain to obtain the variational form
\begin{align}
\label{Eq: Variational form}
&
\left( \mathcal{L}^\textbf{q} u_{NN}(\textbf{x}), v(\textbf{x}) \right)_{\Omega} = \left( f(\textbf{x}) , v(\textbf{x}) \right)_{\Omega} 
\\
\label{Eq: Variational form BC}
& u(\textbf{x}) = h(\textbf{x}), \quad \textbf{x}\in \partial \Omega
\end{align}
where $(\cdot , \cdot)$ denotes the usual inner product. This variational form leads to the \emph{variational residual}, defined as  
\begin{align}
\label{Eq: residue var form}
& Residual^{\mathfrak{v}} 
= \mathcal{R} - F - r^b,
\\ \nonumber 
& \mathcal{R} 
= \left( \mathcal{L}^\textbf{q} u_{NN} , v \right)_{\Omega},
\quad
F = \left( f, v \right)_{\Omega},
\end{align}
which is enforced for any admissible test function $v_k , \,\, k = 1,2,\cdots$ and $r^b$ has the same form as in \eqref{Eq: residue strong form}. Hence, we construct a discrete finite dimensional space $V_K $ by choosing a finite set of admissible test functions and let $$V_K = \text{span} \lbrace v_k , \, k=1, 2, \cdots, K \rbrace. $$ Subsequently, we define the \emph{variational loss function} as
\begin{align}
\label{Eq: loss weak - 1}
&
L^{\mathfrak{v}} = L_{R}^{\mathfrak{v}} + L_{u} ,
\\ \nonumber
&
L_{R}^{\mathfrak{v}} = \frac{1}{K} \sum_{k = 1}^{K} \Big| \mathcal{R}_k - F_k \Big|^2 ,
\quad
L_{u} = \tau \frac{1}{N_u} \sum_{i = 1}^{N_u} \Big|r^b(\textbf{x}_{u_i}) \Big|^2,
\end{align}
in which $\tau $ denotes the penalty parameter. Here, we use the superscript $\mathfrak{v}$ to refer to the loss function associated with the variational form of residual. In this setting, we pose the problem of solving \eqref{Eq: Variational form} and \eqref{Eq: Variational form BC} as: 
\begin{align}
\text{find } \tilde u(\textbf{x}) = u_{NN}(\textbf{x}; \textbf{w}^*, \textbf{b}^*) \text{ such that } \{\textbf{w}^*, \textbf{b}^*\} = \text{argmin}(L^{\mathfrak{v}}(\textbf{w}, \textbf{b})). 
\end{align}

\begin{rem}[Analogy between strong-form and variational residual]
A similar analogy as in standard numerical schemes to solve the strong and variational (weak) forms of differential equations can be constructed here too. Loosely speaking, by adopting the delta dirac function as the test function, i.e. $v_k(\textbf{x}) = \delta(\textbf{x} - \textbf{x}_k)$, the variational residual \eqref{Eq: residue var form} becomes the strong-form residual, given in \eqref{Eq: residue strong form} evaluated at (penalizing point) $\textbf{x}_k$:
\begin{align*}
\mathcal{R}_k - F_k
& = \left( \mathcal{L}^\textbf{q} u_{NN}(\textbf{x}) , v_k(\textbf{x}) \right)_{\Omega} - \left( f(\textbf{x}), v_k(\textbf{x}) \right)_{\Omega}
\\ \nonumber
& = \left( \mathcal{L}^\textbf{q} u_{NN}(\textbf{x}) , \delta(\textbf{x} - \textbf{x}_k) \right)_{\Omega} - \left( f(\textbf{x}), \delta(\textbf{x} - \textbf{x}_k) \right)_{\Omega}
\\ \nonumber
& = \mathcal{L}^\textbf{q} u_{NN}(\textbf{x}_k) -  f(\textbf{x}_k)
\\ \nonumber
& = r(\textbf{x}_k).
\end{align*}
Therefore, in the strong-form, we evaluate the residual at some penalizing points while there are no such points in the variational form and instead we test the residual with a set of test functions.

\end{rem}

A major challenge in VPINN, unlike PINN, is to compute the integrals in the loss function. The compositional structure of neural networks makes it almost impossible to obtain analytic expression of these integrals. Their computation is also not fully practically feasible as even in the case of network with two hidden layer there is no analysis for their quadrature rules. In order to avoid such a complication, we first choose a shallow network with one hidden layer. This enables us to carry out the derivations analytically at least for some special cases. Then, we will introduce deep VPINNs by composing several hidden layers and considering deep networks, where we need to employ numerical integration techniques to compute the integrals in the variational residuals.

%******************************************************************************************
\begin{figure}[t]
	\center
	\includegraphics[width=0.95\linewidth]{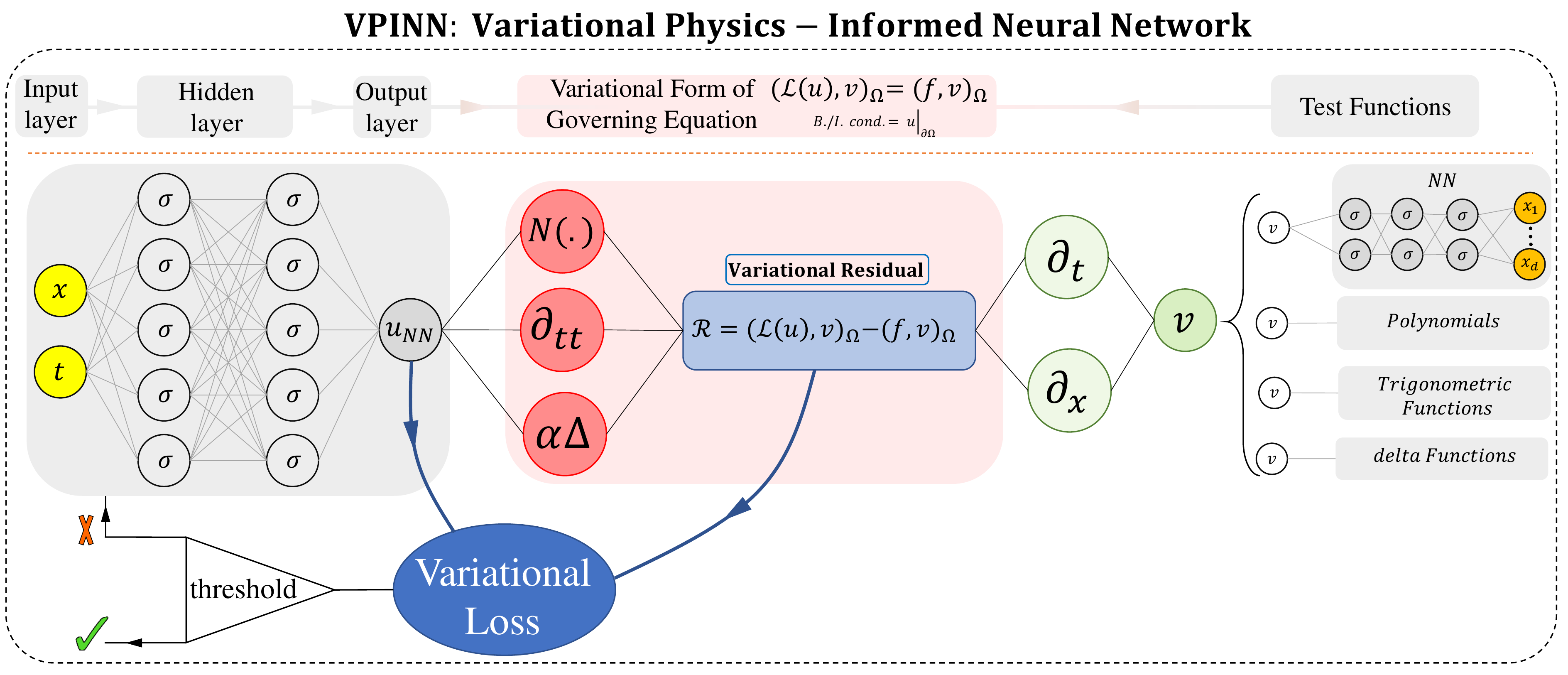}
	\caption{\scriptsize \label{Fig: VPINN} Schematic of VPINN in a Petrov-Galerkin formulation. Trial functions belong to the space of NN and test functions can be chosen from a separate NN or other function spaces such as polynomials and trigonometric functions. Red color represents the differential operators on trial space. Green color represents the test functions and their derivatives. Blue color represents the variational residuals $\mathcal{R}$. }
\end{figure}
% 

%
%%%%%%%%%%%%%%%%%%%%%%%%%%%
\section{Shallow VPINNs}
\label{Sec: vPINN shallow}
%%%%%%%%%%%%%%%%%%%%%%%%%%%
%
We consider a shallow neural network with one hidden layer and $N$ neurons. We let $u(x): \Omega \rightarrow \mathbb{R}$, where $\Omega = (-1,1)$, and consider the following boundary value problem
\begin{align}
\label{Eq: 1-d BVP}
-\frac{d^2 u(x)}{d x^2} & = f(x), \quad x\in(-1,1)
\\ 
\label{Eq: 1-d BVP BC}
u(-1) &= g,  
\\ \nonumber
u(1) & = h, 
\end{align}
where $h$ and $g$ are constants and we assume the force term $f(x)$ is fully available (this assumption can be relaxed as we do not need the force term to be available at all points). Let the approximate solution be $u(x) \approx \tilde u(x) = u_{NN}(x)$, then the strong-form residual \eqref{Eq: residue strong form} becomes  
\begin{align}
\label{Eq: 1-d BVP residue}
residual^{\mathfrak{s}} 
&= r(x) - r^b(x)
\\ \nonumber 
r(x) &= -\frac{d^2 u_{NN}(x)}{d x^2} - f(x), \quad  x\in(-1,1) ,
\\ \nonumber
r^b(x) &=  u_{NN}(x) - u(x), \quad\quad\quad x = \pm 1.
\end{align}
We choose a set of test functions $v_k(x) \in V_K$ with compact support on $\Omega$ such that 
\begin{align}
\label{Eq: test function general - 1}
v_k(x) = 
\begin{cases}
\text{non-zero}, \quad & x\in (-1,1),
\\
0, \quad & \text{else where},
\end{cases}
\quad k=1,2,\cdots,K.
\end{align}
The variational residual \eqref{Eq: residue var form} then becomes
\begin{align}
\label{Eq: 1-d BVP var residue - 0}
Residual^{\mathfrak{v}}_k 
&= \mathcal{R}_k - F_k - r^b
\\ \nonumber 
\mathcal{R}_k & = - \left( \frac{d^2 u_{NN}(x)}{d x^2}, v_k(x) \right)_{\Omega} ,
\quad 
F_k = \left( f(x) , v_k(x)  \right)_{\Omega} ,
%\\ \nonumber
%\mathcal{R}^b_k &= \tau^{-} \left( Q^{-}(x), v_k(x) \right)_{\Omega} (u_{NN}(-1) - g) + \tau^{+} \left( Q^{+}(x), v_k(x) \right)_{\Omega} (u_{NN}(1) - h)
%
\end{align}
where $r^b$ is given in \eqref{Eq: 1-d BVP residue}. By integrating by parts in the first term $\mathcal{R}_k$, we can define three distinctive \emph{variational residual} forms, namely 
\begin{align}
\label{Eq: 1-d BVP var residue - 1}
\mathcal{R}^{(1)}_k &= - \left( \frac{d^2 u_{NN}(x)}{d x^2}, v_k(x) \right)_{\Omega} ,
\\
\label{Eq: 1-d BVP var residue - 2}
\mathcal{R}^{(2)}_k &= \,\,\,\,\, \left( \frac{d u_{NN}(x)}{d x}, \frac{d v_k(x)}{d x} \right)_{\Omega} 
- \frac{d u_{NN}(x)}{d x} \, v_k(x) \bigg\vert_{\partial \Omega},
\\
\label{Eq: 1-d BVP var residue - 3}
\mathcal{R}^{(3)}_k &= - \left( u_{NN}(x) , \frac{d^2 v_k(x)}{d x^2} \right)_{\Omega} 
\, - \frac{d u_{NN}(x)}{d x} \, v_k(x) \bigg\vert_{\partial \Omega} 
+ u_{NN}(x) \, \frac{d v_k(x)}{d x} \bigg\vert_{\partial \Omega} .
\end{align}
The corresponding \emph{variational loss function}s for each case take the form
\begin{align}
\label{Eq: loss VPINN}
&L^{\mathfrak{v}(i)} 
= L_{R}^{\mathfrak{v}(i)} + L_{u}, \quad i=1,2,3,
\\ \nonumber
&L_{R}^{\mathfrak{v}(i)} = \frac{1}{K} \sum_{k = 1}^{K} \Big| \mathcal{R}^{(i)}_k - F_k \Big|^2 ,
\quad L_{u} = \frac{\tau}{2} \left( \Big|u_{NN}(-1) - g \Big|^2 + \Big|u_{NN}(1) - h \Big|^2 \right),
\end{align}

\begin{rem}
\label{Rem: VR3}
Because the test functions have compact support over $\Omega$, the first boundary term in \eqref{Eq: 1-d BVP var residue - 2} and \eqref{Eq: 1-d BVP var residue - 3} vanishes. Moreover, by choosing a proper weight coefficient $\tau$ in \eqref{Eq: loss VPINN}, we make sure that the network learns the boundary accurately. Therefore, we replace the second boundary term in \eqref{Eq: 1-d BVP var residue - 3} by the exact values at boundaries. Hence, we have $$\frac{d u_{NN}(x)}{d x} \, v_k(x) \bigg\vert_{\partial \Omega} = 0, \quad u_{NN}(x) \, \frac{d v_k(x)}{d x} \bigg\vert_{\partial \Omega} \approx u(x) \, \frac{d v_k(x)}{d x} \bigg\vert_{\partial \Omega}.$$
\end{rem}

%
%%%%%%%%%%%%%%%%%%%%%%%
\subsection{Shallow VPINN With Sine Activation Functions And Sine Test Functions}
\begin{SCfigure}
	\centering
	\includegraphics[width=0.35\linewidth]{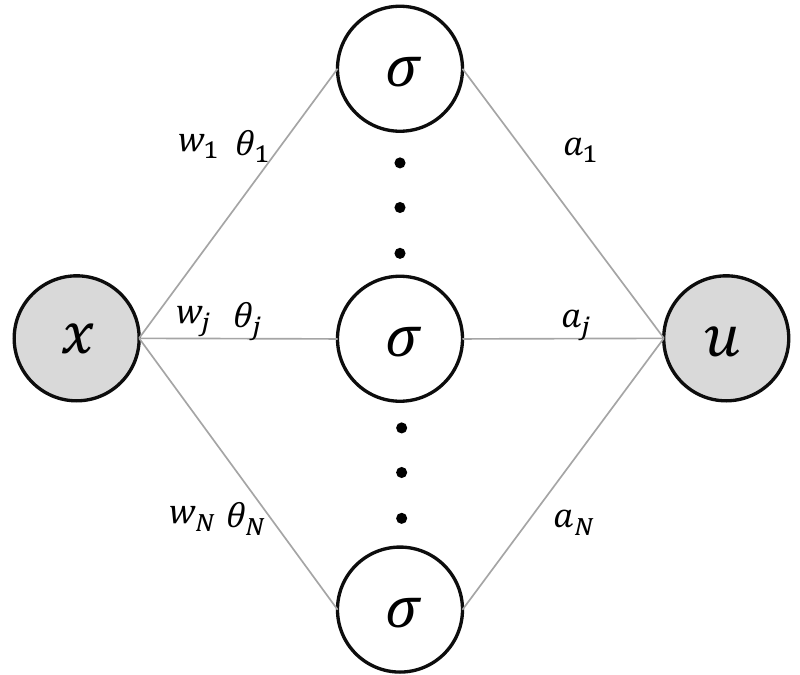}
	\caption{\scriptsize \label{Fig: Shallow NN} A shallow neural network with one hidden layer, $N$ neurons, and activation function $\sigma$. We let $u_{NN_j}(x) = a_j \, \sigma \left(w_j \, x + \theta_j  \right)$ and $u_{NN}(x) = \sum_{j=1}^{N} u_{NN_j}(x)$. }
\end{SCfigure}

%%%%%%%%%%%%%%%%%%%%%%%
%
A shallow network with one hidden layer has a relatively simple representation, which enables us to carry out the integrals analytically. 
Denoting the weights and biases associated with the first layer by $w_j$ and $\theta_j$, respectively, and the weights associated with the output by $a_j$, we construct the neural network with $N$ neurons and represent the approximate solution as
\begin{align}
\label{Eq: shallow NN sine}
u_{NN}(x) = \sum_{j=1}^{N} u_{NN_j}(x) = \sum_{j=1}^{N} a_j \, \sin \left(w_j \, x + \theta_j  \right),
\end{align}   
where $\sigma = \sin$ is the activation function. Figure \ref{Fig: Shallow NN} shows the schematic of the network.
%
%******************************************************************************************
We also choose the test functions to be sine functions of the form
\begin{align}
\label{Eq: sine test functions}
v_k (x) = \sin(k \, \pi \, x  ), \quad k=1,2,\cdots, K.  
\end{align}
The derivation of variational residuals is not complicated but requires careful use of several trigonometric identities. We present the full derivations in the Appendix \ref{Sec: Appx shallow sine} and only present the final results here. The variational residuals \eqref{Eq: 1-d BVP var residue - 1}-\eqref{Eq: 1-d BVP var residue - 3} become
\begin{align}
\label{Eq: 1-d BVP var residue - 2 - shallow sine}
\mathcal{R}^{(1)}_k = \mathcal{R}^{(2)}_k & = 2(-1)^k \, k \, \pi \, \sum_{j=1}^{N} \frac{a_j \, w_j^2 \, \cos(\theta_j) \sin(w_j)}{w_j^2 - k^2 \pi^2}
,
\\
\label{Eq: 1-d BVP var residue - 3 - shallow sine}
\mathcal{R}^{(3)}_k &= 2 (-1)^k \, k \, \pi \,  \sum_{j=1}^{N} \frac{a_j \, k^2 \, \pi^2 \, \cos(\theta_j) \sin(w_j)}{w_j^2 - k^2 \pi^2} + 
(-1)^k \, k \, \pi \, (h-g)
.
\end{align}
We note that the variational forms \eqref{Eq: 1-d BVP var residue - 1} and \eqref{Eq: 1-d BVP var residue - 2} have exact similar analytical expression, however, the form \eqref{Eq: 1-d BVP var residue - 3} is different due to the additional boundary approximation explained in Remark \ref{Rem: VR3}.

\vspace{0.1 in}
\noindent $\bullet$ \textbf{Steady Burger's Equation:}
The Burger's equation is one of the fundamental PDEs arising in various physical fields, including nonlinear acoustics, gas dynamics, and fluid mechanics; see e.g. \cite{whitham2011linear} and references therein. This equation was first introduced in \cite{bateman1915some} and then later in the context of theory of turbulence was studied in \cite{burgers1948mathematical}. Here, we study the one-dimensional steady state Burger's equation, given as
\begin{align}
\label{Eq: steady Burger}
&u \frac{d u}{d x} - \frac{d^2 u}{d x^2} = f(x),
\\ \nonumber
&u(-1) = g, \,\, u(1)=h.
\end{align}
The variational residual \eqref{Eq: residue var form} for this problem becomes
\begin{align}
\label{Eq: St Burger var residue - 0}
Residual^{\mathfrak{v}}_k = \mathcal{R}_k + \mathcal{R}^{NL}_k - F_k - r^b,
\end{align}
in which 
\begin{align}
\label{Eq: St Burger var residue - 1}
\mathcal{R}^{NL}_k = \left( u_{NN}(x)\frac{d u_{NN}(x)}{d x}, v_k(x) \right)_{\Omega} ,
\end{align}
and other terms are the same as in \eqref{Eq: 1-d BVP var residue - 0}. The corresponding \emph{variational loss function}s for this problem can be written as
\begin{align}
\label{Eq: loss VPINN St Burger}
L^{\mathfrak{v}(i)} 
= L_{R}^{\mathfrak{v}(i)} + L_{u}, \quad 
L_{R}^{\mathfrak{v}(i)} 
 = \frac{1}{K} \sum_{k = 1}^{K} \Big| \mathcal{R}^{(i)}_k + \mathcal{R}^{NL}_k - F_k \Big|^2, 
%+ \frac{\tau}{2} \left( \Big|u_{NN}(-1) - g \Big|^2 + \Big|u_{NN}(1) - h \Big|^2 \right),
\quad i=1,2,3.
\end{align}
The shallow network \eqref{Eq: shallow NN sine} with test functions \eqref{Eq: sine test functions} gives 
\begin{align}
\label{Eq: St Burgers var residue - shallow sine}
\mathcal{R}^{NL}_k & = (-1)^k \, k \, \pi \, \sum_{i=1}^{N}\sum_{j=1}^{N} a_i a_j w_i
\left[  
\frac{\sin(w_j + w_i) \cos(\theta_j + \theta_i) }{(w_j + w_i)^2 - k^2 \pi^2}
+\frac{\sin(w_j - w_i) \cos(\theta_j - \theta_i) }{(w_j - w_i)^2 - k^2 \pi^2}
\right].
\end{align}
The detailed derivation is given in the appendix.

\begin{figure}[h]
	\center
	\includegraphics[clip, trim=1cm 0cm 0cm 0cm, width=0.48\linewidth]{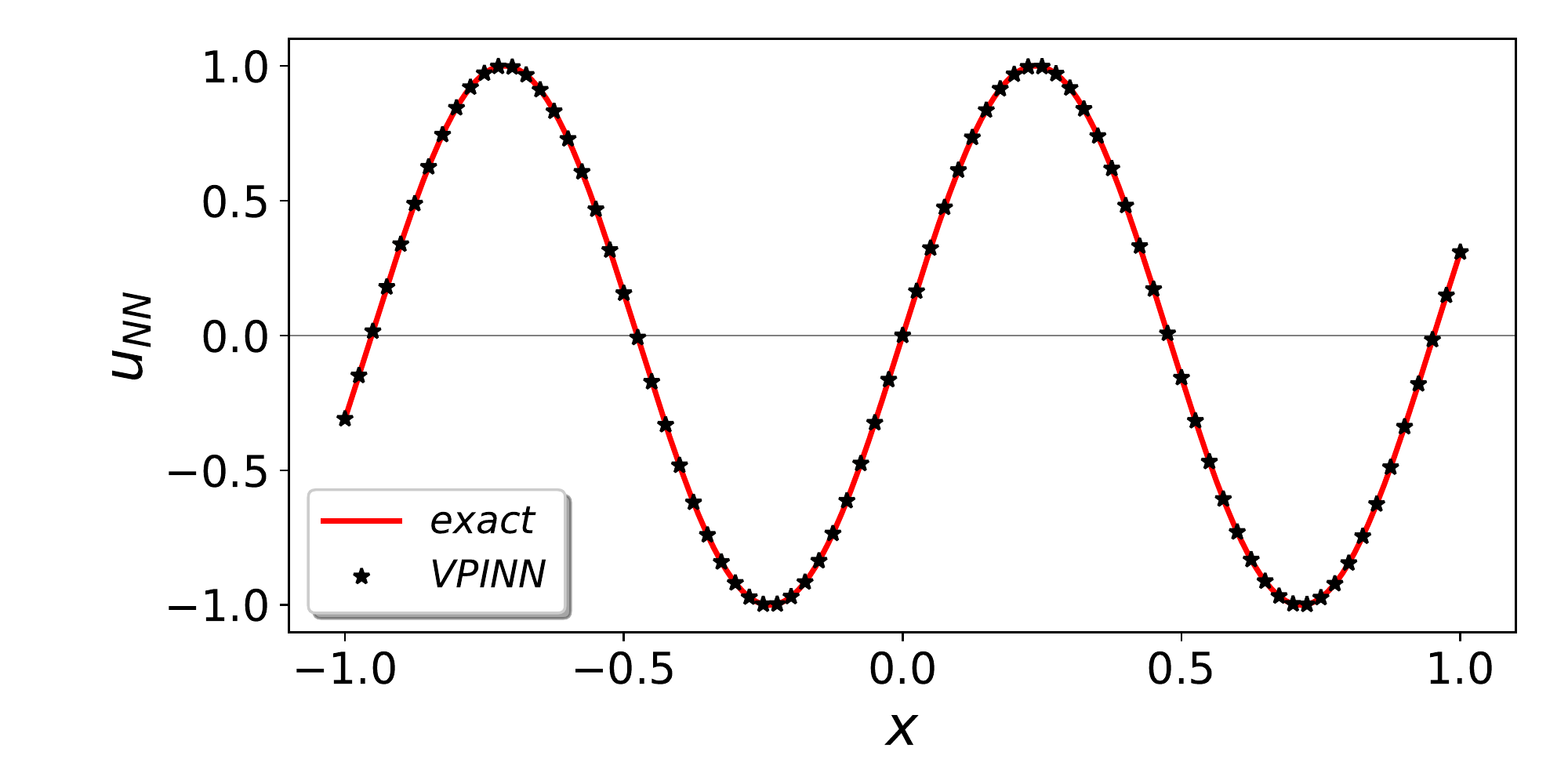}
	\includegraphics[clip, trim=1cm 0cm 0cm 0cm, width=0.48\linewidth]{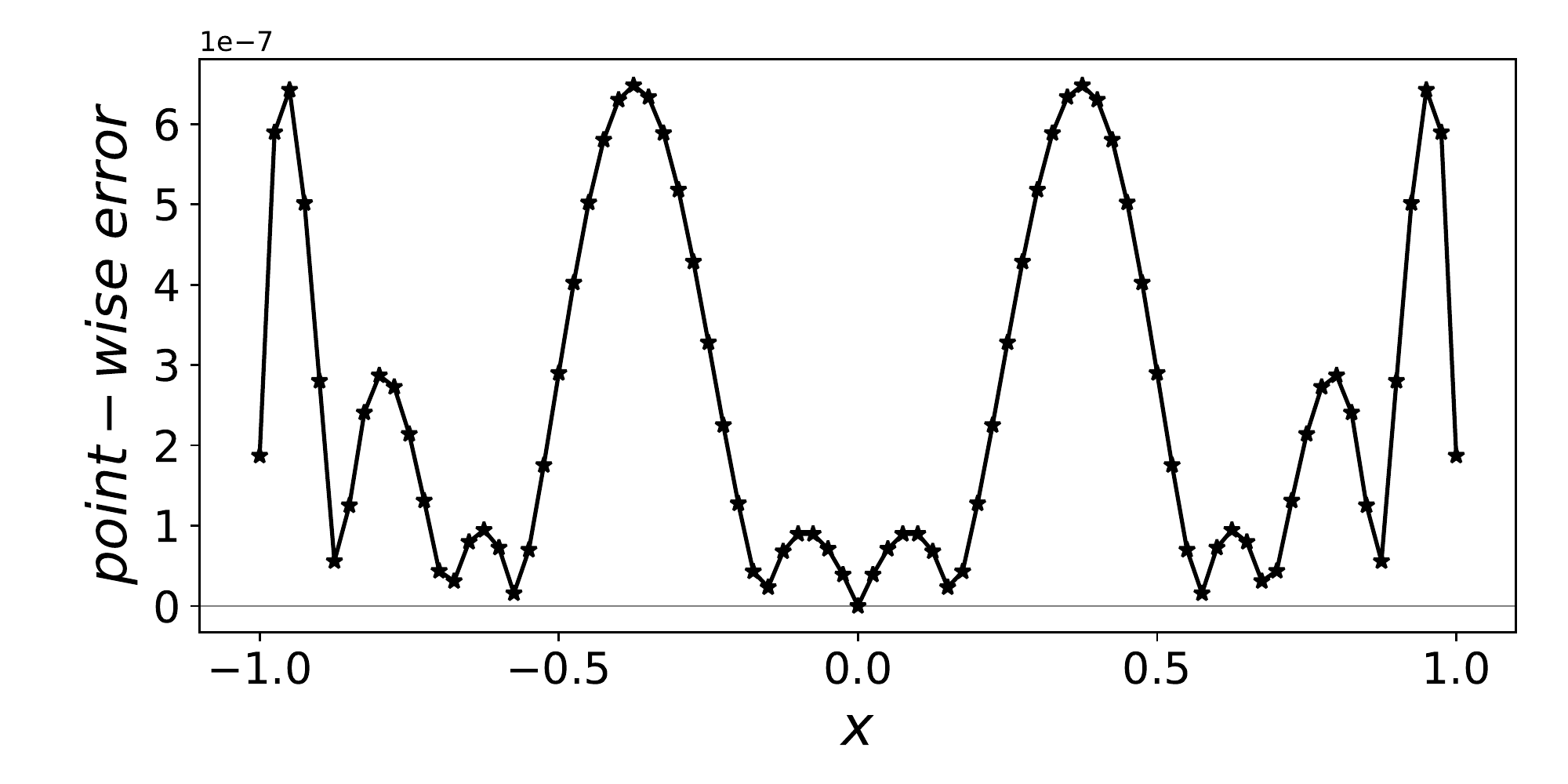}
	\vspace{-0.2 in}
	\caption{\scriptsize \label{Fig: St Burger} One-dimensional steady state Burger's equation: VPINN with $\mathcal{R}^{(1)} = \mathcal{R}^{(2)}$ formulation. Left: exact solution $\sin(2.1 \pi x)$ and VPINN approximation. Right: point-wise error averaged over several random network initializations. See Table \ref{Table: VPINN para burger} for VPINN hyperparameters.}
\end{figure}

\begin{figure}[h]
	\center
	\includegraphics[clip, trim=1cm 0cm 0cm 0cm, width=0.48\linewidth]{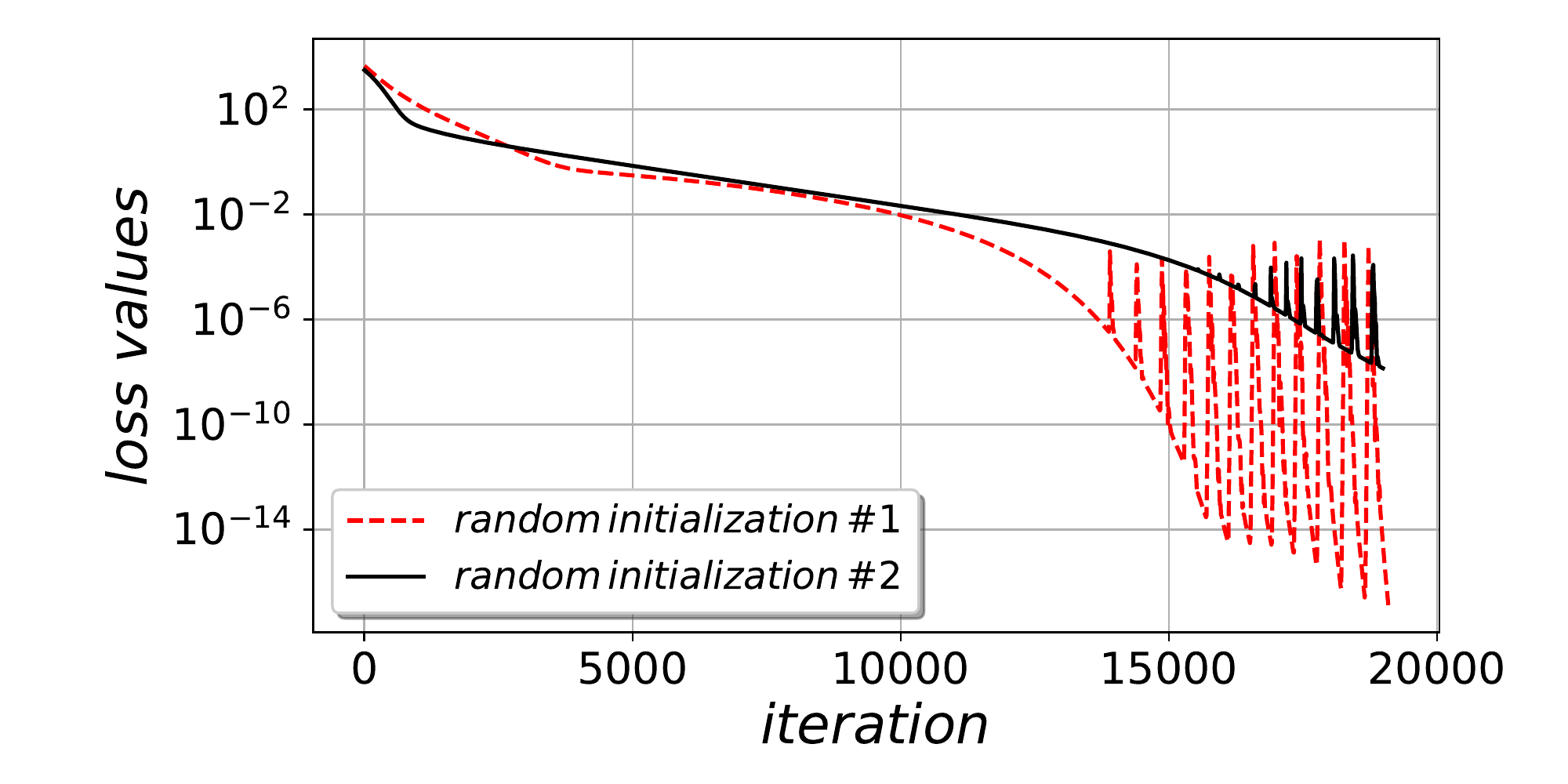}
	\includegraphics[clip, trim=1cm 0cm 0cm 0cm, width=0.48\linewidth]{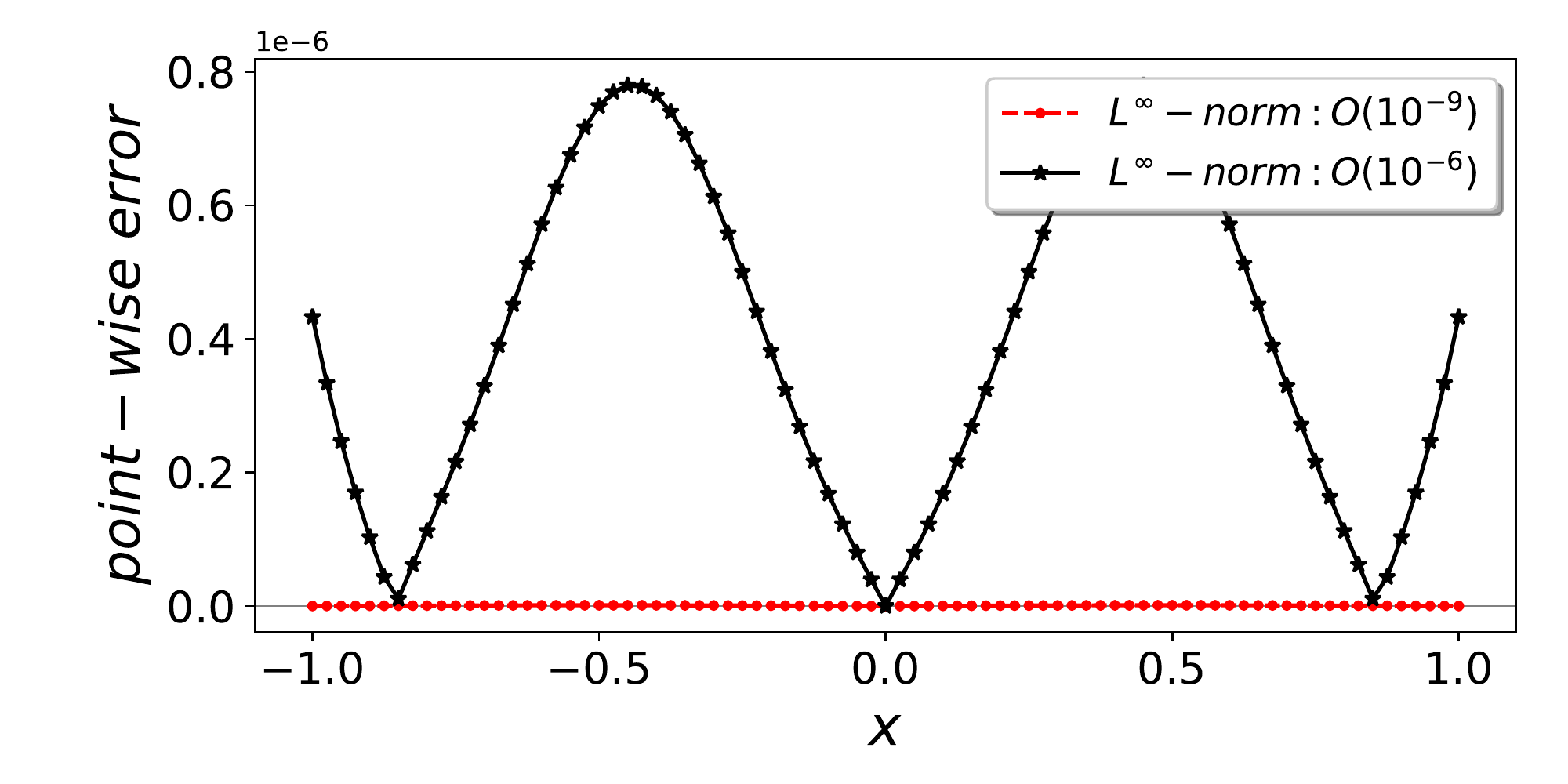}
	\vspace{-0.2 in}
	\caption{\scriptsize \label{Fig: St Burger compare} One-dimensional steady state Burger's equation: network initialization effect on optimization performance. Left: comparison of loss values. Right: comparison of point-wise error. The red dashed line shows a more successful optimization and thus a much lower error. See Table \ref{Table: VPINN para burger} for VPINN hyperparameters.}
\end{figure}

\begin{figure}[h]
	\center
	\begin{tabular}{l c}
		\includegraphics[clip, trim=1.1cm 0cm 0.2cm 0cm, width=0.46\linewidth]{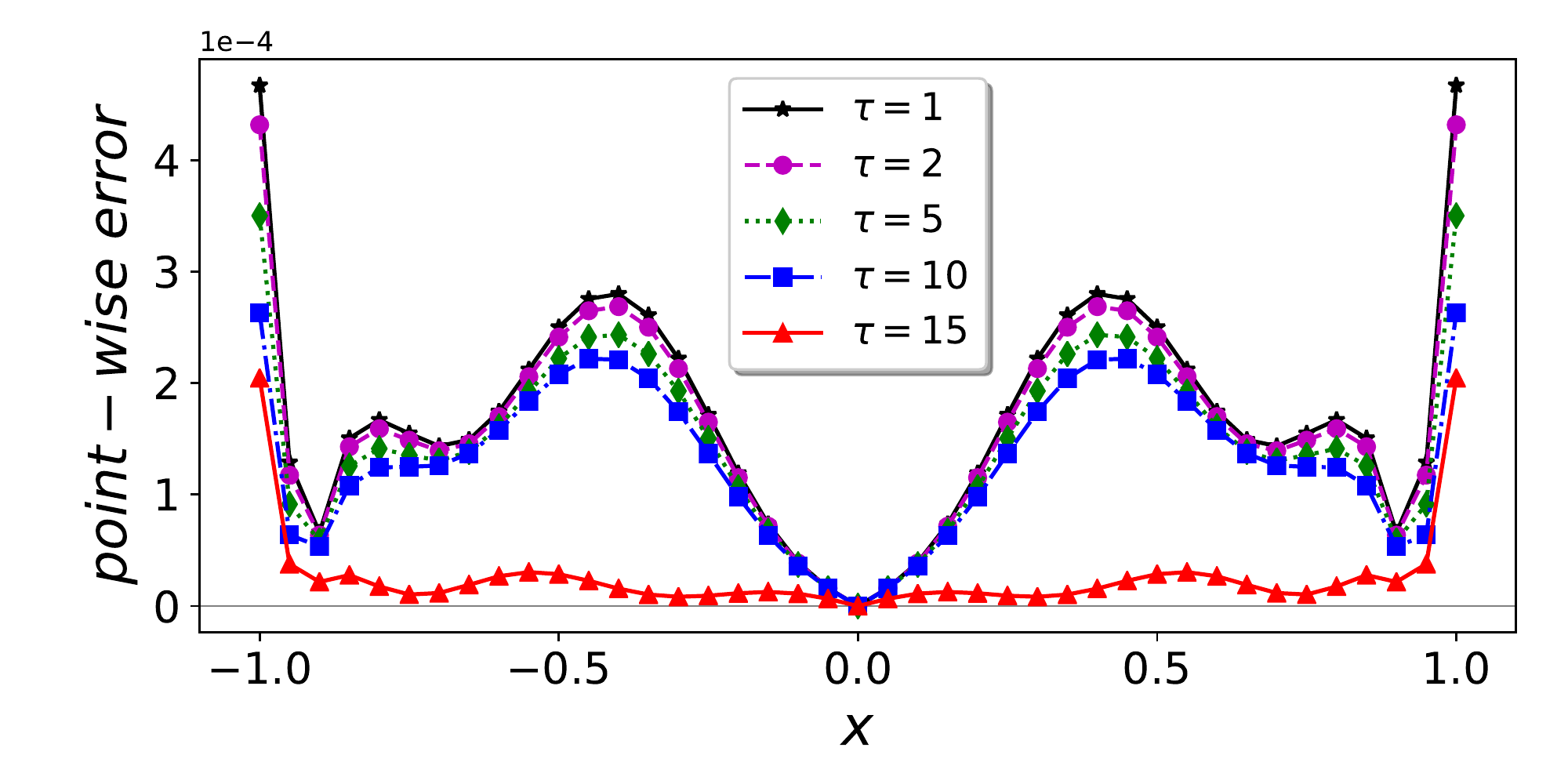}
		&
		\includegraphics[clip, trim=1.2cm 0cm 0.5cm 0cm, width=0.45\linewidth]{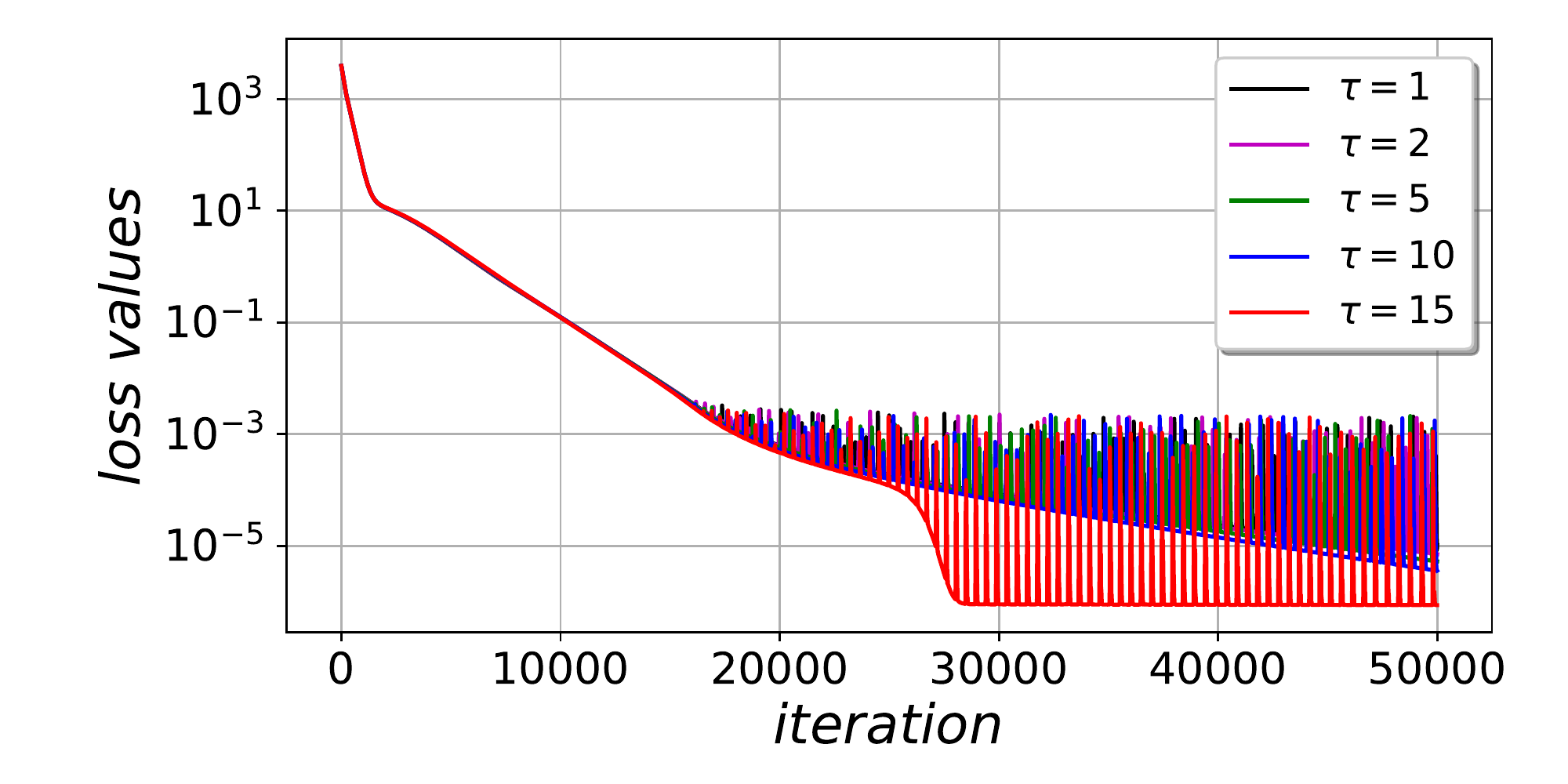}
		\\
		\includegraphics[clip, trim=1cm 0cm 0.8cm 0cm, width=0.46\linewidth]{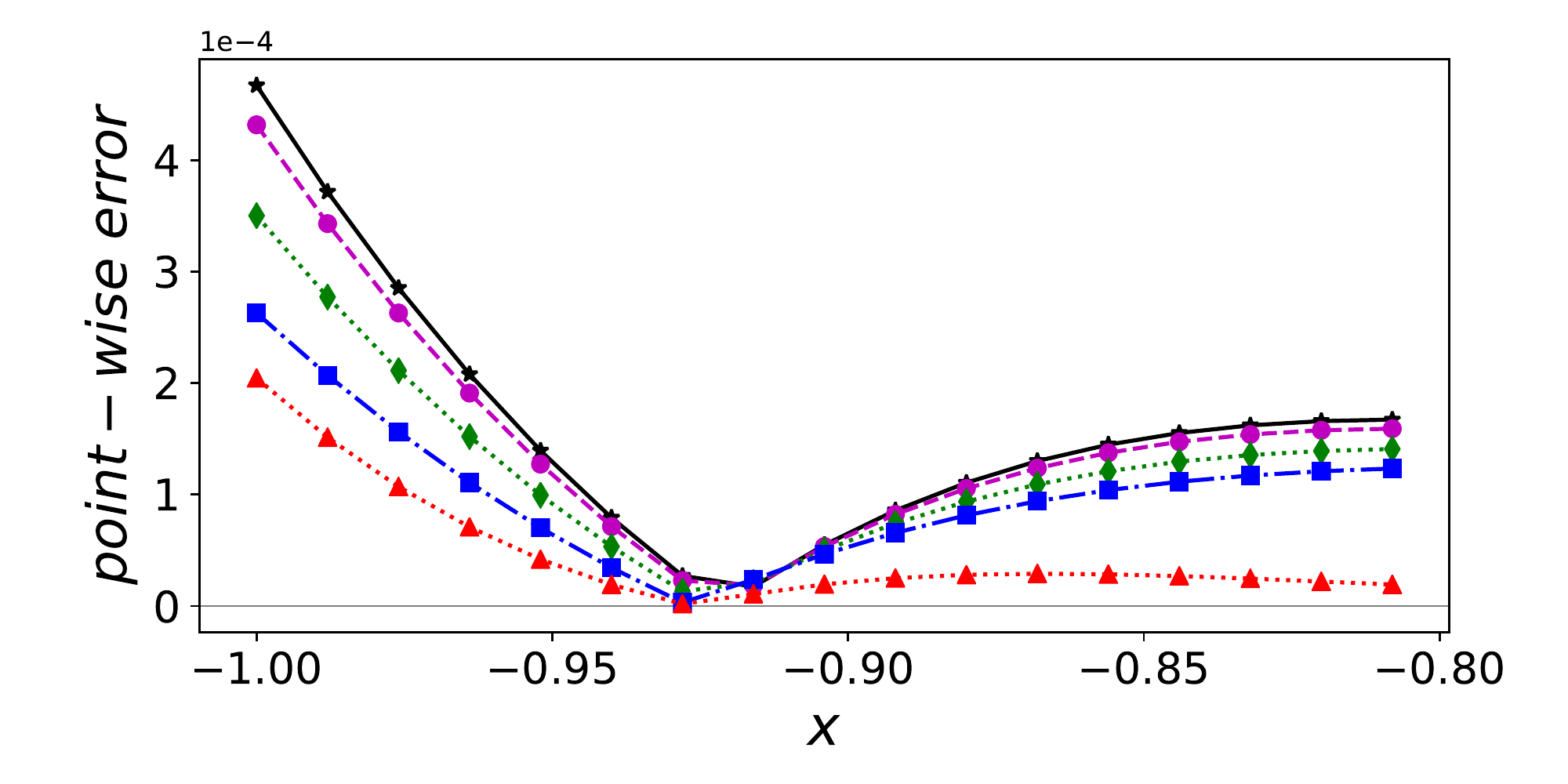}
		&
		\includegraphics[clip, trim=1cm 0cm 0.8cm 0cm, width=0.46\linewidth]{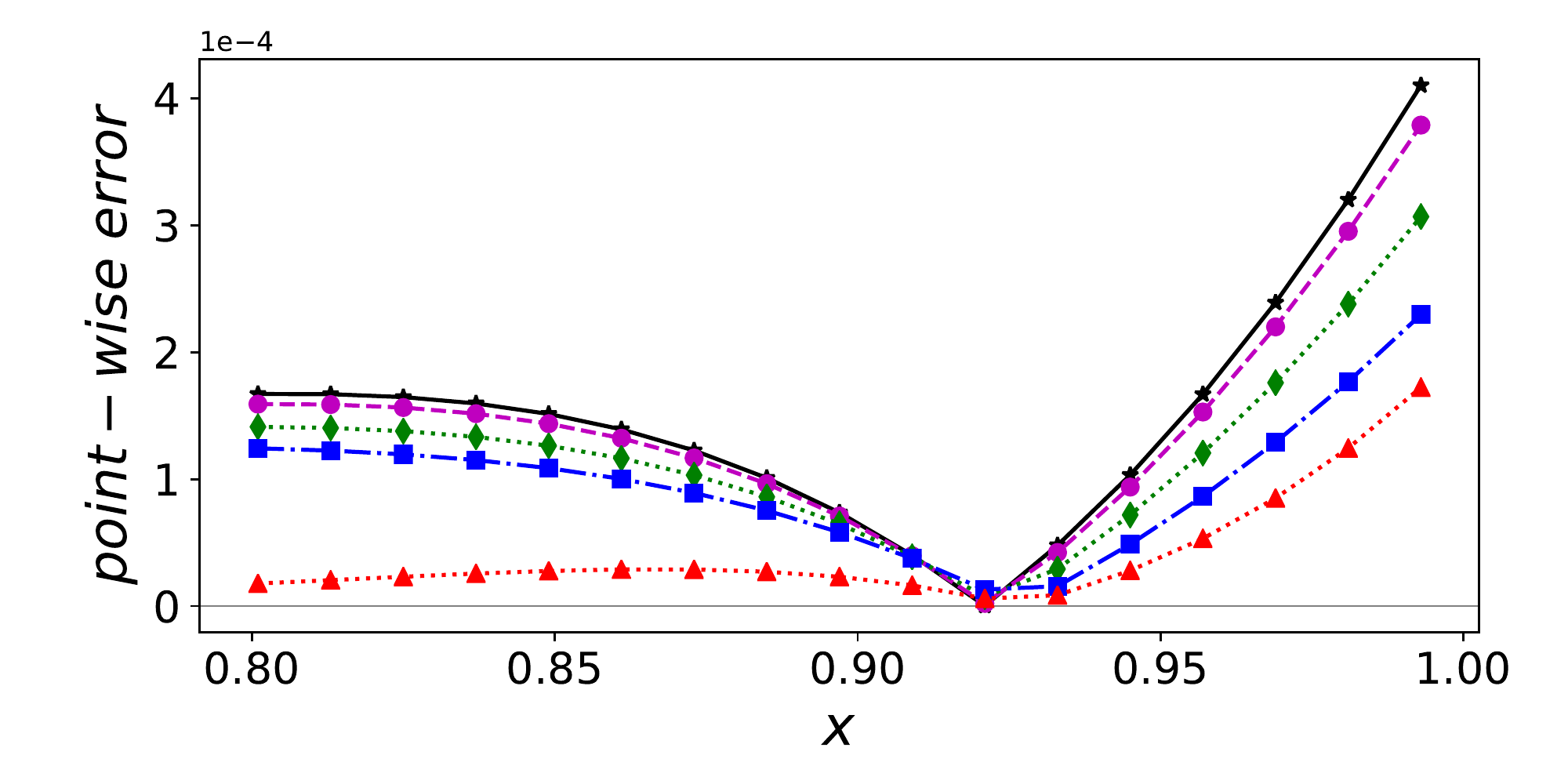}
	\end{tabular}
	\vspace{-0.2 in}
	\caption{\scriptsize \label{Fig: St Burger VR3 Lbw compare} One-dimensional steady state Burger's equation: effect of penalty parameter $\tau$ in the $\mathcal{R}^{(3)}$ formulation. Top Left: comparison of exact solution $\sin(2.1 \pi x)$ and VPINN approximation for different $\tau$. Top Right: loss value. Bottom: the zoom-in display of point-wise error close to boundaries. See Table \ref{Table: VPINN para burger} for VPINN hyperparameters.}
\end{figure}

\vspace{0.1 in}
\noindent $\bullet$ \textbf{Numerical Examples:}
Here, we examine the performance of our proposed method using a shallow network with sine activation function to solve the steady state Burger's equation. We also employ sine test functions and train the network using the loss function defined in \eqref{Eq: loss VPINN St Burger}. 
\begin{exm}
	\label{Ex: STBurger - singlemodal}	
	We consider equation \eqref{Eq: steady Burger} and let the exact solution, boundary values, and force term be of the form 
	$$u^{exact} = A\sin(\omega x), \quad g = A\sin(-\omega), \quad h = A\sin(\omega),$$ 
	$$f(x) =A^2 \omega/2 \sin(2\omega x) +A \omega^2 \sin(\omega x).$$
	We show the results for variational forms $\mathcal{R}^{(1)}$ and $\mathcal{R}^{(2)}$ in Figs. \ref{Fig: St Burger}-\ref{Fig: St Burger compare}, and for variational form $\mathcal{R}^{(3)}$ in Fig. \ref{Fig: St Burger VR3 Lbw compare}. 
\end{exm}

%%%%%%%%%%%%%%%%%%%%%%%%%%%%%%%%%%%%%%%%%%%%%%%%%%%%%%%%%%%%%%%%%%%%%%%%%%%%%%%%%%%%%%%%%%%%%%%%%%%%%%%%%%%%
\begin{table}[h]
\center
\caption{ \scriptsize{ \label{Table: VPINN para burger} One dimensional steady state Burger's equation: Neural network, optimizer, and VPINN parameters.}}
\vspace{-0.1 in}
\scalebox{0.8}{
	\begin{tabular}[t]
		{l l}
		\multicolumn{2}{c}{VPINN}\\
		\hline
		\hline
		variational form & $\mathcal{R}^{(1)}$, $\mathcal{R}^{(2)}$ , $\mathcal{R}^{(3)}$\\ \hline
		$\#$ test functions & 5     \\ \hline 
		test functions & $\sin(k\pi x)$  \\ \hline 
		$\tau$ & 5  \\ \hline
		\hline
		%\vspace{0.2 in}
		%
	\end{tabular}
}
%	\vspace{-0.1 in}
\scalebox{0.8}{
	\begin{tabular}[t]
		{l l}
		\multicolumn{2}{c}{NN}\\
		\hline
		\hline
		$\#$ hidden layers & 1  \\ \hline 
		$\#$ neurons in each hidden layer & 5 \\ \hline   
		activation function & $\text{sine}$ \\ \hline
		optimizer & Adam \\ \hline 
		learning rate & $10^{-3}$ \\ \hline
		\hline \\ 
		%\vspace{0.2 in}
		%
	\end{tabular}
}	
\end{table}
%
%
%******************************************************************************************
%******************************************************************************************
%
%
%******************************************************************************************
%******************************************************************************************
%

%
%******************************************************************************************
%******************************************************************************************
%

\noindent $\bullet$\textbf{Discussion:} In Example \ref{Ex: STBurger - singlemodal}, the exact solution is the sine function with a fixed frequency $\omega$. In the developed VPINN to solve this problem, we construct a shallow network of one hidden layer with $N$ neurons and sine activation function. Therefore, each neuron represents a sine function with adaptive frequency $w_j$. Ideally, one neuron would be sufficient to exactly capture the solution if the neuron frequency $w_j$ could be precisely optimized to exact frequency $\omega$. This is not the case in practice however, as the optimizer may fail to converge if the network initialization is very far from the target values. In general, weights and biases are initialized from known probability distributions. Xavier initialization \cite{glorot2010understanding} is one of the most widely used initialization method, which initializes the weights in the network by drawing them from a distribution with zero mean and a finite variance. The optimal value of variance is given $1/N$, where $N$ is the number of nodes feeding into that layer. This optimal value, however, leads to optimizer failure in our VPINN formulation for the problem at hand. This is due to the fact that the neuron frequencies are initially drawn from a very narrow distribution and therefore they may fall very far from the target frequency. To avoid this failure, we need to widen the distribution and increase the number of neurons to make sure at least one of the neuron frequencies will fall close enough to the target frequency. We also note that the construction of force vector is carried out exactly and we do not use any quadrature rules. Therefore, we can isolate the error associated with the optimization process of the network. Figure \ref{Fig: St Burger} shows the point-wise error, averaged over several different random network initializations. In the most successful case of optimization, we report the $L^{\infty}$-norm error of order $10^{-9}$ for this specific problem.

In the formulation of $\mathcal{R}^{(3)}$ in \eqref{Eq: 1-d BVP var residue - 3} and then later in \eqref{Eq: 1-d BVP var residue - 3 - shallow sine} for a shallow network, we use Remark \ref{Rem: VR3} to replace the boundary terms in the variational loss with the exact known boundary conditions. In this case, the convergence of error depends strongly on how well the network learns the boundary conditions, in that any error from the inaccurate approximation of the boundary condition will propagate through the interior domain. Therefore, by increasing the penalty coefficient in the loss function, we mitigate the error at the boundaries and contain its adverse propagation to the interior domain; see Fig. \ref{Fig: St Burger VR3 Lbw compare}.  

The PINN formulation with sine activation function fails to lead to an accurate approximation. We used a shallow network with single hidden layer, $N = 50$ neurons, $N_r = 1000$ penalizing points, and obtained the $L^{\infty}$ approximation error of or $O(1)$. We note, however, that the choice of tanh activation function performs much more accurately, and with similar network structure results in $L^{\infty}$ approximation error of or $O(10^{-5})$. This is still not comparable with VPINN, as in VPINN, we obtain a much more accurate result with a much simpler network.

\begin{figure}[h]
	\center
	\begin{tabular}{c l}
		\multicolumn{2}{c}{$u^{exact} = (1-x^2) \sin(2.1 \pi x)$} \\  [-2 pt] 
		\multicolumn{2}{c}{\includegraphics[clip, trim=1cm 0cm 0cm 0cm, width=0.46\linewidth]{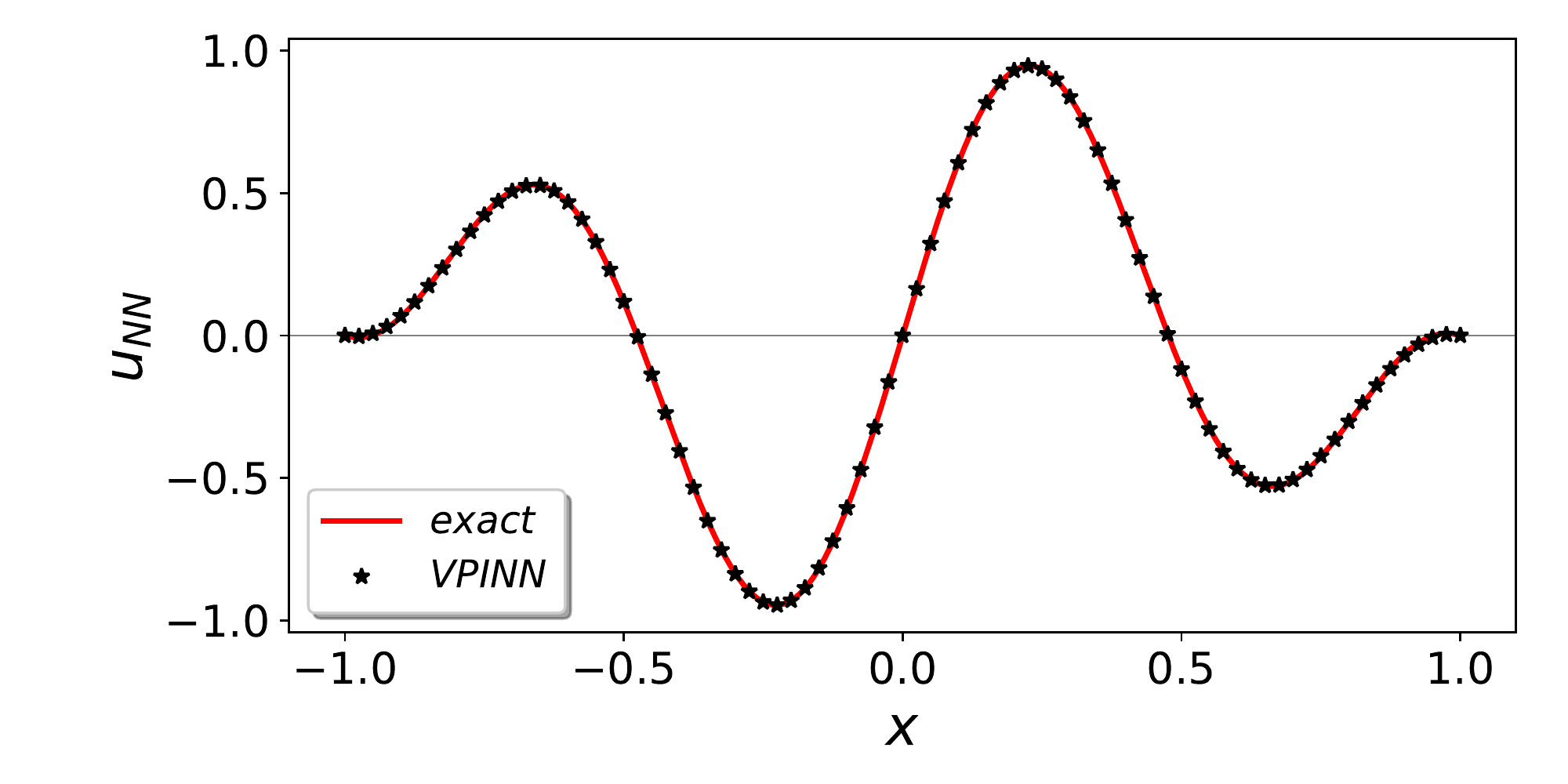}} \\ 
		\multicolumn{2}{c}{$\mathcal{R}^{(2)}$ Formulation} \\  [-2 pt] 
		\includegraphics[clip, trim=1cm 0cm 0cm 0cm, width=0.46\linewidth]{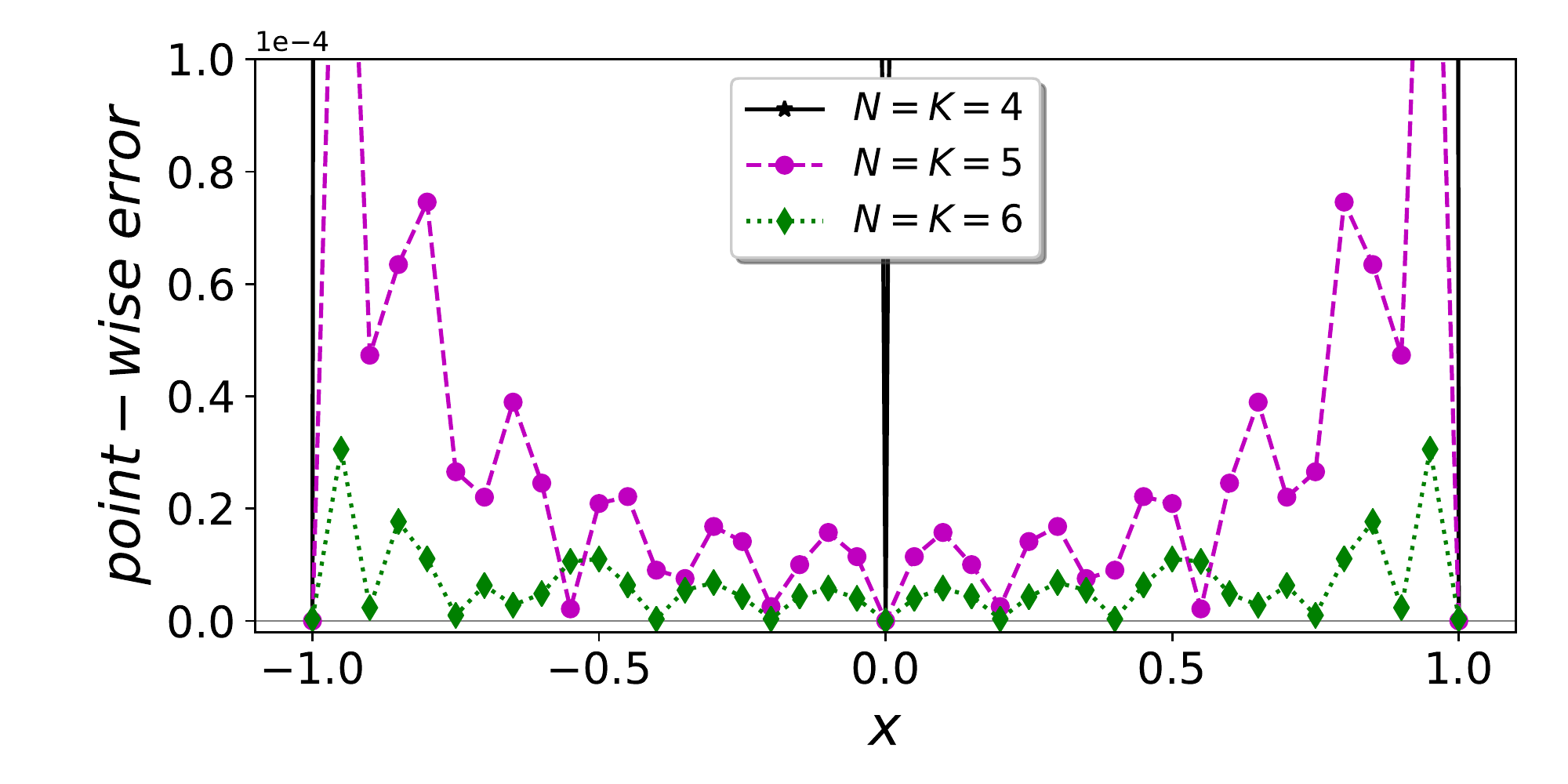}
		&
		\includegraphics[clip, trim=1cm 0cm 0cm 0cm, width=0.46\linewidth]{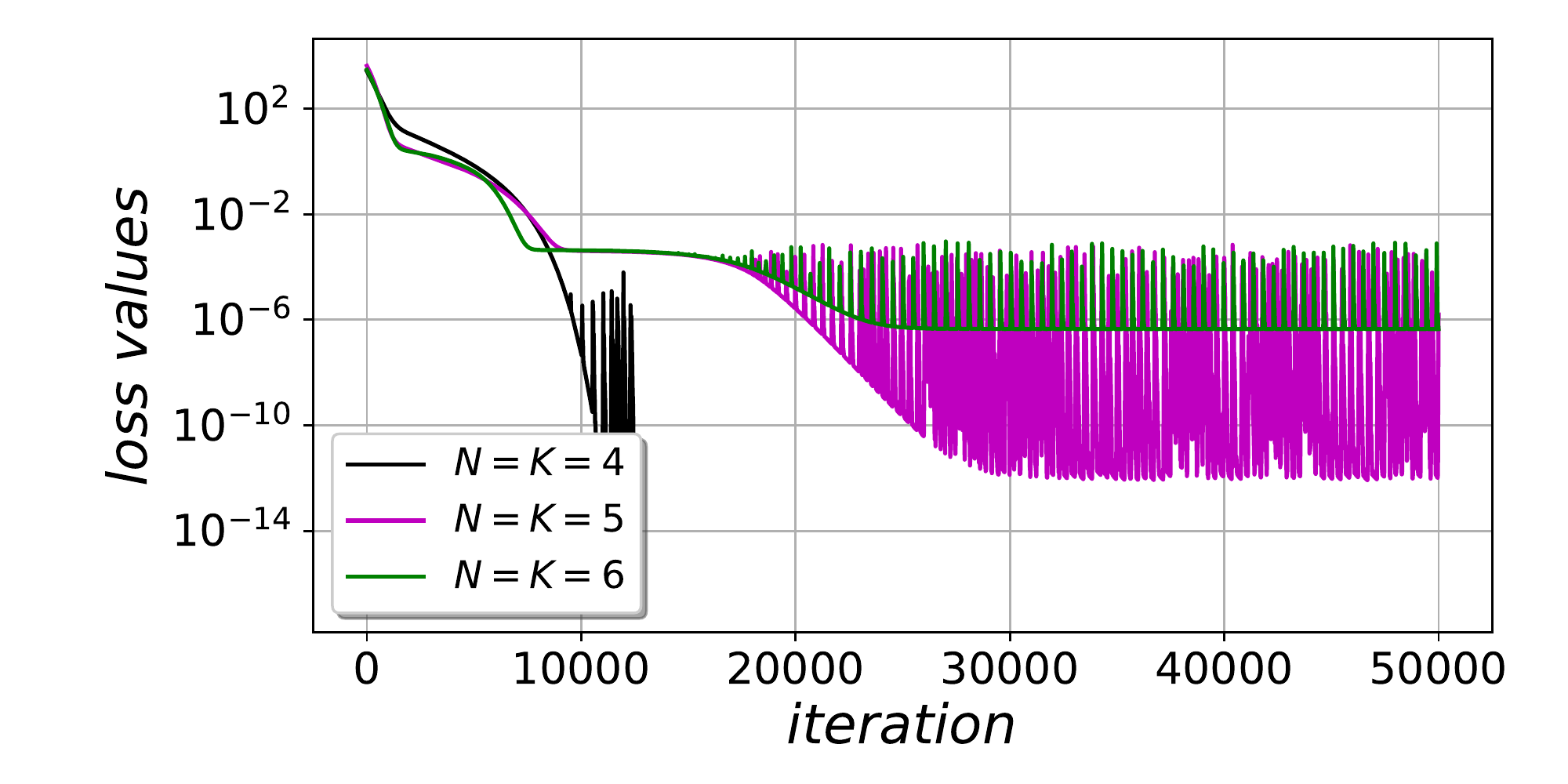}
		\\	
		\multicolumn{2}{c}{$\mathcal{R}^{(3)}$ Formulation} \\  [-2 pt]  
		\includegraphics[clip, trim=1cm 0cm 0cm 0cm, width=0.46\linewidth]{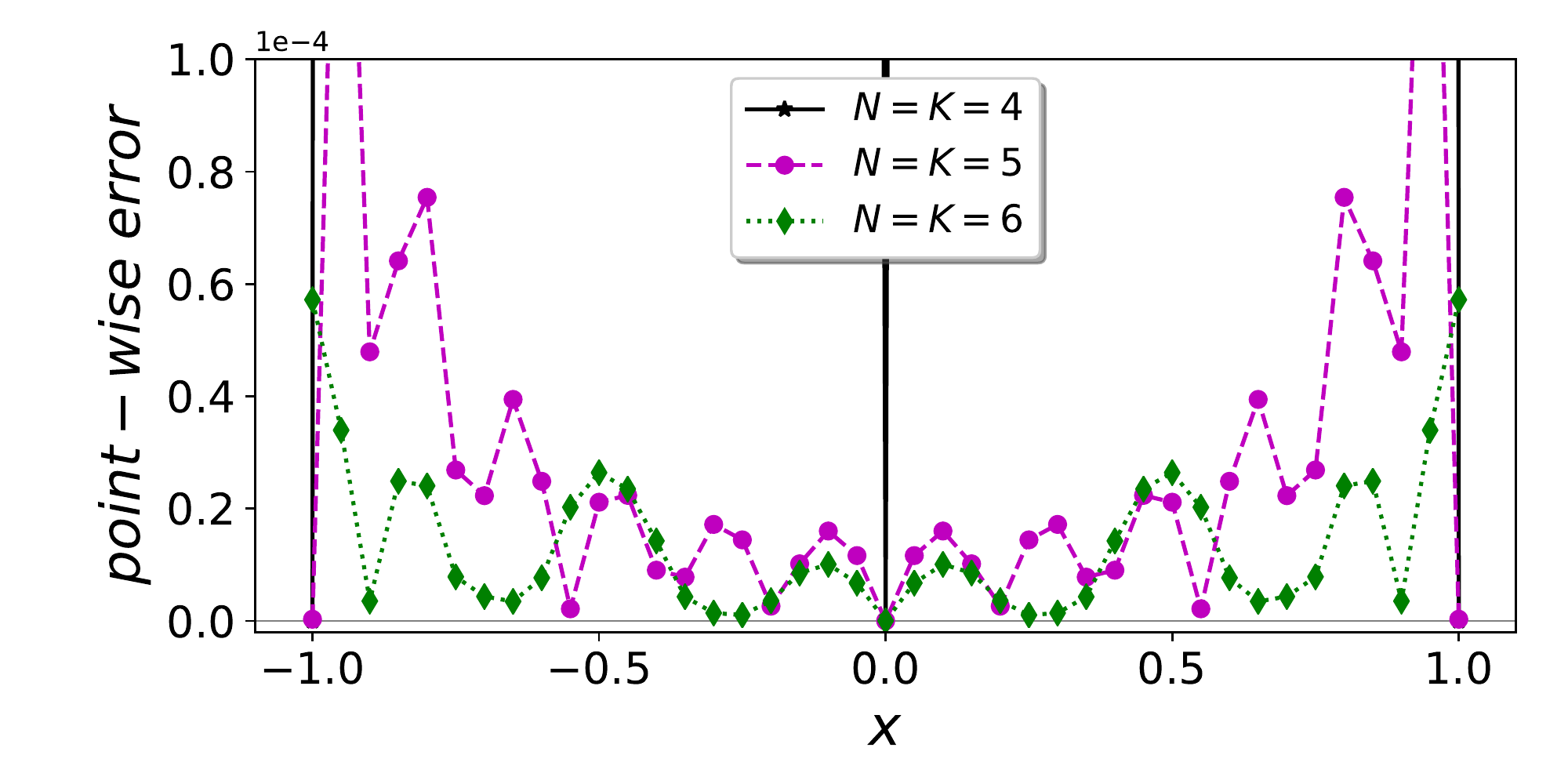}
		&
		\includegraphics[clip, trim=1cm 0cm 0cm 0cm, width=0.46\linewidth]{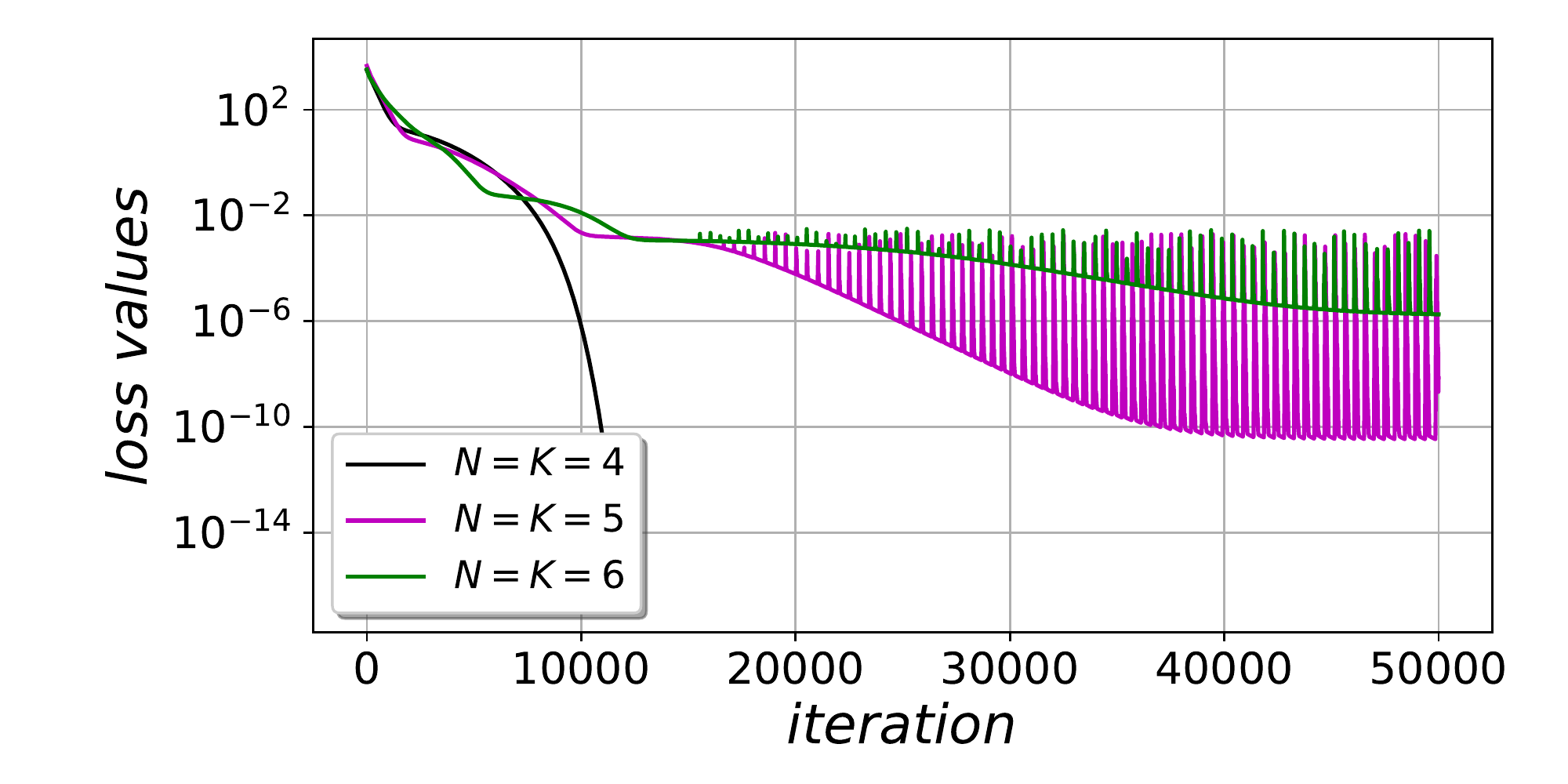}
	\end{tabular}
	\caption{\scriptsize \label{Fig: St Burger Vanishing Bound NK compare} Example \ref{Ex: STBurger - vanishing boundary}. One-dimensional steady state Burger's equation: error convergence by increasing $N$ (number of neurons) and $K$ (number of test functions). Shallow network with $L=1$ hidden layer, sine activation, and sine test function.}
\end{figure}

\begin{exm}
	\label{Ex: STBurger - vanishing boundary}	
	We consider equation \eqref{Eq: steady Burger}, however, we let the exact solution vanish at the two boundaries, i.e.  
	$$u^{exact} = A (1-x^2) \sin(\omega x), \quad g = h = 0.$$ 
	The force term $f(x)$ can be obtained easily by substituting $u^{exact} $ in \eqref{Eq: steady Burger}.
	We show the error convergence by increasing the number of neurons and test functions for variational forms $\mathcal{R}^{(1)} = \mathcal{R}^{(2)}$ and $\mathcal{R}^{(3)}$ in Fig. \ref{Fig: St Burger Vanishing Bound NK compare}. In all cases, the value of the penalty parameter is fixed to $\tau = 100$. 
\end{exm}

%
%******************************************************************************************
%******************************************************************************************

\noindent $\bullet$\textbf{Discussion:} 
We first consider the projection of the exact solution $u^{exact}$ onto the network. By using the sine test functions \eqref{Eq: sine test functions}, we introduce the variational residual of projection and the loss function as
\begin{align*}
%\label{Eq: projection}
u^{exact} & \approx u_{NN}(x) = \sum_{j=1}^{N} a_j \, \sin \left(w_j \, x + \theta_j  \right),
\\
\mathcal{R}^{proj}_k & = \left( u_{NN}(x) , v_k(x) \right)_{\Omega} ,
\quad
U_k  = \left( u^{exact}(x) , v_k(x) \right)_{\Omega} ,
\\
L^{proj} 
&= \frac{1}{K} \sum_{k = 1}^{K} \Big| \mathcal{R}^{p}_k - U_k \Big|^2 
+ \tau \left( \Big|u_{NN}(-1) - g \Big|^2 + \Big|u_{NN}(1) - h \Big|^2 \right).
\end{align*}
Following similar steps as in \eqref{Eq: 1-d BVP var residue - 3 - shallow sine}, we can obtain an analytical expression of the variational residual,
\begin{align}
\label{Eq: proj var residue - shallow sine}
\mathcal{R}^{proj}_k &= 2 (-1)^k \, k \, \pi \,  \sum_{j=1}^{N} \frac{a_j \, \cos(\theta_j) \sin(w_j)}{w_j^2 - k^2 \pi^2}.
\end{align}
The convergence of $u_{NN}(x) \rightarrow u^{exact} $ by minimizing the loss function $L^{proj}$ depends strongly on the number of neurons $N$, number of test functions $K$, and performance of optimization. We observe that we can obtain the best $L^{\infty}$ projection error of order $O(10^{-4})$ by choosing $N = K = 5$, where any other combination leads to less accurate results. In solving the differential equation, we expect VPINN to give an accuracy level of almost similar order as the projection. In example \ref{Ex: STBurger - vanishing boundary}, the periodic boundary conditions simplifies the $\mathcal{R}^{(3)}$ formulation by removing the term $(h-g)$ in \eqref{Eq: 1-d BVP var residue - 3 - shallow sine}. However, the approximation in Remark \ref{Rem: VR3} still exists, and thus, $\mathcal{R}^{(3)}$ does not necessarily perform better, compared to the other formulations. We see that by increasing the number of neurons and test functions, the error over the whole domain decreases. It will, however, saturate at some points, where the optimization error becomes dominant. 

In general, we can decompose the expected error associated with VPINN into four main types: NN approximation (expressivity of NN), spectral projection, numerical integration, and optimization. Yet, it is hard to isolate single sources in the analysis of error as they are strongly related to each other. High expressivity, i.e. higher number of neurons $N$ in the network, may complicate the loss function, leading to poor optimization performance. More importantly, the choice of sine activation function further requires additional care in network initialization, as was explained in example \ref{Ex: STBurger - singlemodal}. In our derivations for the shallow network, we remove the error of numerical integration by analytically expressing the loss functions. Then, by increasing $N$ and $ K$, we seek the best combination that leads to better accuracy.

%
%%%%%%%%%%%%%%%%%%%%%%%
\subsection{Shallow Network with Sine Activation Functions and Polynomial Test Functions}
%%%%%%%%%%%%%%%%%%%%%%%
%
By considering sine activation function, we construct a similar shallow network as in \eqref{Eq: shallow NN sine}; also shown in Fig. \ref{Fig: Shallow NN}. Here, we choose a combination of Legendre polynomials as test functions, i.e.,
\begin{align}
\label{Eq: Legendre test functions}
v_k (x) = P_{k+1}(x) - P_{k-1}(x), \quad k=1,2,\cdots, K,  
\end{align}
which naturally vanish at the boundary points $v_k (-1)=v_k (1)=0 $.
We note that the construction of variational residuals in this case is not unique, as one can use different properties of Jacobi polynomials to take the integrals and derive various formulations. We use the recursion formula of Legendre polynomials to obtain the variational residuals \eqref{Eq: 1-d BVP var residue - 1}-\eqref{Eq: 1-d BVP var residue - 2}. Therefore, 
\begin{align}
\label{Eq: 1-d BVP var residue - shallow sine Leg test - recursion VF1}
\mathcal{R}^{(1)}_k 
& 
= \sum_{j=1}^{N} a_j \, w_j^2 \int_{-1}^{1} \sin(w_j x + \theta_j ) \, \left( P_{k+1}(x) - P_{k-1}(x) \right) \, dx
\\ \nonumber
& 
= \sum_{j=1}^{N} a_j \, w_j^2 \int_{-1}^{1} \text{Im} \lbrace e^{i(w_j x + \theta_j) } \rbrace \, \left( P_{k+1}(x) - P_{k-1}(x) \right) \, dx
%\\ \nonumber
%%
%& 
%= \sum_{j=1}^{N} a_j \, w_j^2 \,\,  \text{Im} \big\lbrace e^{i\theta_j} \int_{-1}^{1} e^{i w_j x }  \, \left( P_{k+1}(x) - P_{k-1}(x) \right) \, dx \big\rbrace
\\ \nonumber
& 
= \sum_{j=1}^{N} a_j \, w_j^2 \,\,  \text{Im} \big\lbrace e^{i\theta_j} \left( I_{k+1}(w_j) - I_{k-1}(w_j) \right)  \big\rbrace
= \sum_{j=1}^{N} a_j \, w_j^2 \,\,  \left( C_{k+1}(w_j) - C_{k-1}(w_j) \right)
\end{align}
and
\begin{align}
\label{Eq: 1-d BVP var residue - shallow sine Leg test - recursion VF2}
\mathcal{R}^{(2)}_k 
& 
= \sum_{j=1}^{N} a_j \, w_j \int_{-1}^{1} \cos(w_j x + \theta_j ) \, \frac{d}{dx}\left( P_{k+1}(x) - P_{k-1}(x) \right) \, dx
\\ \nonumber
& 
= \sum_{j=1}^{N} a_j \, w_j \int_{-1}^{1} \text{Re} \lbrace e^{i(w_j x + \theta_j) } \rbrace \, (2k+1) P_{k}(x)  \, dx
%\\ \nonumber
%%
%& 
%= (2k+1) \sum_{j=1}^{N} a_j \, w_j \,\,  \text{Re} \big\lbrace e^{i\theta_j} \int_{-1}^{1} e^{i w_j x }  \, P_{k}(x)  \, dx \big\rbrace
\\ \nonumber
& 
= (2k+1) \sum_{j=1}^{N} a_j \, w_j \,\,  \text{Re} \big\lbrace e^{i\theta_j} I_{k}(w_j)  \big\rbrace
= (2k+1) \sum_{j=1}^{N} a_j \, w_j \,\,  B_{k+1}(w_j) ,
\end{align}
where $i = \sqrt{-1}$, $\text{Re}\{\cdot\}$ and $\text{Im}\{\cdot\}$ are the real and imaginary parts, respectively, and the following quantities
\begin{align*}
\label{Eq: 1-d BVP var residue - shallow sine Leg test - recursion formulas}
I_{k}(w_j) = \int_{-1}^{1} e^{i w_j x  } P_{k}(x)  \, dx, \quad
C_{k}(\theta_j, w_j) = \text{Im} \big\lbrace e^{i\theta_j} I_{k}(w_j)  \big\rbrace, \quad
B_{k}(\theta_j, w_j) = \text{Re} \big\lbrace e^{i\theta_j} I_{k}(w_j)  \big\rbrace,
\end{align*}
have the following recursion formulas for $k = 2,3, \cdots, K$
\begin{align*}
\label{Eq: 1-d BVP var residue - shallow sine Leg test - recursion formulas 2}
&I_{k} = i \frac{2k-1}{w_j} I_{k-1} + I_{k-2}, \,\,
&&I_0 = \frac{2 \sin(w_j)}{w_j}, \,\,
&&I_1 = -\frac{2 i }{w_j}\left( \sin(w_j) - \frac{\cos(w_j)}{w_j} \right),
\\
&C_{k} = \frac{2k-1}{w_j} B_{k-1} + C_{k-2}, \,\,
&&B_0 = \frac{2 \sin(w_j)\cos(\theta_j)}{w_j} \,\,
&&C_1 = -\frac{2 \cos(\theta_j) }{w_j}\left( \sin(w_j) - \frac{\cos(w_j)}{w_j} \right),
\\
&B_{k} = -\frac{2k-1}{w_j} C_{k-1} + B_{k-2}, \,\,
&&C_0 = \frac{2 \sin(w_j)\sin(\theta_j)}{w_j}, \,\,
&&B_1 = \frac{2 \sin(\theta_j) }{w_j}\left( \sin(w_j) - \frac{\cos(w_j)}{w_j} \right).
\end{align*}
The full derivations of recursive formulas are given in Appendix \ref{Sec: Appx shallow sine test Leg}.

\begin{figure}[h]
	\center
	\begin{tabular}{c l}
		\multicolumn{2}{c}{$\quad\quad u^{exact} = 0.1 \sin(4\pi x) + \tanh(5 x)$} \\  [-2 pt] 
		\multicolumn{2}{c}{\includegraphics[clip, trim=1cm 0cm 0cm 0cm, width=0.46\linewidth]{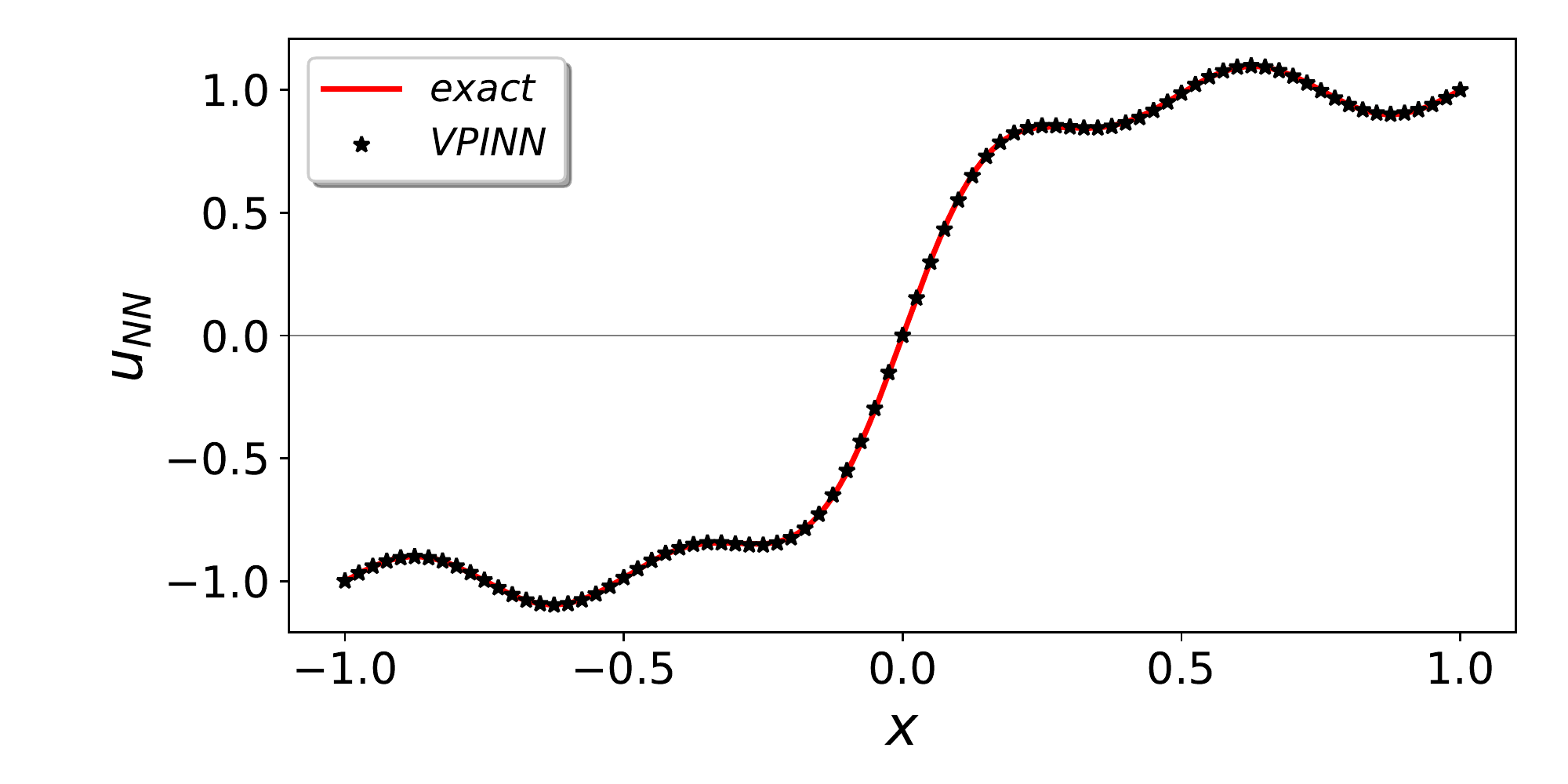}} \\ 
		%		\multicolumn{2}{c}{VPINN: increasing $N=K$ } \\  [-1.5 pt]  
		\includegraphics[clip, trim=1cm 0cm 0cm 0cm, width=0.46\linewidth]{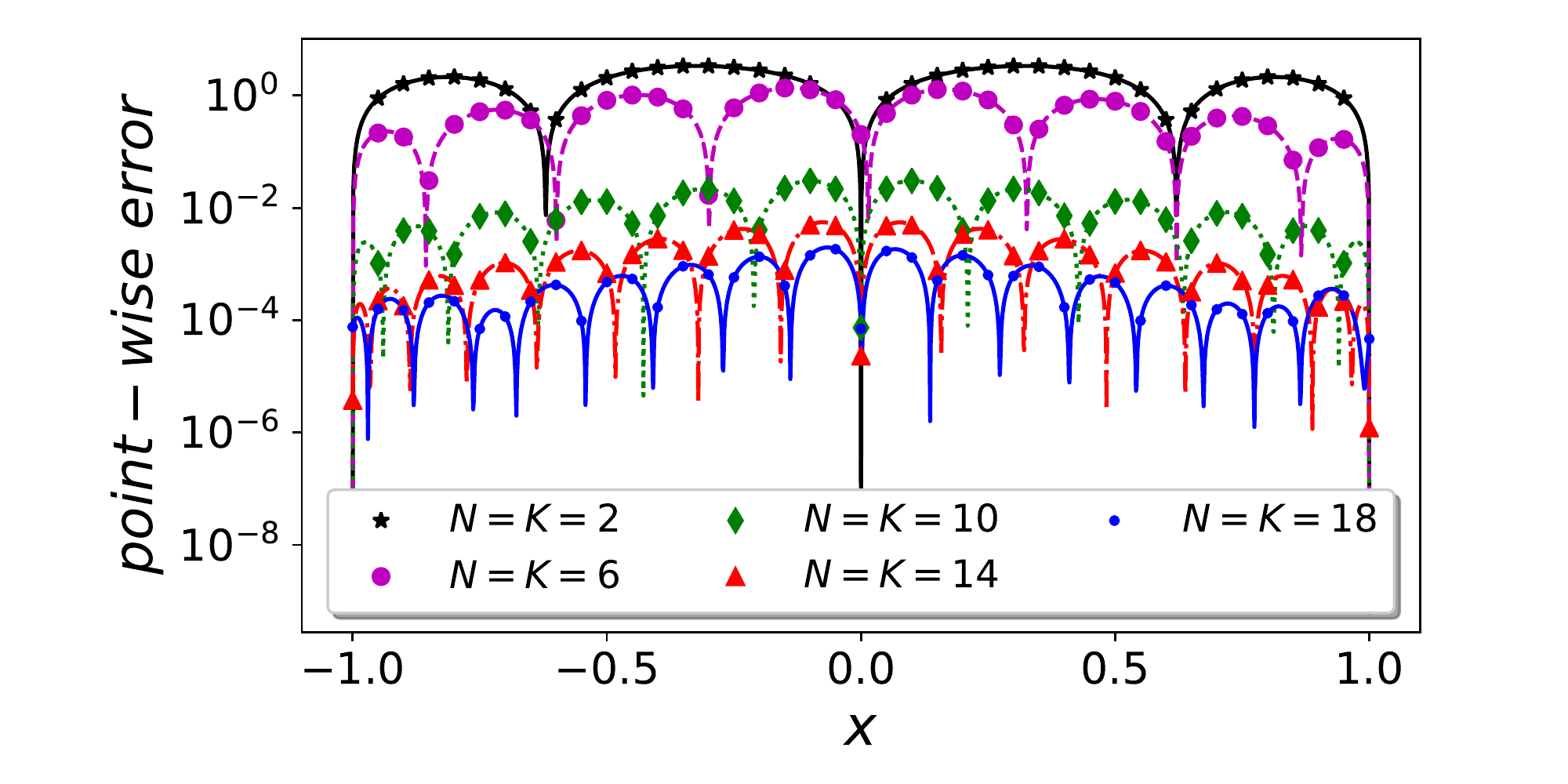}
		&
		\includegraphics[clip, trim=1cm 0cm 0cm 0cm, width=0.46\linewidth]{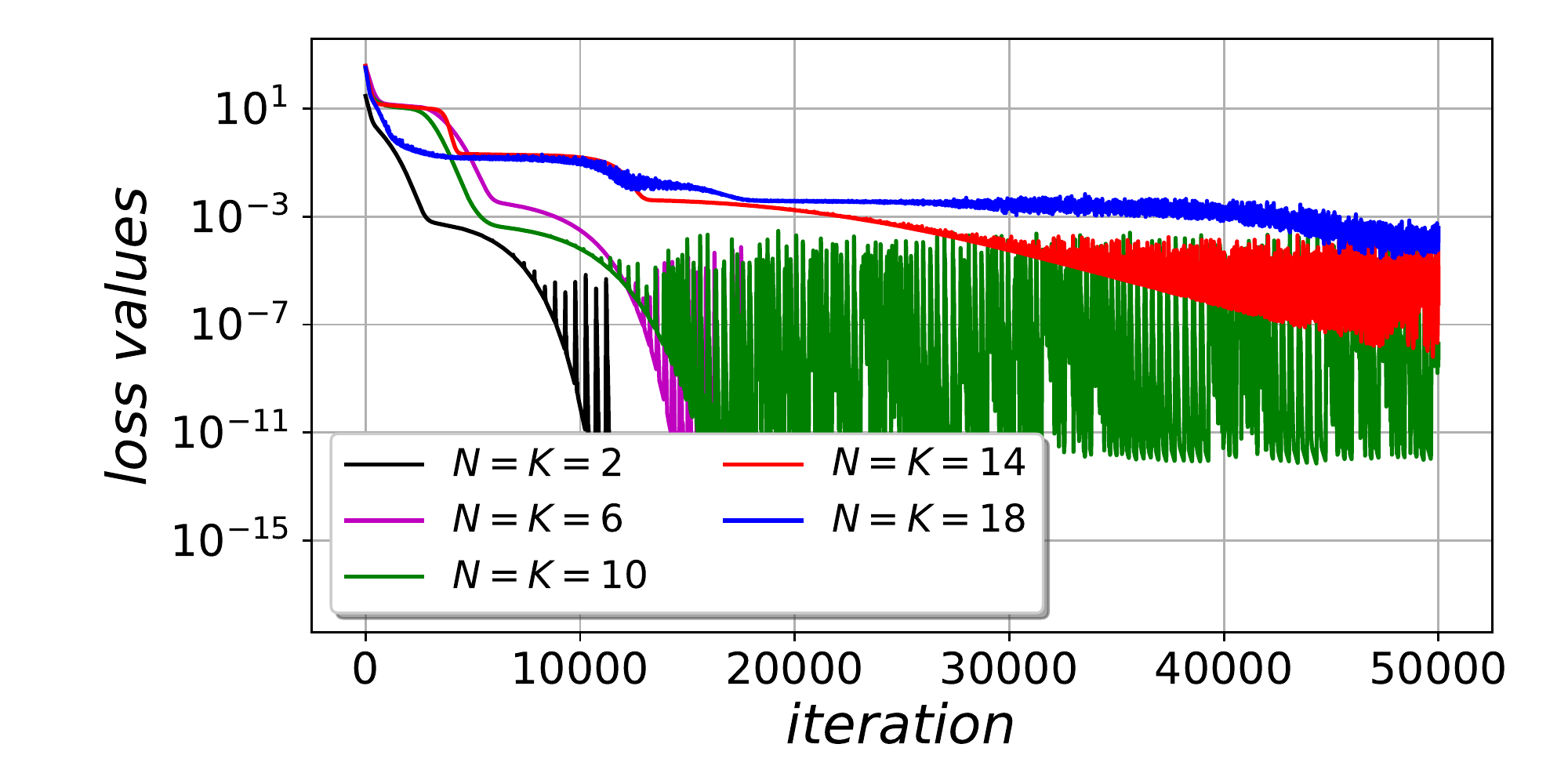}
	\end{tabular}
	\caption{\scriptsize \label{Fig: Poisson NK compare} One-dimensional Poisson's equation: error convergence by increasing number of neurons $N$ and number of test functions $K$. VPINN: $\mathcal{R}^{(1)}$ formulations of shallow network with $L=1$ hidden layer, sine activation, and Legendre test function. A shallow PINN with $L=1$ hidden layer, $N=50$ neurons, sine activation, and $N_r = 1000$ penalizing points has the $L^{\infty}$ error of $O(10^{-1})$.}
\end{figure}

\vspace{0.1 in}
\noindent $\bullet$ \textbf{Numerical Examples:}
Here, we consider the Poisson's equation as the analytical expression of nonlinearity becomes very complicated, if not impossible, using polynomial test functions. 
Thus,
\begin{align}
\label{Eq: Poisson Eqn}
 - \frac{d^2 u}{d x^2} = f(x),
\quad u(-1) = g, \,\, u(1)=h.
\end{align}
The variational residual of the network then becomes
\begin{align}
\label{Eq: Poisson var residue - 0}
Residual^{\mathfrak{v}}_k = \mathcal{R}_k - F_k - \mathcal{R}^b_k,
\end{align}
and therefore, the corresponding \emph{variational loss function} for this problem can be written as
\begin{align}
\label{Eq: loss VPINN Poisson}
L^{\mathfrak{v}(i)} 
&= \frac{1}{K} \sum_{k = 1}^{K} \Big| \mathcal{R}^{(i)}_k - F_k \Big|^2 
+ \frac{\tau}{2} \left( \Big|u_{NN}(-1) - g \Big|^2 + \Big|u_{NN}(1) - h \Big|^2 \right),
\quad i=1,2.
\end{align}
\begin{exm}
	\label{Ex: Poisson - steep}	
	We consider equation \eqref{Eq: Poisson Eqn} and let the exact solution be of the form 
	$$u^{exact} = A\sin(\omega x) + \tanh(r x).$$
	The force term is simply obtained by substituting exact solution in the equation.  
	We show the results for $A = 0.1$, $\omega = 4 \pi$, and $r = 5$ in Fig. \ref{Fig: Poisson NK compare}. We set the penalty parameter $\tau = 10$. 
\end{exm}	

%
%******************************************************************************************
%******************************************************************************************

\noindent $\bullet$ \textbf{Discussion:}
The performance of VPINN by employing polynomial test functions \eqref{Eq: Legendre test functions} is much better compared to sine test functions \eqref{Eq: sine test functions} when the exact solution is not of a simple sinusoidal form. However, the analytical expression for the variational residual and the loss function become more complicated. When a high number of test functions is used, the recursion formulas lead to a high exponent of weights in the denominator, which makes the loss function very sensitive to the initialization of weights and biases. In Fig. \ref{Fig: Poisson NK compare}, we show the error convergence of a shallow VPINN for different number of neurons and test functions, based on variational residual $\mathcal{R}^{(1)}$; we observe similar results for the formulation $\mathcal{R}^{(2)}$ too. We note that in Fig. \ref{Fig: Poisson NK compare}, the best approximation for $N=K = 18$ (blue line) has the highest loss value, compared to other cases. This shows that although the network has higher expressivity, the optimizer has more difficulty to train the network, and in case that the optimizer performs better, one can expect more accurate approximation by increasing $N$ and $K$. Compared to a shallow PINN with $L=1$ hidden layer and sine activation, but a much higher number of neurons, $N=50$, the network fails to accurately capture the exact solution and with $N_r = 1000$ penalizing points gives the $L^{\infty}$ error of $O(10^{-1})$. We should mention that we choose the sine activation in PINN to compare with VPINN, however, the tanh activation function can provide more accurate approximation in PINN; we further discuss this in the next section.

%%
%
%%%%%%%%%%%%%%%%%%%%%%%%%%%
\section{Shallow to Deep VPINNs}
\label{Sec: vPINN Deep}
%%%%%%%%%%%%%%%%%%%%%%%%%%%
%
The advantage of neural networks is their high expressivity in their deep constructions. The composition of layers provides a high nonlinear representation of networks to more accurately approximate a function. On the other hand, unlike the shallow networks, the analysis of deep constructions becomes more challenging as they do not lend themselves to simple analytical expressions. In the formulation of our proposed VPINN using deep neural networks, we cannot take the integrals of variational residuals analytically anymore and have to employ a numerical integration technique such as quadrature rules. Yet, there exist no works in the literature on analysis of quadrature integration for compositional function spaces of (deep) neural networks. In this section, we extend our formulation of shallow VPINN to deep networks by employing Gauss type quadrature rules to compute the corresponding integrals in the loss function. Therefore, by choosing quadrature points and weights $\{x_q, W_q\}_{q=1}^{q=Q}$, test functions defined in \eqref{Eq: test function general - 1} and by using Remark \ref{Rem: VR3}, the variational residuals \eqref{Eq: 1-d BVP var residue - 1}-\eqref{Eq: 1-d BVP var residue - 3} can be written as
\begin{align*}
%\label{Eq: 1-d BVP var residue - 1 - quad}
%
\mathcal{R}^{(1)}_k 
&
%= - \left( \frac{d^2 u_{NN}(x)}{d x^2}, v_k(x) \right)_{\Omega} 
\approx \widetilde{\mathcal{R}}^{(1)}_k 
= - \sum_{q=1}^{Q} W_q \, \frac{d^2 u_{NN}(x_q)}{d x^2} v_k(x_q)
\\
%\label{Eq: 1-d BVP var residue - 2 - quad}
\mathcal{R}^{(2)}_k 
&
%= \left( \frac{d u_{NN}(x)}{d x}, \frac{d v_k(x)}{d x} \right)_{\Omega} 
\approx \widetilde{\mathcal{R}}^{(2)}_k 
= \sum_{q=1}^{Q} W_q \, \frac{d u_{NN}(x_q)}{d x} \frac{d v_k(x_q)}{dx}
\\
%\label{Eq: 1-d BVP var residue - 3 - quad}
\mathcal{R}^{(3)}_k 
&
%= - \left( u_{NN}(x) , \frac{d^2 v_k(x)}{d x^2} \right)_{\Omega} + u_{NN}(x) \, \frac{d v_k(x)}{d x} \bigg\vert_{\partial \Omega}
%\\ \nonumber
%& 
\approx \widetilde{\mathcal{R}}^{(3)}_k 
= - \sum_{q=1}^{Q} W_q \, u_{NN}(x_q) \frac{d^2 v_k(x_q)}{d x^2} +  u(x) \, \frac{d v_k(x)}{d x} \bigg\vert_{-1}^{1}
\end{align*}
which only requires the evaluation of network (and/or its derivatives) at several quadrature points $x_q$'s. The above approximation gives us the flexibility to combine different choices of activation functions, test functions, and quadrature rules.

%
%%%%%%%%%%%%%%%%%%%%%%%
\subsection{Numerical Examples}
%%%%%%%%%%%%%%%%%%%%%%%
%
We examine the performance of developed VPINN by considering several numerical examples with different exact solutions. In each case, we use the method of fabricated solutions and obtain the forcing term by substituting the exact solution into the corresponding equation.

\begin{figure}[h]
	\center
	\begin{tabular}{c l}
		\multicolumn{2}{c}{$\quad\quad u^{exact} = 0.1 \sin(4\pi x) + \tanh(5 x)$} \\  [-2 pt] 
		\multicolumn{2}{c}{\includegraphics[clip, trim=1cm 0cm 0cm 0cm, width=0.46\linewidth]{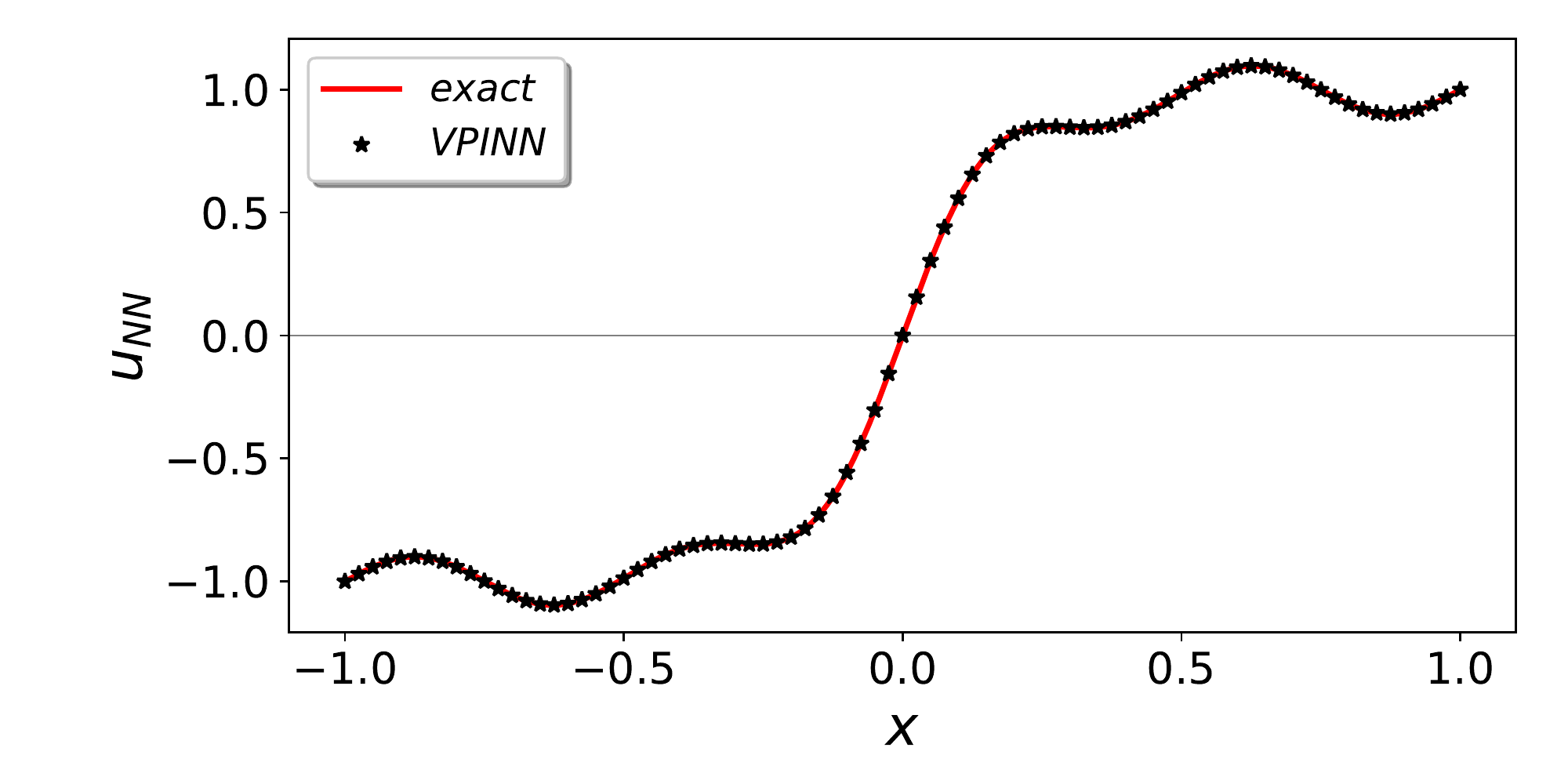}} \\ 
		\multicolumn{2}{c}{VPINN $\widetilde{\mathcal{R}}^{(1)}$} \\  [-1.5 pt]  
		\includegraphics[clip, trim=1cm 0cm 0cm 0cm, width=0.46\linewidth]
		{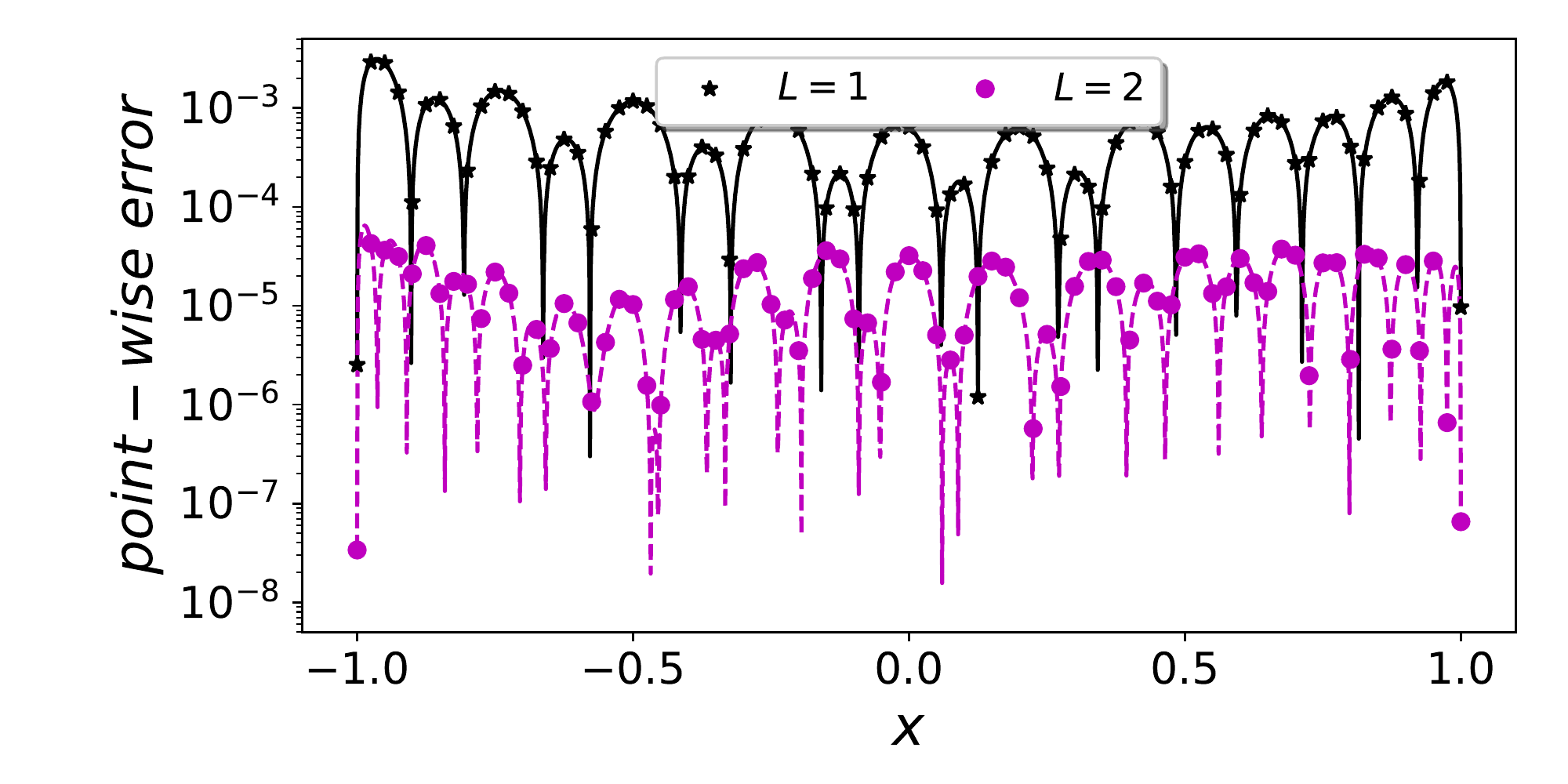}
		&
		\includegraphics[clip, trim=1cm 0cm 0cm 0cm, width=0.46\linewidth]
		{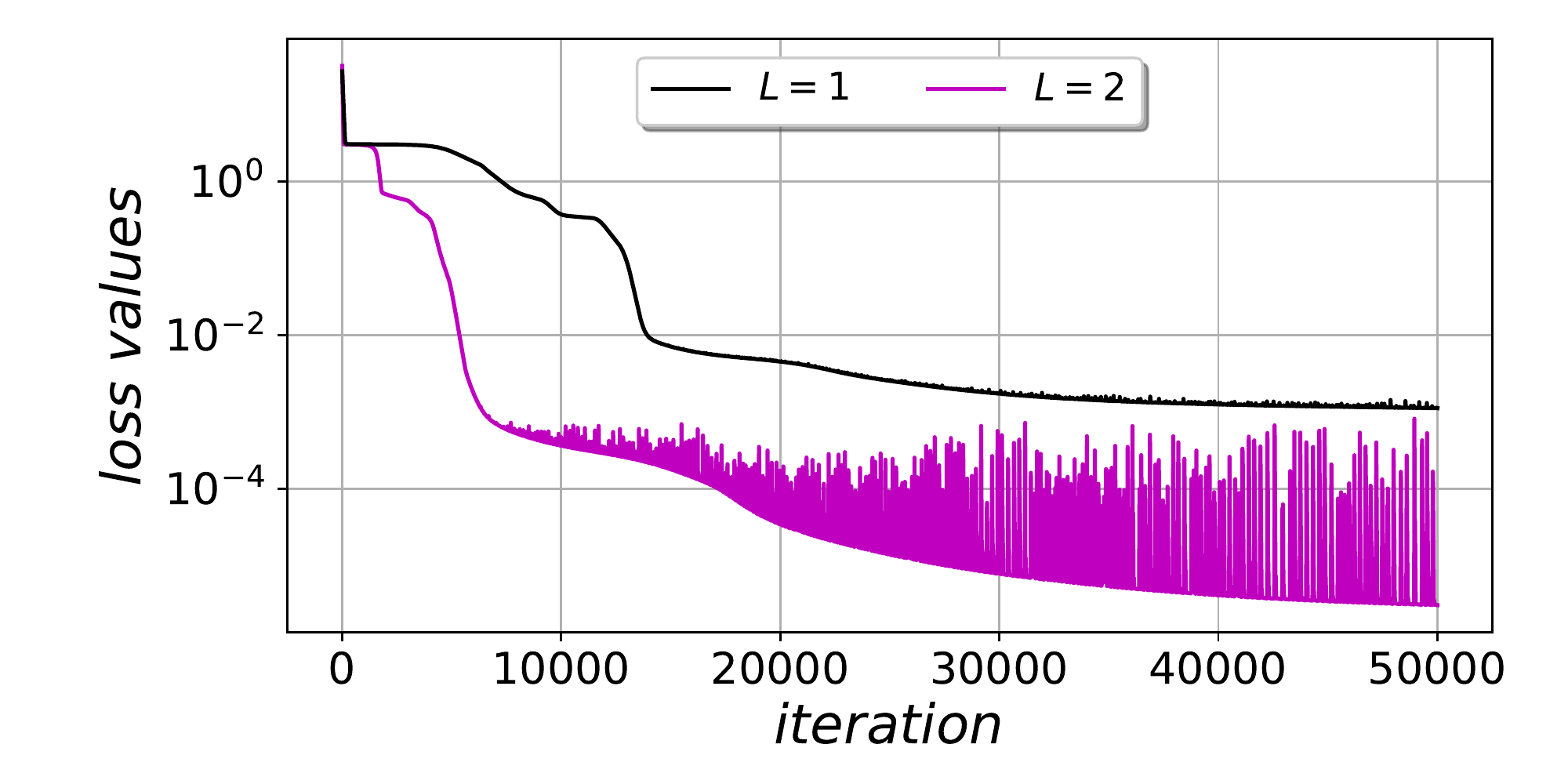}
		\\
		\multicolumn{2}{c}{VPINN $\widetilde{\mathcal{R}}^{(2)}$} \\  [-1.5 pt] 
		\includegraphics[clip, trim=1cm 0cm 0cm 0cm, width=0.46\linewidth]
		{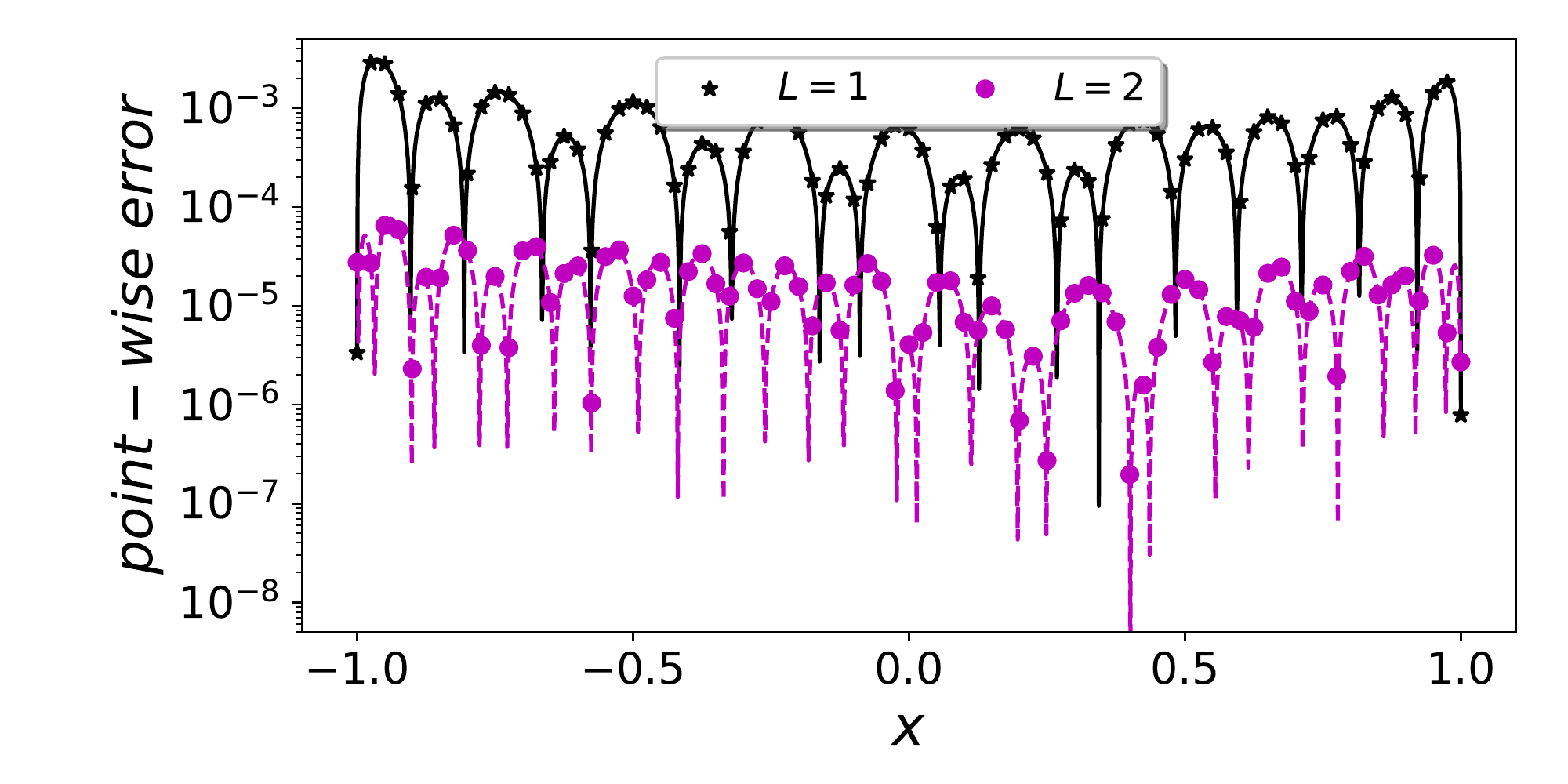}
		&
		\includegraphics[clip, trim=1cm 0cm 0cm 0cm, width=0.46\linewidth]
		{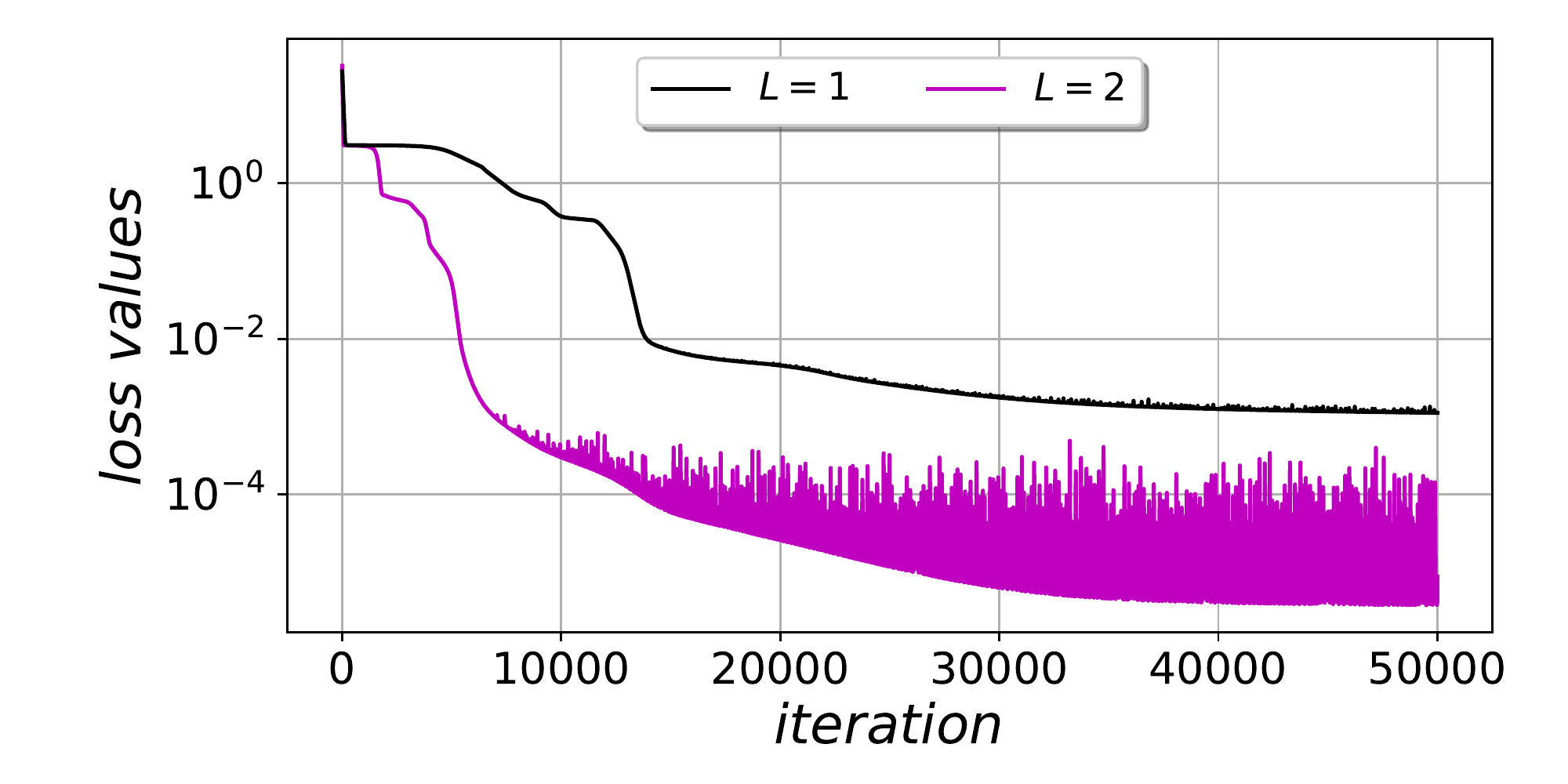}
		\\
		\multicolumn{2}{c}{PINN} \\  [-1.5 pt] 
		\includegraphics[clip, trim=1cm 0cm 0cm 0cm, width=0.46\linewidth]
		{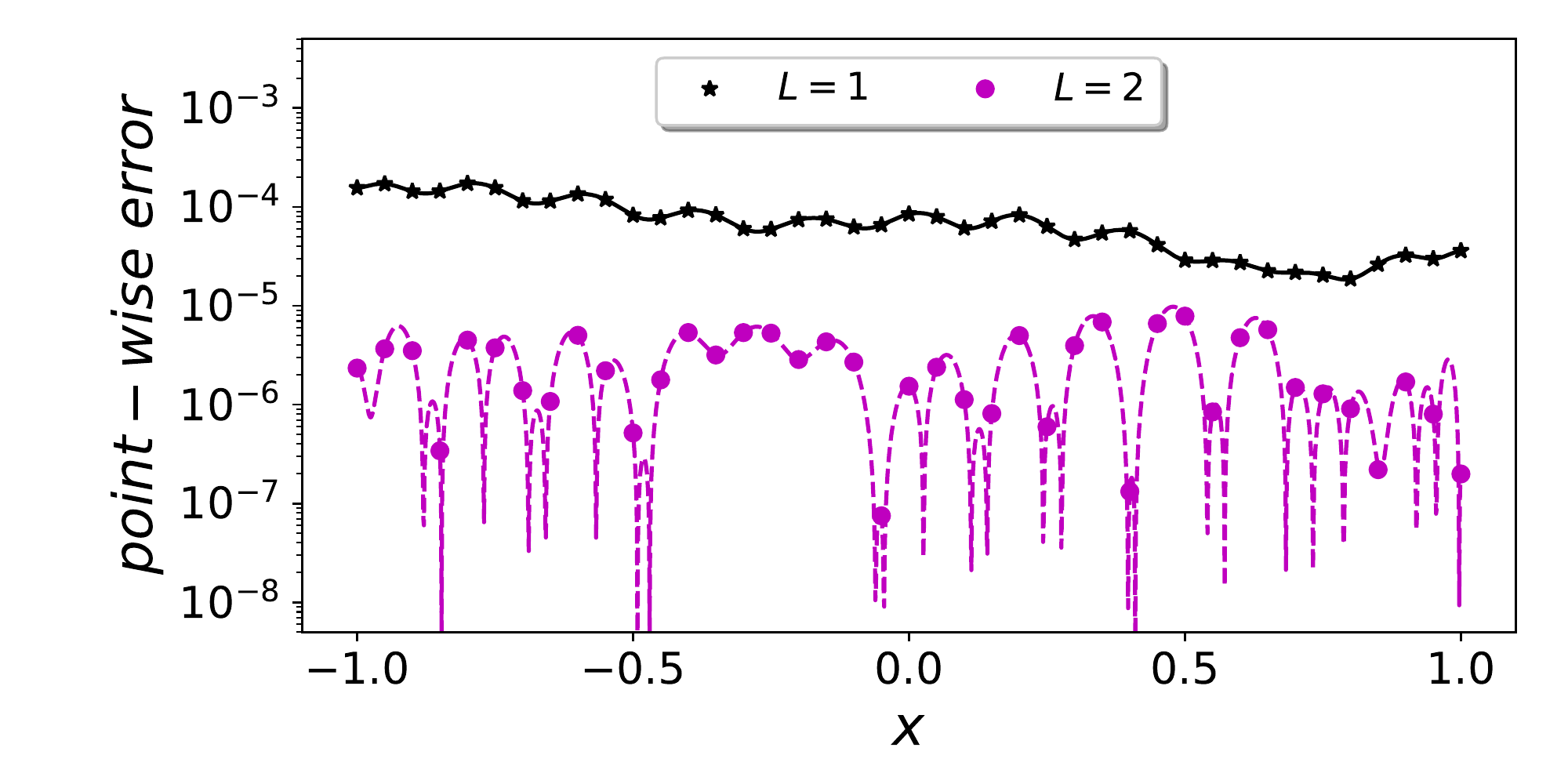}
		&
		\includegraphics[clip, trim=1cm 0cm 0cm 0cm, width=0.46\linewidth]
		{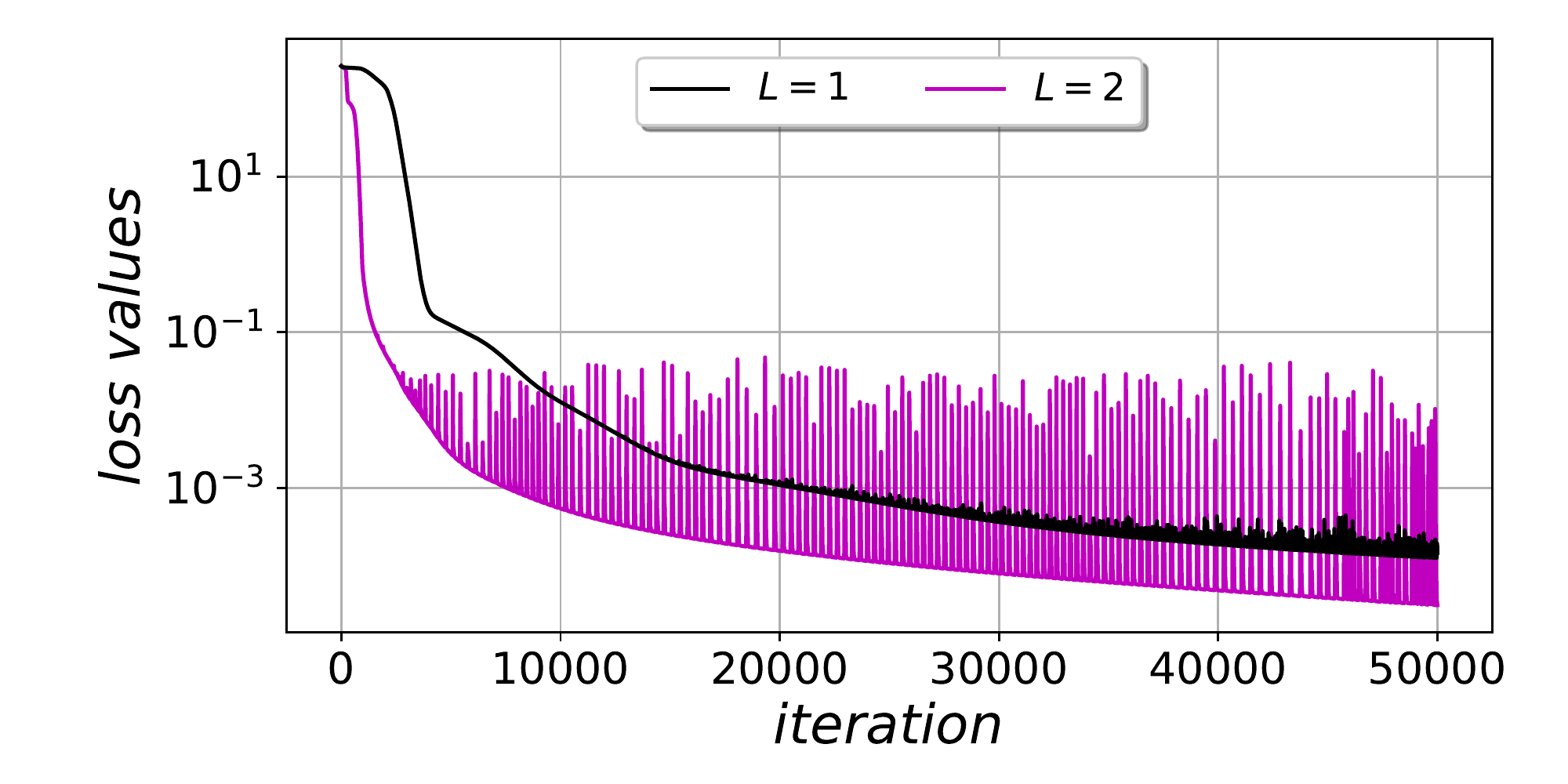}
	\end{tabular}
	\caption{\scriptsize \label{Fig: Poisson steep compare} One-dimensional Poisson's equation. VPINN: $\widetilde{\mathcal{R}}^{(1)}$ and $\widetilde{\mathcal{R}}^{(2)}$ formulations with polynomial test functions $v_k (x) = P_{k+1}(x) - P_{k-1}(x), \,\, k=1,2,\cdots, 60$, $ Q = 100$ Gauss-Jacobi quadrature points, and penalty parameter $\tau = 25$. PINN: 500 randomly selected penalizing points and penalty parameter $\tau = 10$. The network has $N=20$ neurons in each layer with tanh activation. The errors are averaged over at least 5 different network initializations.}
\end{figure}

\begin{figure}[h]
	\center
	\begin{tabular}{c l}
		\multicolumn{2}{c}{$\quad\quad u^{exact} =0.1 \sin(4 \pi x) + e^{\frac{0.01-(x+1)}{0.01}}$} \\  [-2 pt] 
		\multicolumn{2}{c}{\includegraphics[clip, trim=1cm 0cm 0cm 0cm, width=0.46\linewidth]{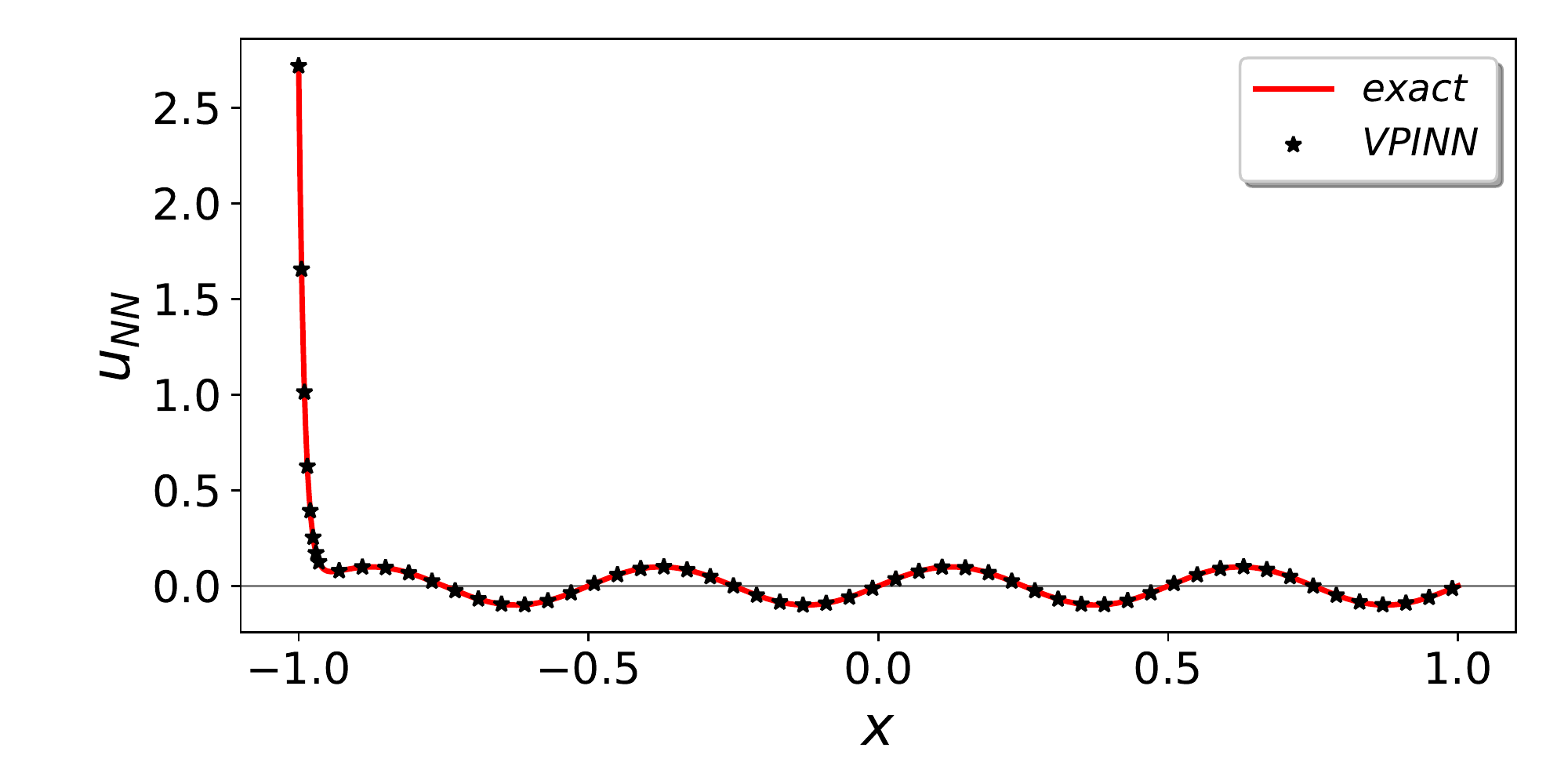}} \\ 
		\multicolumn{2}{c}{VPINN $\widetilde{\mathcal{R}}^{(1)}$} \\  [-1.5 pt]  
		\includegraphics[clip, trim=1cm 0cm 0cm 0cm, width=0.46\linewidth]
		{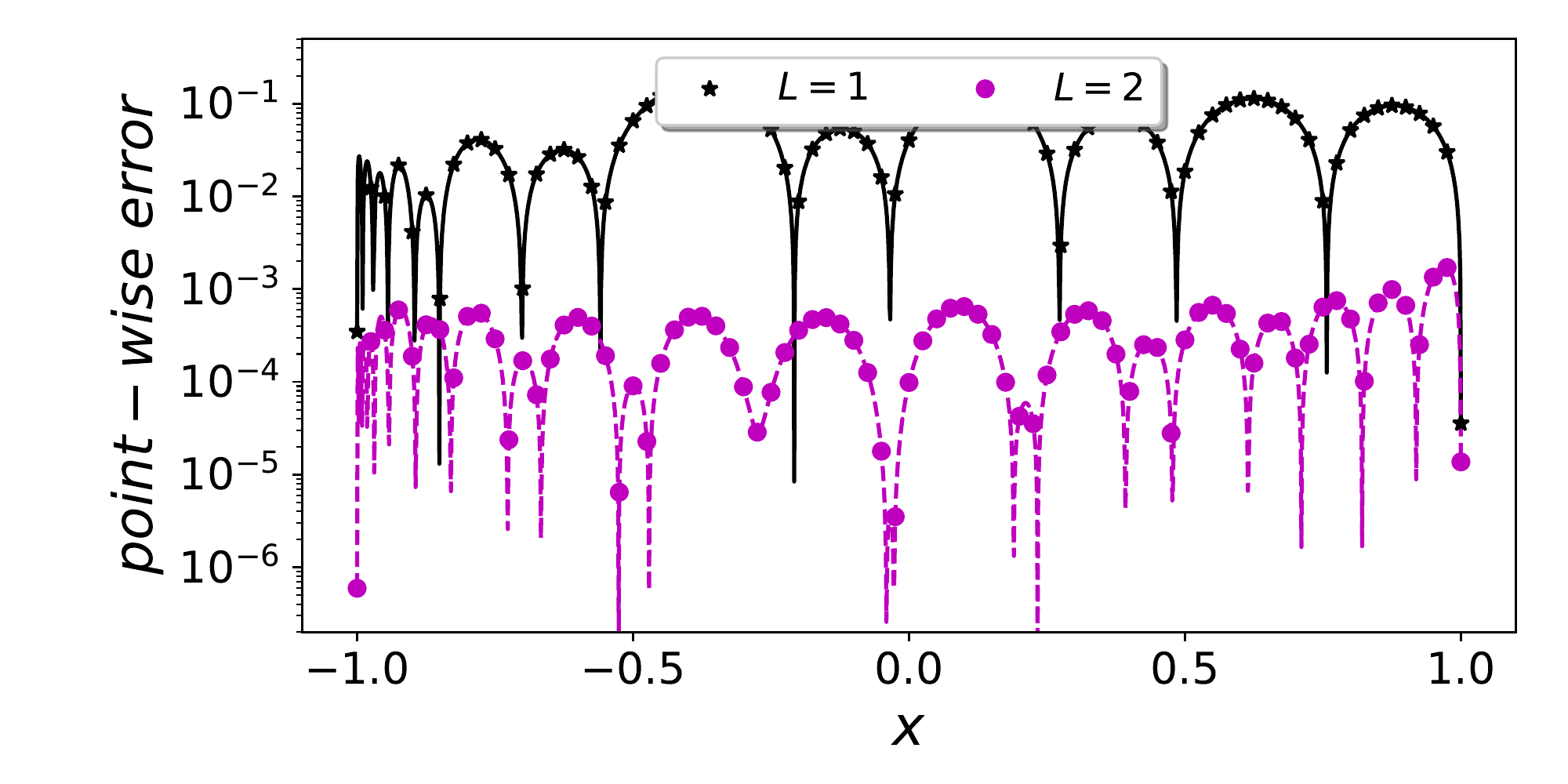}
		&
		\includegraphics[clip, trim=1cm 0cm 0cm 0cm, width=0.46\linewidth]
		{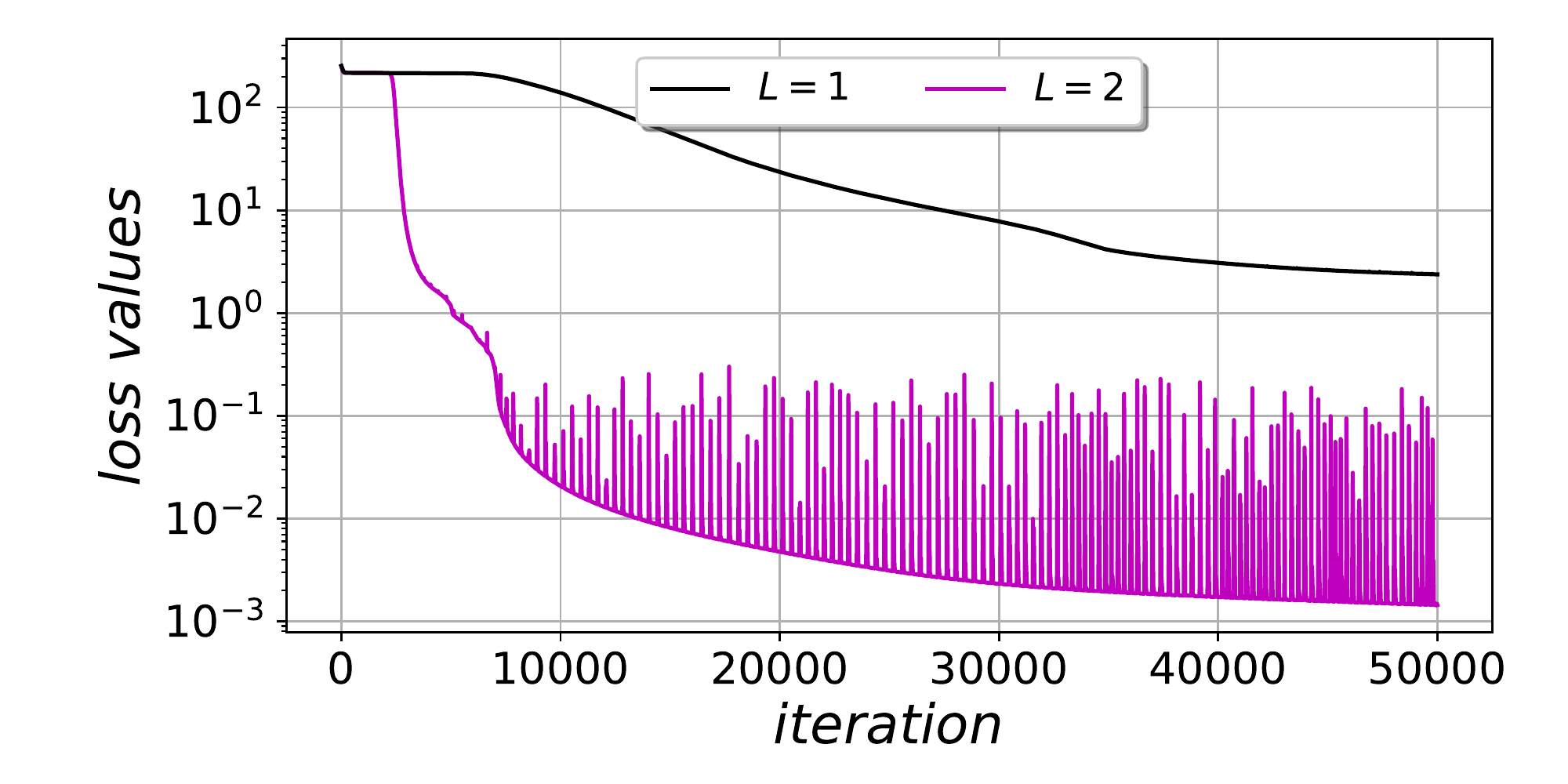}
		\\
		\multicolumn{2}{c}{VPINN $\widetilde{\mathcal{R}}^{(2)}$} \\  [-1.5 pt] 
		\includegraphics[clip, trim=1cm 0cm 0cm 0cm, width=0.46\linewidth]
		{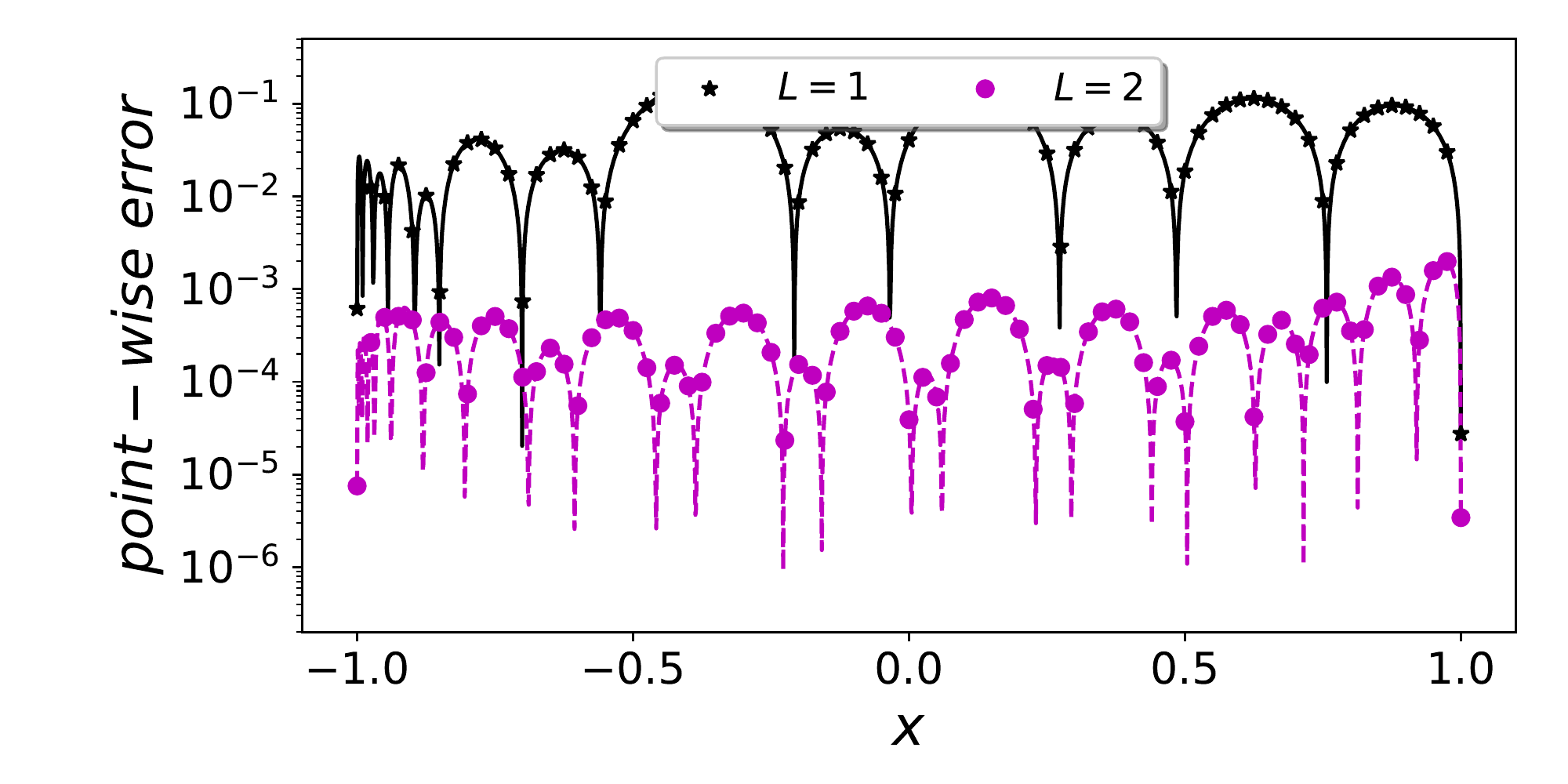}
		&
		\includegraphics[clip, trim=1cm 0cm 0cm 0cm, width=0.46\linewidth]
		{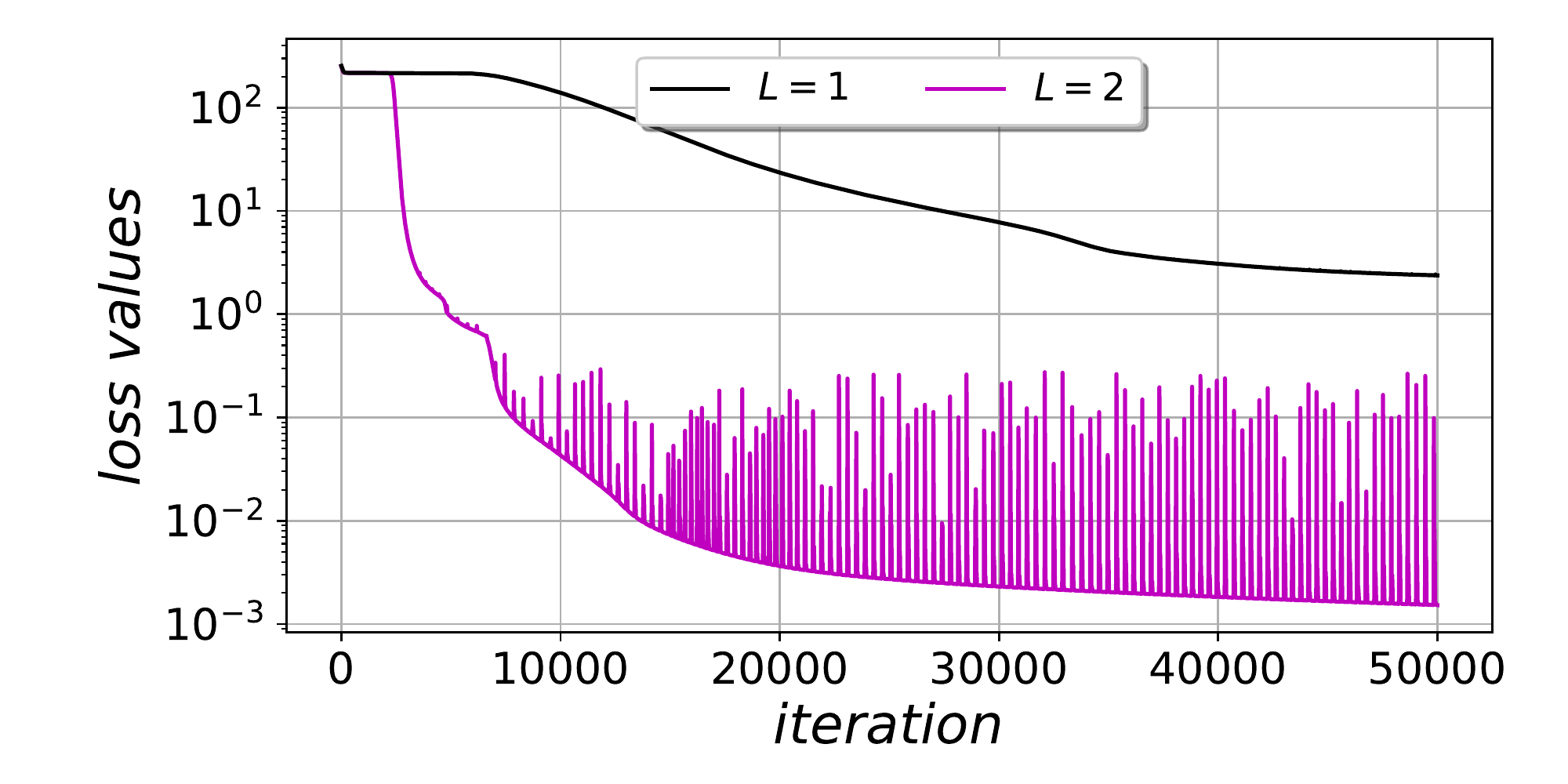}
		\\
		\multicolumn{2}{c}{PINN} \\  [-1.5 pt] 
		\includegraphics[clip, trim=1cm 0cm 0cm 0cm, width=0.46\linewidth]
		{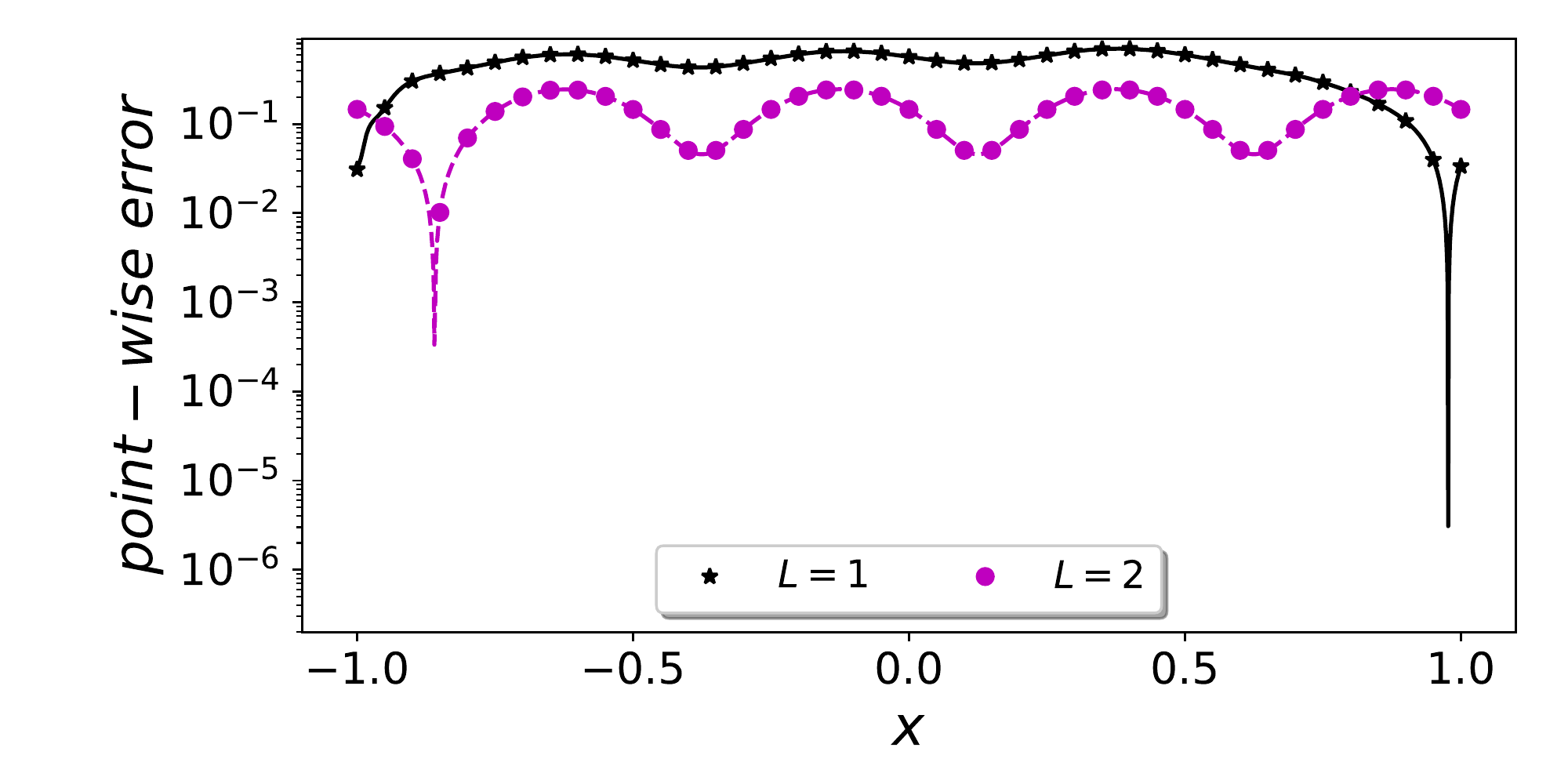}
		&
		\includegraphics[clip, trim=1cm 0cm 0cm 0cm, width=0.46\linewidth]
		{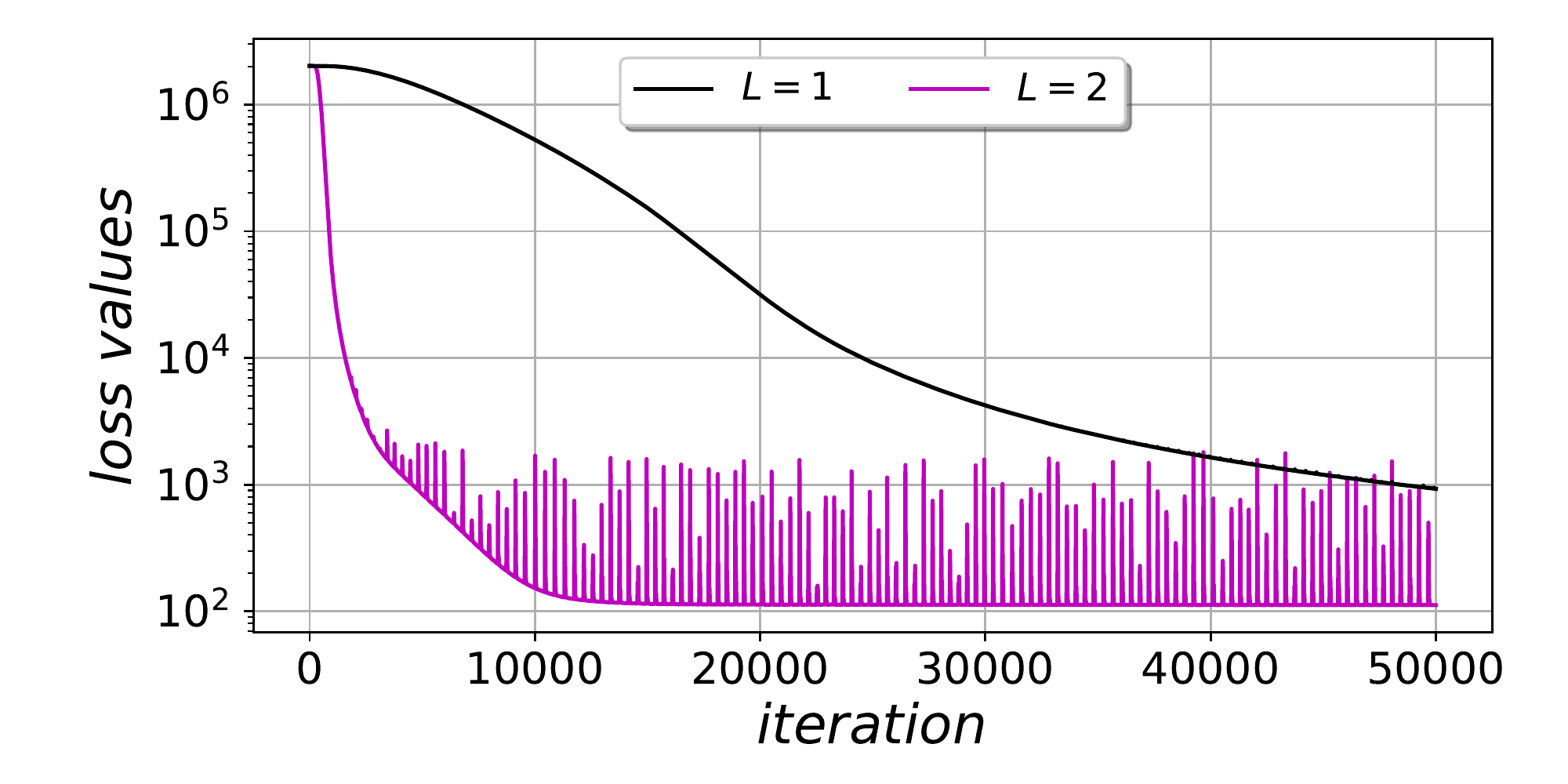}
	\end{tabular}
	\caption{\scriptsize \label{Fig: Poisson boundarylayer compare} One-dimensional Poisson's equation. VPINN: $\widetilde{\mathcal{R}}^{(1)}$ and $\widetilde{\mathcal{R}}^{(2)}$ formulations with polynomial test functions $v_k (x) = P_{k+1}(x) - P_{k-1}(x), \,\, k=1,2,\cdots, 60$, $ Q = 100$ Gauss-Jacobi quadrature points, and penalty parameter $\tau = 10$. PINN: 500 randomly selected penalizing points and penalty parameter $\tau = 10$. The network has $N=20$ neurons in each layer with tanh activation. The errors are averaged over at least 5 different network initialization.}
\end{figure}

%\vspace{0.3 in}
%
\begin{exm}[One-Dimensional Poisson's Equation]
	\label{Ex: 1D Poisson-quad}	
	We consider the Poisson's equation given in \eqref{Eq: Poisson Eqn}. We let the exact solution be of different forms 
	\begin{align}
	\label{Eq: exact solution steep}
	\text{steep solution:  }&  u^{exact}(x) =  0.1 \sin(4 \pi x) + \tanh(5 x)
	\\
	\label{Eq: exact solution BL-2}
	\text{boundary layer solution:  }&  u^{exact}(x) = 0.1 \sin(4 \pi x) + e^{\frac{0.01-(x+1)}{0.01}}
	\end{align}
	The force term is obtained by substituting $u^{exact}(x)$ into the equation. We show the results in Figs. \ref{Fig: Poisson steep compare} and \ref{Fig: Poisson boundarylayer compare}.
\end{exm}

%
%******************************************************************************************
%******************************************************************************************

%
%******************************************************************************************
%******************************************************************************************

%\vspace{0.1 in}
\noindent $\bullet$ \textbf{Discussion:} As expected, by increasing the number of hidden layer while keeping the other hyperparameters unchanged, we observe that the error drops by magnitudes of orders. This explicitly shows the expressivity of deep networks compared to shallow ones. By employing similar optimizer, however, we see that the training process and loss convergence become less stable as there are large numbers of spikes in the loss value over the training iterations. We also observe that the depth of network changes the learning pattern as the sudden drop in the loss value plots happens much earlier. Compared to PINN, the results are relatively similar for the case of steep solution. However, for the boundary layer solution, VPINN performs much more accurately. In this case, for PINN to capture the very sharp boundary layer, we need to cluster a large number of penalizing points in the neighborhood of boundary layer. We also note the oscillatory behavior of point-wise error in VPINN compared to PINN, which is expected due to the projection of neural network residuals onto the hierarchical polynomial test functions.

\begin{figure}[h]
	\center
	\includegraphics[clip, trim=2.5cm 0cm 0cm 0.1cm, width=0.3\linewidth]{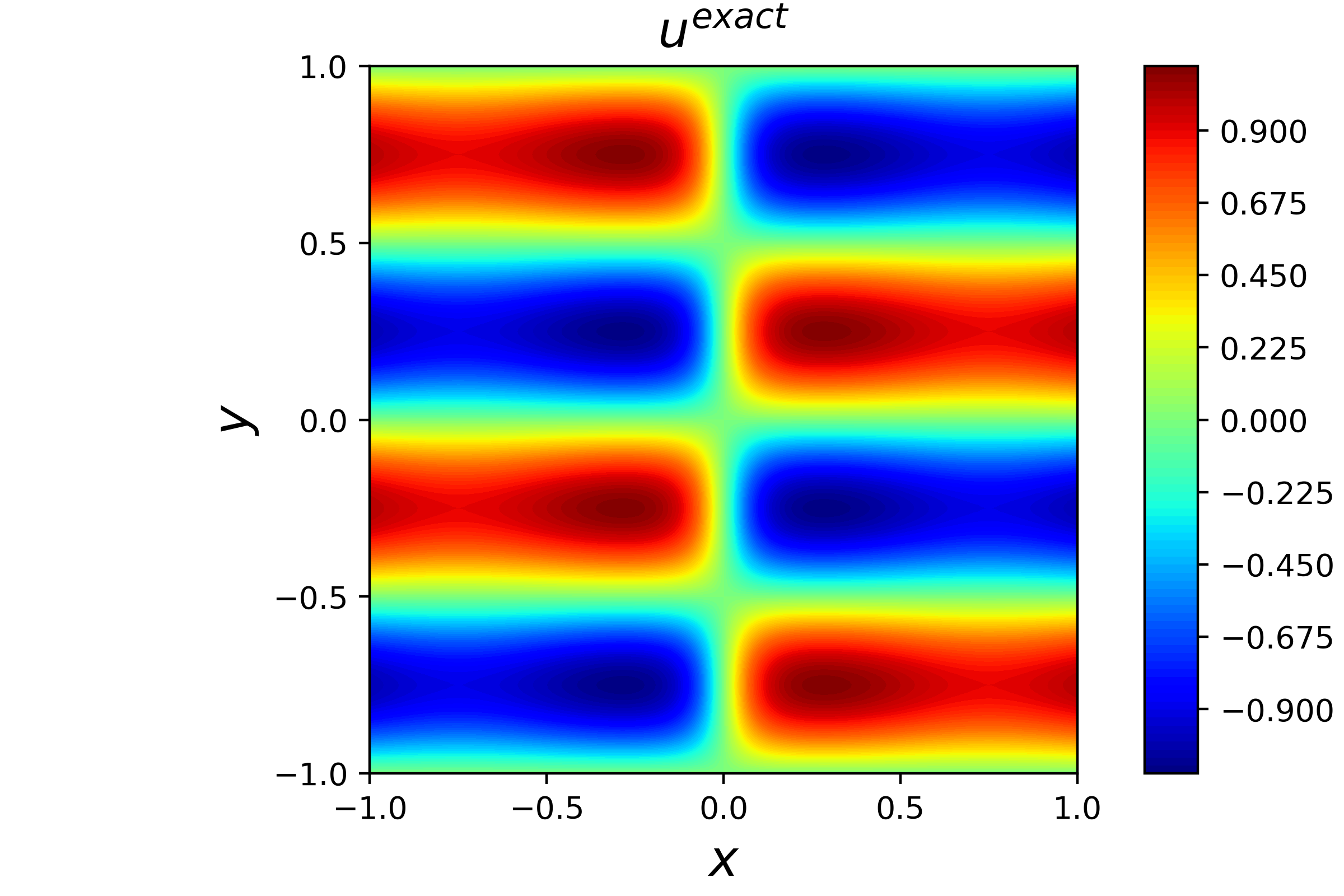}
	\includegraphics[clip, trim=2.5cm 0cm 0cm 0.1cm, width=0.3\linewidth]{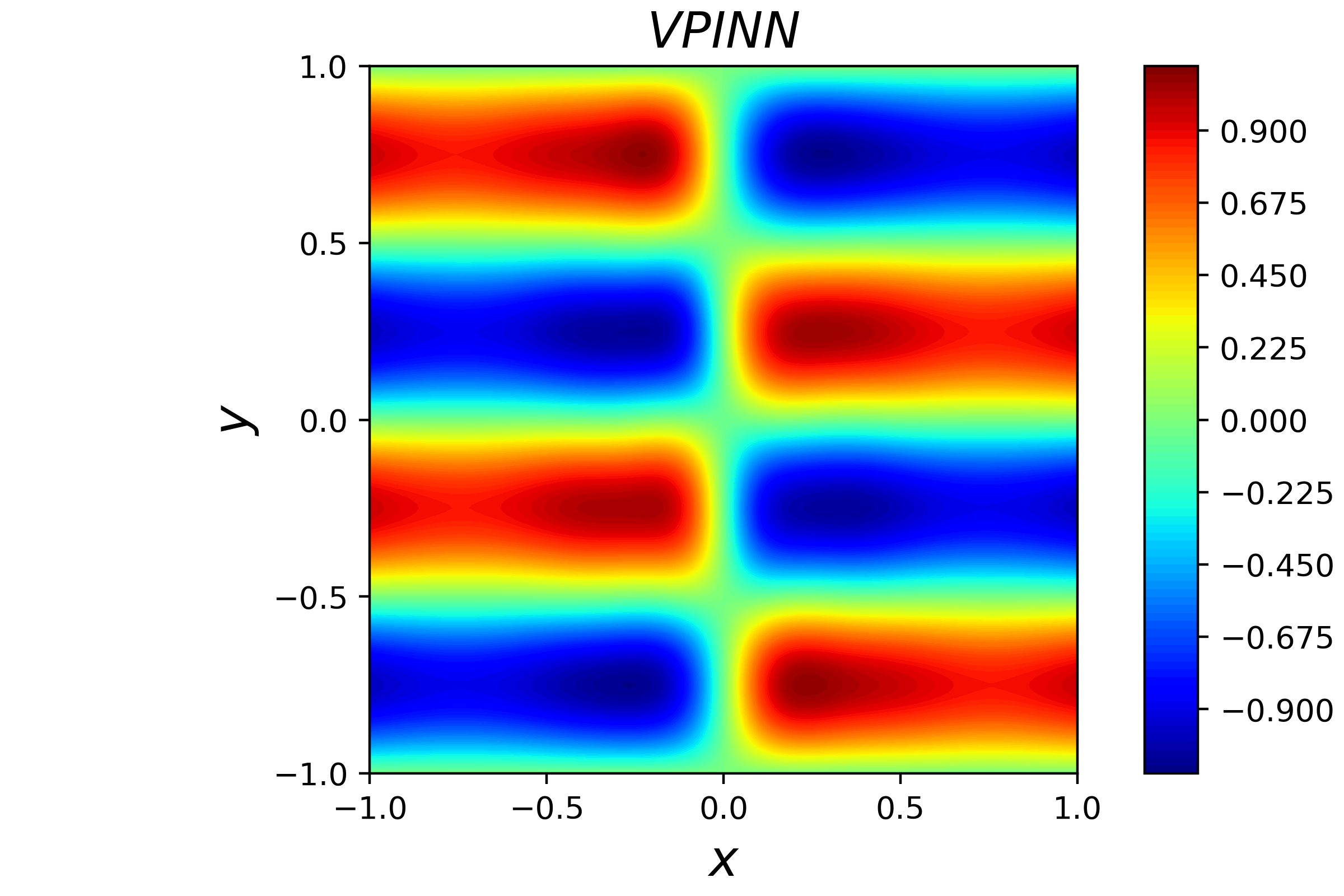}
	\includegraphics[clip, trim=2.5cm 0cm 0cm 0.1cm, width=0.3\linewidth]{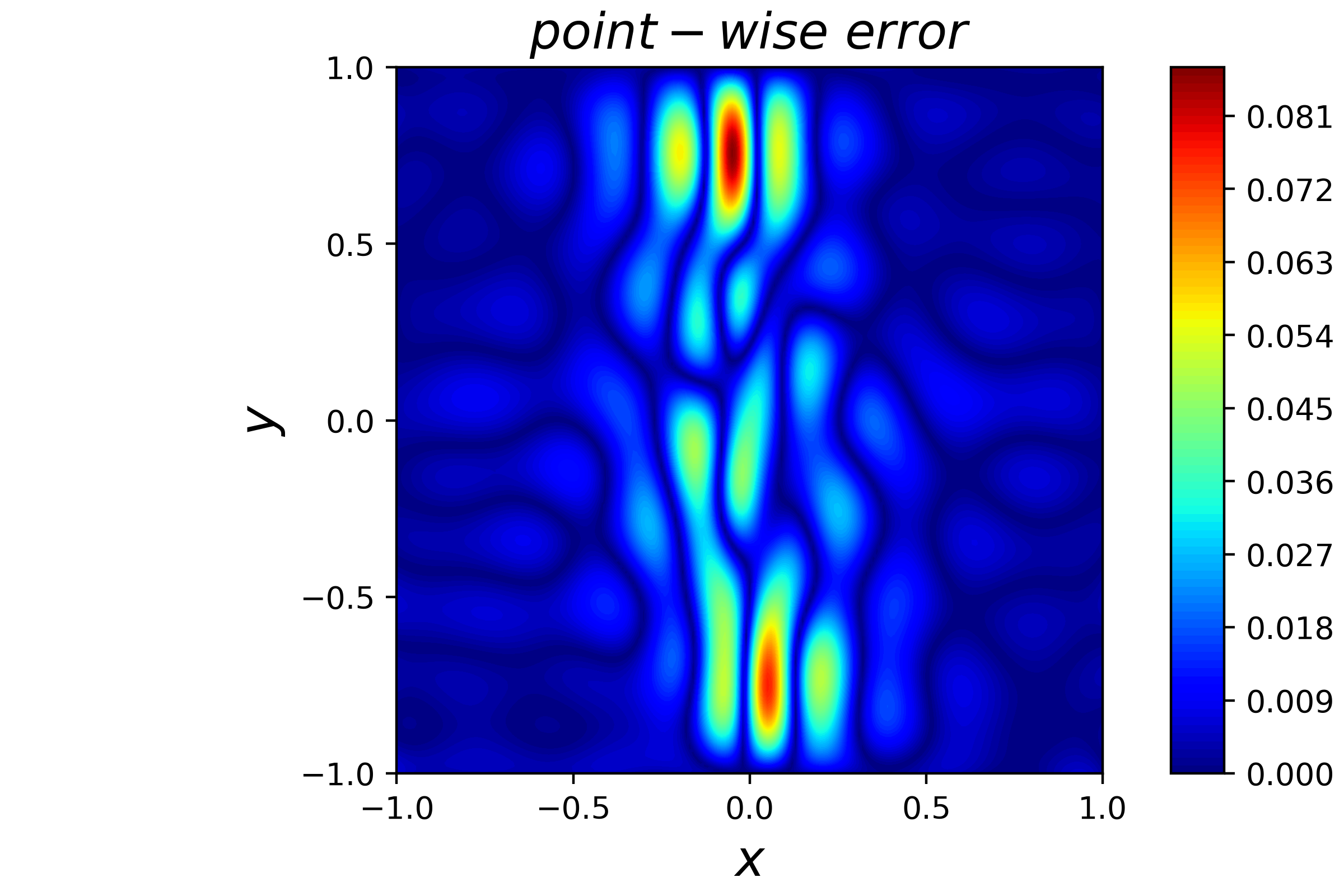}	%
	\caption{\scriptsize \label{Fig: Poisson 2 D} Two-dimensional Poisson's equation. Left: exact solution \eqref{Eq: exact solution 2-d}. Middle: VPINN. Right: point-wise error. Neural network, optimizer, and VPINN parameters are given in Table \ref{Table: VPINN para}.  }
\end{figure}

\begin{figure}[h]
	\center
	\begin{tabular}{c}
		{\scriptsize  x slices, \,\, {\color{red} \textemdash{}} Exact, \,\, $\sqbullet \sqbullet \sqbullet$  VPINN } \\ \hline \\ [-6 pt]
		\includegraphics[clip, trim=1cm 3.5cm 0.5cm 0.5cm, width=0.8\linewidth]{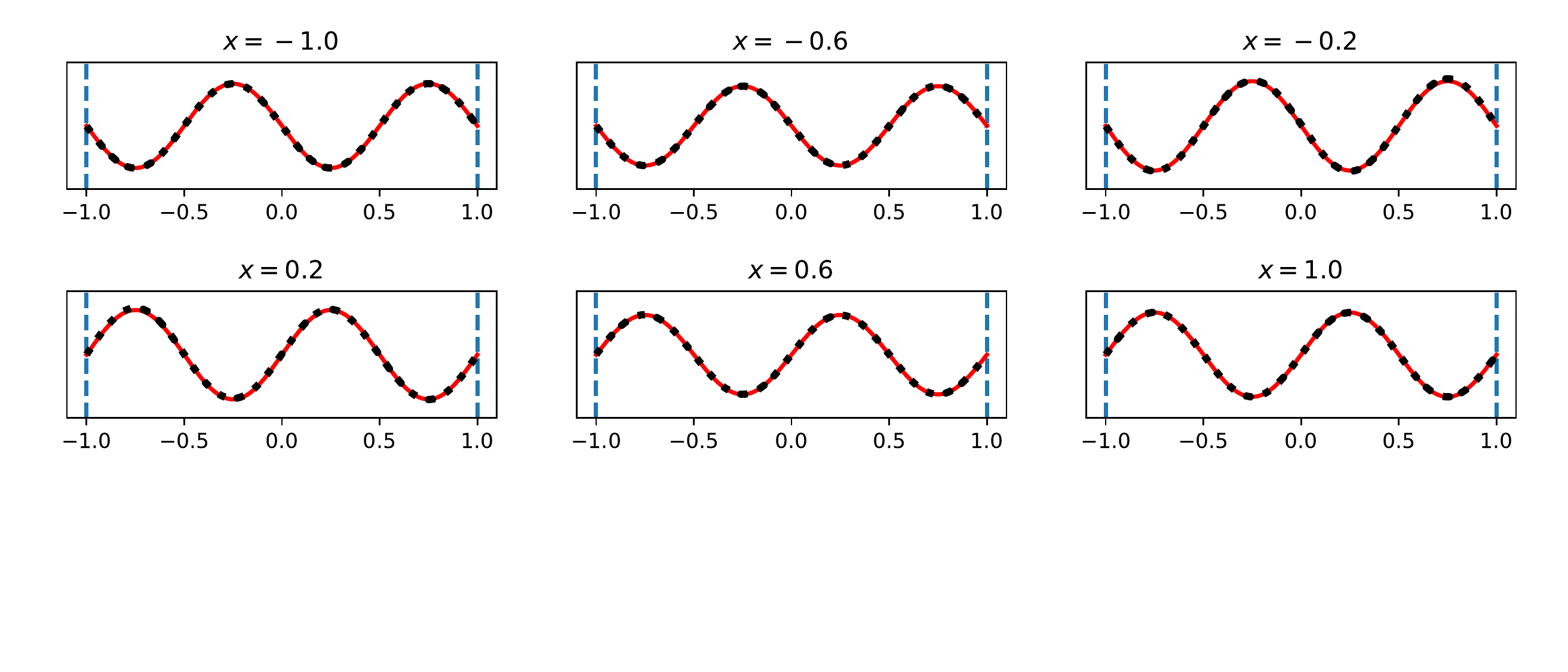}
		\\ 	\hline\hline \\ [-5 pt] 
		{\scriptsize  y slices, \,\, {\color{red} \textemdash{}} Exact, \,\, $\sqbullet \sqbullet \sqbullet$  VPINN } \\ \hline \\ [-6 pt]
		\includegraphics[clip, trim=1cm 3.5cm 0.5cm 0.5cm, width=0.8\linewidth]{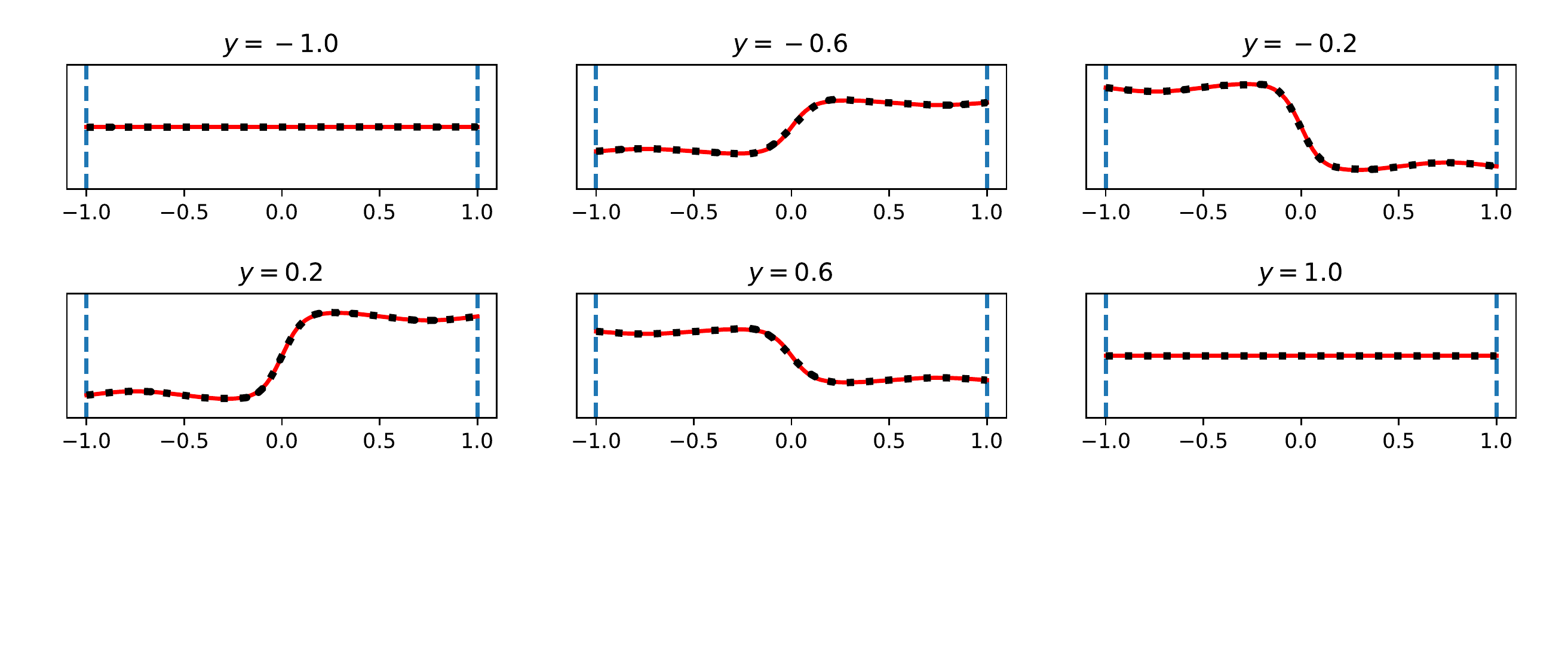}
		\\
		\hline\hline
	\end{tabular}
	\caption{\scriptsize \label{Fig: Poisson 2 D slices} Two-dimensional Poisson's equation with solution \eqref{Eq: exact solution 2-d}. The red line is the exact solution and the dotted black line is VPINN. Neural network, optimizer, and VPINN parameters are given in Table \ref{Table: VPINN para}.}
\end{figure}

\vspace{1 in}
\begin{exm}[Two-Dimensional Poisson's Equation]
	\label{Ex: 2D Poisson-quad}	
	We consider the following two-dimensional Poisson's equation
	\begin{align}
	\label{Eq: 2-d Poisson}
	\Delta u(x,y) = f(x,y), 
	\end{align}
	subject to homogeneous Dirichlet boundary conditions over $\Omega = [-1,1] \times [-1,1]$, where we assume that the forcing term $f(x,y)$ is available at some quadrature points. We consider the exact solution as
	\begin{align}
	\label{Eq: exact solution 2-d}
	u^{exact}(x, y) =  \left( 0.1 \sin(2 \pi x) + \tanh(10 x) \right) \times \sin(2 \pi y),
	\end{align}
	where the forcing function is obtained by substituting the exact solution in \eqref{Eq: 2-d Poisson}. We show the results in Figs. \ref{Fig: Poisson 2 D} and \ref{Fig: Poisson 2 D slices}.
\end{exm}

%\vspace{20 in}

\begin{table}[h]
	\center
	\caption{\scriptsize{ \label{Table: VPINN para} Two-Dimensional Poisson's Equation: Neural network, optimizer, and VPINN parameters.}}
	\vspace{-0.1 in}
	\scalebox{0.7}{
		\begin{tabular}
			{l l}
			\multicolumn{2}{c}{VPINN}\\
			\hline
			\hline
			variational form & $\mathcal{R}^{(2)}$ \\ \hline
			$\#$ test function  & $K_x=10,  K_y = 10$     \\ \hline 
			test functions & $P_{k+1}(.) - P_{k-1}(.)$  \\ \hline 
			$\#$ quadrature points &  $Q_x = 70, Q_y =70$  (Gauss-Lobatto)    \\ \hline
			$\#$ (boundary) training points & $x:\, 2 \times 80$, $y: \, 2 \times 80$    \\ \hline			
			\hline
			%\vspace{0.2 in}
			%
		\end{tabular}
	}
	\scalebox{0.7}{
		\begin{tabular}
			{l l}
			\multicolumn{2}{c}{NN}\\
			\hline
			\hline
			$\#$ hidden layers  & 4  \\ \hline 
			$\#$ neurons in each layer & 20 \\ \hline   
			activation function & sine \\ \hline
			optimizer & Adam \\ \hline 
			learning rate & $10^{-3}$ \\ \hline
			\hline
			%\vspace{0.2 in}
			%
		\end{tabular}
	}
	%
	%	\vspace{-0.1 in}
\end{table}
%

%
%******************************************************************************************
%******************************************************************************************
\vspace{-0.3 in}
%
%******************************************************************************************
%******************************************************************************************

Given a neural network that returns an approximation $u_{NN}(x,y)$ and similar to the one dimensional case, we define $r(x,y) = \Delta u_{NN}(x,y) - f(x)$ as the network residual. To define the variational residual of the network, we choose proper test functions $v(x,y) \in V$ and then following similar formulation as in \eqref{Eq: 1-d BVP var residue - 1}-\eqref{Eq: 1-d BVP var residue - 3}, we have 
\begin{align}
\label{Eq: 2-d Poisson var residue - 1}
\mathcal{R} &= \left( \Delta \, u_{NN}(x,y), v(x,y) \right)_{\Omega}  ,
\\
\nonumber
F &= \left( f(x,y) , v(x,y)  \right)_{\Omega} .
\end{align}
The finite dimensional space of test functions is comprised of the tensor product of the subspaces $V_x = \text{span} \lbrace  \phi_{k_x}(x), \,\, k_x=1,2, \cdots, K_x   \rbrace$, and $V_y = \text{span} \lbrace  \phi_{k_y}(y), \,\, k_y=1,2, \cdots, K_y   \rbrace$. Similar to one-dimensional problem setting, we can define three distinctive \emph{variational residuals} $\mathcal{R}^{(1)}$, $\mathcal{R}^{(2)}$, and $\mathcal{R}^{(3)}$ by taking integration-by-parts, and thus, similarly define three distinctive variational loss functions. Table \ref{Table: VPINN para} shows the structure of the network and the parameters used in VPINN. Figure \ref{Fig: Poisson 2 D} shows the exact solution, prediction, and point-wise error. Figure \ref{Fig: Poisson 2 D slices} shows $x$, and $y$ slices of exact solution and prediction, respectively.

\vspace{0.1 in}
\noindent $\bullet$ \textbf{Discussion:} The exact solution of \eqref{Eq: exact solution 2-d} has a steep change along the $x$ direction and a sinusoidal behavior in the $y$ direction, where each behavior is individually studied in the previous one-dimensional problems. Thus, by employing similar test functions in each direction, we observe that VPINNs can accurately capture the exact solution. The point-wise error in Fig. \ref{Fig: Poisson 2 D} shows that the maximum error occurs close to the steep change, while in the other regions, the error is relatively smaller. This can be also observed from the slight mismatch of VPINN and exact solution close to the steep part of solution in the y slices in Fig. \ref{Fig: Poisson 2 D slices}.  

In higher-dimensional problems, the computation of integrals in the variational residuals in VPINNs requires multi-dimensional numerical integrations, which can further impose extra computational cost. In example \ref{Ex: 2D Poisson-quad}, we employ 70 quadrature points in each direction and thus need to evaluate the integrands in the variational residuals at $70\times70$ points. For the case of more complicated solutions, higher dimensional problems, and employing higher number of test functions, one need to further increase the quadrature points, which adversely affect the computational costs. Other numerical methods, such as sparse grids, can be employed to help this issue.

%%
%
%%%%%%%%%%%%%%%%%%%%%%%%%%%
\section{Summary}
\label{Sec: sum}
%%%%%%%%%%%%%%%%%%%%%%%%%%%
%
We developed the variational physics-informed neural network (VPINN) in the context of a Petrov-Galerkin method based on the nonlinear approximation of DNNs as the trial functions and polynomials and trigonometric functions as test functions. We showed that since VPINN considers the variational form of the underlying mathematical model, it has the advantage of reducing the order of differential operator by integration-by-parts, which can effectively lower the required regularity in the output of NN. For the case of shallow networks with one hidden layers, we analytically obtained distinct forms of the variational residual of the network, which can open up possibilities of further investigating the numerical analysis of VPINN. However, a deep network can provide a more accurate approximation, and therefore, the numerical integration of the variational formulation is a necessity in the case of deep networks. The VPINNs formulation compared to PINNs penalizes the strong-form residual of the network by testing it with several test functions, and thus replaces the penalizing points with quadrature points. To the best of our knowledge, there is no proper quadrature rule in the literature developed for integrals of DNNs, and thus we aim to carry out an extensive investigation on this subject in our future works, by taining NN to perform such integrations. 
%%

%\newpage
\appendix
%
%%%%%%%%%%%%%%%%%%%%%%%
\section{Derivation of Variational Residuals for Shallow Network With Sine Activation Functions and Sine Test Functions}
\label{Sec: Appx shallow sine}
%%%%%%%%%%%%%%%%%%%%%%%
%
We chose the sine activation functions and sine test functions to obtain the variational residuals. The variational residual $\mathcal{R}^{(2)}_k$ is given in \eqref{Eq: 1-d BVP var residue - 2}. We assume that $w_j \neq k \pi$ and therefore, have
\begin{align*}
%\label{Eq: 1-d BVP var residue - 2}
\mathcal{R}^{(2)}_k 
&= \int_{-1}^{1} \frac{d u_{NN}(x)}{d x} \frac{d v_k(x)}{d x} dx 
\\ \nonumber
&= \sum_{j=1}^{N} a_j \, w_j \, k \, \pi  
\int_{-1}^{1} \cos (w_j \, x + \theta_j ) \cos(k \pi x) dx 
\\ \nonumber
%%
%&= \frac{k \, \pi}{2} \sum_{j=1}^{N} a_j \, w_j  
%\int_{-1}^{1} \cos ( (w_j + k \pi ) x + \theta_j ) + \cos ( (w_j - k \pi ) x + \theta_j ) \, dx 
%\\ \nonumber
%%
%&= \frac{k \, \pi}{2} \sum_{j=1}^{N} 
%  \frac{a_j \, w_j}{w_j + k \pi} \sin ( (w_j + k \pi ) x + \theta_j ) \Big|_{-1}^{1} 
%+ \frac{a_j \, w_j}{w_j - k \pi} \sin ( (w_j - k \pi ) x + \theta_j ) \Big|_{-1}^{1}
%\\ \nonumber
%
&= \frac{k \, \pi}{2} \sum_{j=1}^{N} 
  \frac{a_j \, w_j}{w_j + k \pi} \left[ \sin ( w_j + k \pi + \theta_j ) - \sin ( - w_j - k \pi + \theta_j ) \right] 
\\ \nonumber & \quad\quad\quad
+ \frac{a_j \, w_j}{w_j - k \pi} \left[ \sin ( w_j - k \pi + \theta_j ) - \sin ( - w_j + k \pi + \theta_j ) \right]
\\ \nonumber
&= k \, \pi \sum_{j=1}^{N} a_j \, w_j \, \cos (\theta_j ) \,  
\left[ \frac{\sin ( w_j + k \pi)}{w_j + k \pi} + \frac{\sin ( w_j - k \pi)}{w_j - k \pi}  \right]
\\ \nonumber
&=(-1)^k k \, \pi \sum_{j=1}^{N} a_j \, w_j \, \cos (\theta_j ) \, \sin ( w_j )\,   
\left[ \frac{1}{w_j + k \pi} + \frac{1}{w_j - k \pi}  \right]
\\ \nonumber
&=2(-1)^k \, k \, \pi \, \sum_{j=1}^{N} \frac{a_j \, w_j^2 \, \cos(\theta_j) \sin(w_j)}{w_j^2 - k^2 \pi^2}.
\end{align*}
The variational residual $\mathcal{R}^{(3)}_k$ is given in \eqref{Eq: 1-d BVP var residue - 3}. For the first term, we have
\begin{align*}
%\label{Eq: 1-d BVP var residue - 3}
& - \int_{-1}^{1} u_{NN}(x) \frac{d^2 v_k(x)}{d x^2} dx 
\\ \nonumber
&= \sum_{j=1}^{N} a_j \, k^2 \, \pi^2  
\int_{-1}^{1} \sin (w_j \, x + \theta_j ) \sin(k \pi x) dx 
\\ \nonumber
&= \frac{k^2 \, \pi^2}{2} \sum_{j=1}^{N} a_j   
\int_{-1}^{1} \cos ( (w_j - k \pi ) x + \theta_j ) - \cos ( (w_j + k \pi ) x + \theta_j ) \, dx 
\\ \nonumber
%%
%&= \frac{k^2 \, \pi^2}{2} \sum_{j=1}^{N} a_j 
%\left[ \frac{\sin ( (w_j - k \pi ) x + \theta_j )}{w_j - k \pi}  
%- \frac{\sin ( (w_j + k \pi ) x + \theta_j )}{w_j + k \pi}  \right] \Big|_{-1}^{1}
%\\ \nonumber
%%
%&= \frac{k^2 \, \pi^2}{2} \sum_{j=1}^{N} a_j 
%\left[ \frac{\sin ( w_j - k \pi + \theta_j ) - \sin ( - w_j + k \pi + \theta_j )}{w_j - k \pi}  
%- \frac{\sin ( w_j + k \pi + \theta_j ) - \sin ( - w_j - k \pi + \theta_j )}{w_j - k \pi}  \right] 
%\\ \nonumber
%
&= k^2 \, \pi^2 \sum_{j=1}^{N} a_j 
\left[ \frac{\cos ( \theta_j ) \sin ( w_j - k \pi )}{w_j - k \pi}  
- \frac{\cos ( \theta_j ) \sin ( w_j + k \pi )}{w_j + k \pi}  \right] 
\\ \nonumber
&= k^2 \, \pi^2 \sum_{j=1}^{N} a_j \cos ( \theta_j ) 
\left[ \frac{\sin ( w_j - k \pi )}{w_j - k \pi}  
- \frac{ \sin ( w_j + k \pi )}{w_j + k \pi}  \right]
\\ \nonumber
&= (-1)^k k^2 \, \pi^2 \sum_{j=1}^{N} a_j \cos ( \theta_j ) \, \sin ( w_j ) 
\left[ \frac{1}{w_j - k \pi} - \frac{ 1}{w_j + k \pi}  \right]
\\ \nonumber
&=2 (-1)^k k^3 \, \pi^3 \sum_{j=1}^{N}  \frac{a_j \cos ( \theta_j ) \, \sin ( w_j )}{w_j^2 - k^2 \pi^2}
\\ \nonumber
\end{align*}
Also, the boundary term becomes
\begin{align*}
&u_{NN}(x) \, \frac{d v_k(x)}{d x} \bigg\vert_{-1}^{1}
\approx u(x) \, \frac{d v_k(x)}{d x} \bigg\vert_{-1}^{1}
= u(x) \, k \, \pi \, \cos(k\pi x) \bigg\vert_{-1}^{1}
%\\ \nonumber
%= & h \, k \, \pi \, \cos(k\pi) - g \, k \, \pi \, \cos(-k\pi)
= (-1)^k \, k \, \pi \, (h- g) .
\end{align*}
Therefore, 
\begin{align}
\mathcal{R}^{(3)}_k &= 2 (-1)^k \, k \, \pi \, \left[ \frac{h-g}{2} + k^2 \, \pi^2 \, \sum_{j=1}^{N} \frac{a_j \, \cos(\theta_j) \sin(w_j)}{w_j^2 - k^2 \pi^2} \right]
.
\end{align}
The nonlinear part of the variational residual for the steady state Burger's equation takes the form
\begin{align*}
%\label{Eq: St Burger var residue - 1}
%
\mathcal{R}^{NL}_k 
&= \left( u_{NN}(x)\frac{d u_{NN}(x)}{d x}, v_k(x) \right)_{\Omega} ,
\\ \nonumber
& = \sum_{i=1}^{N}\sum_{j=1}^{N} a_i \, a_j \, w_i  
\int_{-1}^{1} \sin (w_j \, x + \theta_j ) \cos (w_i \, x + \theta_i ) \sin(k \pi x) dx 
\\ \nonumber
& = \sum_{i=1}^{N}\sum_{j=1}^{N} a_i \, a_j \, w_i  
\int_{-1}^{1} \frac{\sin ((w_j + w_i) x + \theta_j + \theta_i ) + \sin ((w_j - w_i) x + \theta_j - \theta_i ) }{2} \sin(k \pi x) dx ,
\end{align*}
and using the same steps as in derivations of $\mathcal{R}^{(3)}_k$, we obtain
\begin{align*}
\mathcal{R}^{NL}_k 
& =  k \pi \sum_{i=1}^{N}\sum_{j=1}^{N} a_i \, a_j \, w_i
\Big[ 
\frac{\sin ( w_j + w_i) \cos( \theta_j + \theta_i + k\pi )}{(w_j+w_i)^2-k^2\pi^2} 
+ \frac{\sin ( w_j - w_i) \cos( \theta_j - \theta_i + k\pi )}{(w_j-w_i)^2-k^2\pi^2} 
\Big]
\\ \nonumber
& =  (-1)^k\,k \pi \sum_{i=1}^{N}\sum_{j=1}^{N} a_i \, a_j \, w_i
\Big[ 
\frac{\sin ( w_j + w_i) \cos( \theta_j + \theta_i )}{(w_j+w_i)^2-k^2\pi^2} 
+ \frac{\sin ( w_j - w_i) \cos( \theta_j - \theta_i )}{(w_j-w_i)^2-k^2\pi^2} 
\Big]
\end{align*}
%

%
%%%%%%%%%%%%%%%%%%%%%%%
\section{Derivation of Variational Residuals for Shallow Network With Sine Activation Functions and Legendre Test Functions}
\label{Sec: Appx shallow sine test Leg}
%%%%%%%%%%%%%%%%%%%%%%%
%
Let $P_{k}(x)$ be Legendre polynomials of order $k = 0,1,\cdots$. The recursion formula for construction of Legendre polynomials is given as
\begin{align}
\label{Eq: Legendre recursion}
p_{k}(x) & = \frac{2k-1}{k} \, x \, p_{k-1}(x) -  \frac{k-1}{k} \, p_{k-2}(x), \quad k=2,3,\cdots,
\\
\nonumber
p_0(x) & = 1, \quad p_1(x) = x.
\end{align}
The integral $I_k$ for $k=2,3,\cdots$ takes the form
\begin{align*}
I_k & = \int_{-1}^{1} e^{i \, w_j x} \, p_{k}(x) \, dx = \frac{2k-1}{k} \int_{-1}^{1} e^{i \, w_j x} \, \, x \, p_{k-1}(x) \, dx  -  \frac{k-1}{k} \, \int_{-1}^{1} e^{i \, w_j x} \, p_{k-2}(x) \, dx.
\end{align*}
By integration by parts and using the property $x \, p_{k-1}(x) = \frac{k-1}{2k-1} \, p^{\prime}_{k}(x) +  \frac{k}{2k-1} \, p^{\prime}_{k-2}(x)$, where $( \, ^{\prime} \, )$ denotes $d/dx$, we obtain
\begin{align*}
I_k & = \frac{2k-1}{k} \left[ \mathcal{C}_1 - \frac{1}{i \, w_j} \int_{-1}^{1} e^{i \, w_j x} \, \left( p_{k-1}(x) + x p^{\prime}_{k-1}(x) \right) \, dx \right] -  \frac{k-1}{k} \, \int_{-1}^{1} e^{i w_j x} \, p_{k-2}(x) \, dx,
\\
& = \frac{2k-1}{k} \left[ \mathcal{C}_1 - \frac{1}{i \, w_j} I_{k-1} - \frac{1}{i \, w_j} \left( \frac{k-1}{2k-1} \mathcal{C}_2 +  \frac{k}{2k-1} \mathcal{C}_3 \right) + \frac{k-1}{2k-1} \, I_{k} +  \frac{k}{2k-1} \, I_{k-2}  \right] -  \frac{k-1}{k} \, I_{k-2}
\\
& = - \frac{1}{i \, w_j} (2k-1) I_{k-1} + I_{k-2} + (2k-1) \mathcal{C}_1  - \frac{k}{i \, w_j} \left( \frac{k-1}{k} \mathcal{C}_2 +  \mathcal{C}_3 \right), 
\end{align*}
where
\begin{align*}
\mathcal{C}_1 = \frac{1}{i \, w_j} e^{i \, w_j x} \, x \, p_{k-1}(x) \bigg\vert_{-1}^{1}, \quad
\mathcal{C}_2 = e^{i \, w_j x} \, p_{k}(x) \bigg\vert_{-1}^{1}, \quad
\mathcal{C}_3 = e^{i \, w_j x} \, p_{k-2}(x) \bigg\vert_{-1}^{1}.
\end{align*}
By using the special values $p_k(\pm 1) = (\pm 1)^k$, we can show that $\mathcal{C}_2 = \mathcal{C}_3$ and thus, $$(2k-1) \mathcal{C}_1  - \frac{k}{i \, w_j} \left( \frac{k-1}{k} \mathcal{C}_2 +  \mathcal{C}_3 \right) = 0. $$ Therefore, 
\begin{align*}
I_{k} & = i \frac{2k-1}{w_j} I_{k-1} + I_{k-2}, \quad k=2,3,\cdots,
\\
\nonumber
I_0 & = \frac{2 \sin(w_j)}{w_j}, \quad I_1 = \frac{- 2 i}{w_j} \left( \cos(w_j) - \frac{\sin(w_j)}{w_j} \right).
\end{align*}
Subsequently, we can simply obtain the recursion formulas for $C_k$ and $B_k$.

%%\clearpage

%\newpage
\bibliographystyle{siam}
\bibliography{BIB_Temp}

\end{document}